\documentclass[10pt,letterpaper]{article}

\usepackage{wacv}
\usepackage{times}
\usepackage{epsfig}
\usepackage{graphicx}
\usepackage{amsmath}
\usepackage{amssymb}
\usepackage{booktabs}
\usepackage{algorithm}
\usepackage{algpseudocode}
\usepackage{multirow}
\usepackage{multicol}
\usepackage{adjustbox}
\usepackage{subcaption}
\usepackage{bbding}
\usepackage{pifont}
\usepackage{wasysym}
\usepackage{amssymb}
\wacvfinalcopy

\usepackage[breaklinks=true,bookmarks=false]{hyperref}

\begin{document}

\title{Local Lesion Generation is Effective for Capsule Endoscopy Image Data Augmentation in a Limited Data Setting}

\author{Adrian B. Chłopowiec\textsuperscript{1,2,*,a}
\,
Adam R. Chłopowiec\textsuperscript{1,2,a}
\,
Krzysztof Galus\textsuperscript{1}
\\
Wojciech Cebula\textsuperscript{1}
\,
Martin Tabakov\textsuperscript{1,2}
}

\maketitle

\begin{center}
    \textsuperscript{1} Department of Artificial Intelligence, Wroclaw University of Science and Technology, Wroclaw, Poland \\
    \textsuperscript{2} BioCam LLC, Wroclaw, Poland \\
\end{center}

\renewcommand{\thefootnote}{\fnsymbol{footnote}}
\footnotetext[1]{ Corresponding author at: Department of Artificial Intelligence, Wroclaw University of Science and Technology, Wroclaw, Poland. \textit{Email address:} chlopowiec.adrian@gmail.com (Adrian Chłopowiec).}

\renewcommand{\thefootnote}{\alph{footnote}}
\footnotetext[1]{ The first two authors have contributed equally to this paper.}

\thispagestyle{empty}
\pagenumbering{arabic}
\pagestyle{plain}

\begin{abstract}
Limited medical imaging datasets challenge deep learning models by increasing risks of overfitting and reduced generalization, particularly in Generative Adversarial Networks (GANs), where discriminators may overfit, leading to training divergence. This constraint also impairs classification models trained on small datasets. Generative Data Augmentation (GDA) addresses this by expanding training datasets with synthetic data, although it requires training a generative model. We propose and evaluate two local lesion generation approaches to address the challenge of augmenting small medical image datasets. The first approach employs the Poisson Image Editing algorithm, a classical image processing technique, to create realistic image composites that outperform current state-of-the-art methods. The second approach introduces a novel generative method, leveraging a fine-tuned Image Inpainting GAN to synthesize realistic lesions within specified regions of real training images. A comprehensive comparison of the two proposed methods demonstrates that effective local lesion generation in a data-constrained setting allows for reaching new state-of-the-art results in capsule endoscopy lesion classification. Combination of our techniques achieves a macro F1-score of 33.07\%, surpassing the previous best result by 7.84 percentage points (p.p.) on the highly imbalanced Kvasir Capsule Dataset, a benchmark for capsule endoscopy.  To the best of our knowledge, this work is the first to apply a fine-tuned Image Inpainting GAN for GDA in medical imaging, demonstrating that an image-conditional GAN can be adapted effectively to limited datasets to generate high-quality examples, facilitating effective data augmentation. Additionally, we show that combining this GAN-based approach with classical image processing techniques further improves the results.
\end{abstract}

\section{Introduction}

The classification of medical data using machine learning models presents a significant challenge \cite{kim2022transfer,chlopowiec2023counteracting,bansal2022systematic} due to limited data availability \cite{chlopowiec2023counteracting,koh2021wilds} and class imbalance \cite{chlopowiec2023counteracting,zhang2020active,tu2024towards}, which often arises from natural disparities in pathology prevalence within a population. Gastrointestinal tract cancers, which account for 63\% of cancer-related deaths and cause 2.2 million fatalities annually, underscore the critical need for accurate diagnostic tools \cite{borgli2020hyperkvasir}.

Video capsule endoscopy (VCE) offers an alternative to traditional gastroscopy and colonoscopy, serving as a gold standard for diagnosing small bowel lesions \cite{enns2017clinical, costamagna2002prospective}. Unlike conventional endoscopy, VCE can visualize parts of the small intestine inaccessible through other methods, making it particularly useful for identifying occult bleeding and small mucosal lesions that standard imaging techniques might miss \cite{pennazio2015small, robertson2019repeat}. Capsule variants, such as PillCam Colon, also allow for large intestine examination \cite{adler2011pillcam}. Additionally, VCE is less invasive than traditional methods and demonstrates superior sensitivity in detecting small bowel lesions compared to push enteroscopy \cite{appleyard2000randomized, adler2011pillcam,enns2017clinical,costamagna2002prospective}. However, it does have limitations: VCE does not permit biopsy collection \cite{costamagna2002prospective}, and determining the exact location of a lesion remains challenging despite advancements in the field \cite{than2012review}.

A major challenge in VCE is the sheer volume of data. The capsule moves through the gastrointestinal (GI) tract via peristalsis, generating lengthy videos with over 50,000 frames \cite{yang2020future}. Manually analyzing such footage can take up to two hours \cite{costamagna2002prospective,yang2020future} of a healthcare professional time, which increases the likelihood of errors due to lapses in concentration or insufficient expertise \cite{zheng2012detection, chetcuti2021capsule, rondonotti2012can}. An effective machine learning-based lesion classification method could accelerate this process, leading to faster diagnoses, earlier cancer detection, reduced mortality rates, and greater capacity for VCE examinations \cite{smedsrud2021kvasir, yang2020future}.

Capsule endoscopy datasets are prone to issues commonly seen in medical imaging, such as data scarcity and class imbalance \cite{smedsrud2021kvasir, yokote2024small}. The Kvasir Capsule dataset \cite{smedsrud2021kvasir}, the largest publicly available dataset for capsule endoscopy, suffers from a lack of diverse lesion representations, with some pathological classes containing images from only two or three patients. Lesions in the GI tract typically present as fine-grained, localized changes in tissue, often appearing in only one or two frames per video, and exhibit significant variation in color, shape, and size \cite{yang2020future}. The arbitrary orientation of these images further complicates classification tasks, where even state-of-the-art methods achieve only a 25.23\% macro F1-score in multiclass classification on the official data split. These models predominantly excel at identifying normal samples but misclassify most pathological cases.

Over the past four decades, numerous methods have been developed to address issues related to low data availability and class imbalance. These include sampling techniques \cite{garcia2010exploring, khushi2021comparative, tomek1976two, chawla2002smote, dablain2022deepsmote}, transfer learning \cite{pan2009survey, tajbakhsh2016convolutional}, and data augmentation strategies \cite{wang2022anomaly,lee2023improving}, particularly Generative Data Augmentation (GDA) \cite{zheng2024toward,yamaguchi2020effective,chen2022generative, azizi2023synthetic, cheng2024breaking}. However, training effective models in a limited data regime remains complicated.

The most common use case for GDA is \textit{de novo} generation using deep generative models like GANs, VAEs, and diffusion models \cite{bissoto2021gan,bowles2018gan,trabucco2023effective}. These models learn a data distribution from a training set and generate synthetic labeled examples to augment the original dataset. Empirical \cite{yamaguchi2020effective,chen2022generative,azizi2023synthetic, cheng2024breaking} and theoretical \cite{zheng2024toward} evidence shows that GDA improves model generalization, especially when overfitting occurs.

Image inpainting \cite{criminisi2004region, barnes2009patchmatch, chan2001nontexture, liu2018image, suvorov2022resolution} and image composition \cite{niu2021making, perez2023poisson} have also been employed for data augmentation. Image inpainting, dominated by GANs \cite{suvorov2022resolution,zhao2021large,liu2018image,yu2019free,nazeri2019edgeconnect} and diffusion models \cite{lugmayr2022repaint,tao2024erasing}, modifies parts of an image by generating plausible content, whereas image composition combines elements from different images using techniques like Poisson Image Editing, known also as Poisson Blending \cite{perez2023poisson}. However, most inpainting research focuses on removing objects, which contrasts with our need to generate lesions. Both methods have shown promise in augmenting data for image recognition tasks \cite{wang2022anomaly, lee2023improving, he2024image, wang2024generative, tao2024erasing}.

In this work, we introduce two data augmentation methods — one based on image composition, the other on image inpainting — that modify parts of healthy tissue images to generate synthetic lesions. The first method employs Poisson Blending Data Augmentation (PBDA) to realistically combine healthy and pathological tissue, while the second, Image Inpainting Data Augmentation (IIDA), introduces a novel generative approach. IIDA fine-tunes an image inpainting GAN to insert lesions into healthy tissue. Both techniques leverage the abundance of healthy tissue images in medical datasets to produce high-quality synthetic data, effectively augmenting small medical imaging datasets. By addressing the challenges of GAN and classifier overfitting, these approaches enhance data quality and model performance. Notably, when synthetic data created with both techniques is combined to enrich the training dataset for the classifier, the model achieved a new state-of-the-art performance on the Kvasir Capsule Dataset, with a 33.07\% macro F1-score, surpassing the previous best result by 7.84 p.p.

The contributions of this paper can be summarised as follows:

\begin{itemize}
    \item We propose IIDA method, a novel data augmentation approach using a fine-tuned image inpainting model, particularly effective in low-data settings. This new approach leverages the abundance of healthy tissue, common in many medical imaging domains, to generate lesions.
    \item We designed PBDA method, a data augmentation pipeline adjusted to specific characteristics of VCE data. This pipeline uses image processing algorithms to generate high-quality synthetic lesions by modifying images of healthy tissue. Unlike IIDA, PBDA can be used with any dataset size.
    \item Our qualitative and quantitative results demonstrate the superiority of the generated synthetic samples in improving classification performance.
    \item We compare PBDA with IIDA for augmenting local pathological changes, showing that both techniques can enhance machine learning models. IIDA shows better performance than PBDA as a standalone data augmentation technique, however their combination provides the best results.
    \item We demonstrate that our local lesion generation methods outperform modern generative models, such as NVAE \cite{vahdat2020nvae} and LDM \cite{rombach2022high}, in data augmentation for classification tasks.

\end{itemize}

The rest of the paper is organized as follows: Section \ref{sec:related-works} reviews the related literature. Section \ref{sec:methods} introduces our two proposed data augmentation pipelines. In Section \ref{sec:experiments} we described the dataset and performed an extensive qualitative and quantitative study of proposed data augmentation techniques. Results and methods are discussed in Section \ref{sec:discussion} and conclusions are drawn in Section \ref{sec:conclusion}.

\section{Related works}
\label{sec:related-works}
\subsection{Data scarcity and data imbalance}
Training deep learning models from scratch typically requires extensive annotated datasets and prolonged training time to achieve acceptable performance, especially in comparison to leveraging pre-trained models. In image classification, pre-trained models, especially those trained on ImageNet, are widely adopted for their performance benefits \cite{deng2009imagenet,huh2016makes,ridnik2021imagenet,xie2018pre,kornblith2019better}. Despite notable differences between natural and medical images, Tajbakhsh et al. \cite{tajbakhsh2016convolutional} demonstrated that pre-trained CNNs perform comparably or even outperform CNNs trained from scratch on medical imaging tasks. They further highlighted that these CNNs exhibit greater robustness to variations in training set size and emphasized the importance of selecting an optimal fine-tuning strategy.

In the context of Generative Adversarial Networks (GANs), Karras et al. \cite{karras2020training} investigated training StyleGAN2 under limited data conditions, utilizing datasets of only 5,000, 2,000, and even 1,000 images. They introduced Adaptive Discriminator Augmentation (ADA) to mitigate discriminator overfitting, thus enhancing training stability. This approach dynamically adjusts the augmentation probability without leaking into the generator. Additionally, they validated a GAN fine-tuning technique as proposed by Mo et al. \cite{mo2020freeze}, underpinning their methods with proofs to support theoretical claims.

\subsection{Image generation in capsule endoscopy}
A growing body of research has focused on generating capsule endoscopy images using various methods. Vats et al. \cite{vats2023evaluating} trained a StyleGAN2 model for the unconditional \textit{de novo} generation of capsule endoscopy images, utilizing a dataset of 200,000 images. They conducted an expert evaluation to assess the quality of the generated samples.

Diamantis et al. \cite{diamantis2024intestine} introduced a VAE model trained on a subset of Kvasir Capsule and another VCE dataset. They trained distinct models for normal and abnormal image generation, achieving high-quality synthetic images. Xiao et al. \cite{xiao2023wce} proposed a GAN architecture to generate synthetic capsule endoscopy images from Kvasir Capsule data, which were then used in GDA for object detection. Unlike Xiao et al. \cite{xiao2023wce}, who utilized a random data split, our approach adopts the official data split, enhancing comparability with other studies and better simulating clinical environments.

\subsection{Image inpainting}
Image inpainting is the process of reconstructing missing or damaged parts of an image in a way that blends seamlessly with the original content. Applications include object removal \cite{criminisi2004region, barnes2009patchmatch, yu2019free, suvorov2022resolution}, artifact elimination \cite{daher2023temporal, elharrouss2020image}, restoration of old or damaged photographs \cite{elharrouss2020image, chan2001nontexture}, and enhancement in medical imaging \cite{armanious2020ipa, armanious2019adversarial, wang2021medical}.

Image inpainting is a non-trivial, ill-posed problem \cite{guillemot2013image,wang2022diverse}. The non-uniqueness of the problem arises because there are often multiple plausible ways to fill missing regions. This complexity is heightened when the missing areas cover substantial portions of the image, leaving minimal conditional information for inpainting models to rely on \cite{zhao2021large}.

Suvorov et al. \cite{suvorov2022resolution} utilized Fast Fourier Convolutions (\textit{FFC}) \cite{chi2020fast} to develop a class of Large Mask Inpainting (LaMa) models. They showed that a large receptive field is essential for high-quality inpainting of large masks. FFC significantly enhances LaMa's ability to recreate repetitive patterns. Additionally, Suvorov et al. \cite{suvorov2022resolution} proposed the High Resolution Field Perceptual Loss, leveraging a segmentation network to extract high-level features and enhance the model’s focus on object structure over texture.

\subsection{Generative data augmentation}
GANs and diffusion models have shown promise in medical image augmentation by synthesizing samples to expand training datasets. He et al. \cite{he2024image} developed a GAN specifically for image inpainting to generate synthetic pathological samples in histopathology. Their model, trained with global L1, perceptual, style, and local L1 losses, produced augmented data that improved segmentation performance for pathological tissues.
In another medical application, Lee et al. \cite{lee2024intraoperative} applied a GAN-based model, DeepFillv2, to reconstruct glandular regions that had been partially removed. The inpainting demonstrated superior effectiveness compared to geometric transformations and baseline methods. 

Zhang et al. \cite{zhang2021progressive} proposed a dual-network approach, where one network generates normal thyroid ultrasound images and the other focuses on generating nodules. For normal samples, random image regions were masked, whereas for nodule samples, only the nodule region was masked.

Beyond medical imaging, image inpainting-based data augmentation has also proven effective in other domains. Wang et al. \cite{wang2024generative} designed an inpainting GAN for multi-class object detection in infrared images. Their model incorporated reconstruction and perceptual losses alongside a novel multiscale erosion-based MSE loss, enhancing object detection accuracy.

Tao et al. \cite{tao2024erasing} proposed a Denoising Diffusion Probabilistic Model (DDPM)-based GDA method for surface defect inspection in a limited data regime. Synthetic samples are generated by conditioning the model on partially erased real defect samples.

\subsection{Image composition}
Image composition is a process of combining multiple images, elements of images, or specifically a foreground object and background image to create a realistic composite image.

It has been used in data augmentation to improve downstream task performance in medical imaging. Wang et al.\cite{wang2022anomaly} utilized Poisson Blending to augment diabetic retinopathy lesions in fundus images from retinography. Lee et al. \cite{lee2023improving} created an augmentation pipeline using Poisson Blending to augment training sets in classical endoscopy. Their solution involved the use of Grad-CAM \cite{selvaraju2017grad} for automatic segmentation of lesions in images. They use the lowest RGB variance sliding window search to find a suitable area for blending. 

\subsection{Classification in capsule endoscopy}

The Kvasir Capsule Dataset \cite{smedsrud2021kvasir} is currently the largest publicly available dataset of capsule endoscopy videos. Smedsrud et al. \cite{smedsrud2021kvasir} introduced the official data split, where no patient images are shared between different splits. They established baseline performance for this dataset using ResNet-152 \cite{he2016deep} and DenseNet-161 \cite{huang2017densely}. Recent work \cite{srivastava2022video} validated the effectiveness of their FocalConvNet, alongside several modern CNNs and Vision Transformers (ViTs) also using that split.

In contrast, many works \cite{malik2024multi, alam2022rat, oukdach2023conv, oukdach2022gastrointestinal} evaluate their methods using a random data partitioning, where sequential images from videos can be included in both the training and test sets. Other studies \cite{fonseca2022abnormality} do not specify their data breakdown methodology, while some \cite{bordbar2023wireless, lima2022classification} do not use the official split. Our work distinguishes itself by using data split introduced by Smedsrud et al. \cite{smedsrud2021kvasir}, which increases the reliability of comparisons and ensures the reproducibility of the experiments.

\section{Methods}
\label{sec:methods}
\begin{figure}
    \centering
    \includegraphics[width=\linewidth]{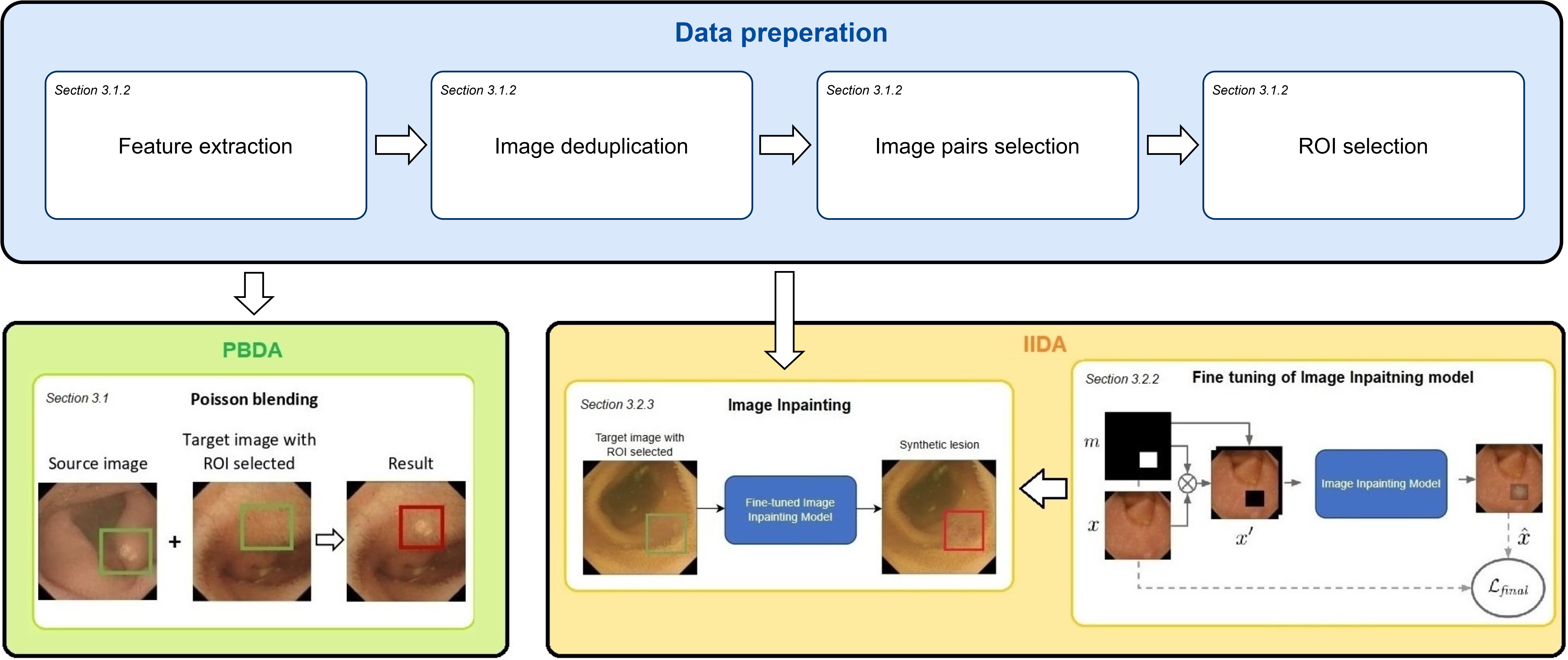}
    \caption{Diagram of the proposed methods. The data preparation step is shared for PBDA and IIDA. This stage increases data variability and quality by addressing the specific characteristics of VCE data. Image pairs and ROIs obtained in this stage are subsequently used for further processing. PBDA uses image pairs and ROIs to create synthetic lesions through Poisson Blending. IIDA fine-tunes image inpainting models for each lesion class, which are then used to inpaint lesions in locations selected through the data preparation stage.}
    \label{fig:figure1}
\end{figure}

This section introduces the PBDA and IIDA methods developed in this study, with a clear presentation of each method’s pipeline, background, and data preparation steps. A visual summary of the methods is provided in Figure \ref{fig:figure1}.

In Section \ref{sec:poisson_blending_data_augmentation}, we outline the PBDA pipeline, beginning with background information on the Poisson Blending (Section \ref{sec:poisson_background}) and detailing the data preparation steps (Section \ref{sec:methods_data_preparation}). These data preparation techniques improve data augmentation outcomes and are used consistently with both PBDA and IIDA to enable a direct comparison between the two approaches.

Section \ref{sec:methods_image_inpainting} then presents the IIDA, beginning with background information on LaMa (Section \ref{sec:lama_background}), followed by the finetuning process (Section \ref{sec:methods_fine_tuning}), and concluding with the data generation process (Section \ref{sec:methods_data_generation}). This structured approach ensures a coherent understanding of PBDA and IIDA, emphasizing the methods' individual contributions and shared comparative framework.

\subsection{Poisson Blending Data Augmentation}
\label{sec:poisson_blending_data_augmentation}

To optimize data augmentation for VCE images, we developed a Poisson Blending Data Augmentation pipeline that increases data diversity while addressing high redundancy of VCE data. The method aims to create augmented samples that appear natural, while introducing variability. Our pipeline incorporates the following stages:

\begin{itemize}
    \item Feature extraction.
    \item Image deduplication.
    \item Image pairs selection.
    \item Region of interest (ROI) selection.
    \item Poisson Blending.
\end{itemize}

The method is illustrated in Figure \ref{fig:figure1} and each of these steps will be explained in detail in the following sections. While Poisson Blending has been applied in recent studies as a data augmentation technique \cite{lee2023improving, wang2022anomaly}, our proposed pipeline is novel and specifically tailored for VCE data, although it can be adapted to other medical imaging fields.

To support understanding of our methodology, the following section provides background on Poisson Blending, laying the groundwork for the data preparation procedure described in Section \ref{sec:methods_data_preparation}.

\subsubsection{Background on Poisson Blending}
\label{sec:poisson_background}
There are various algorithms for merging two images, with Poisson Blending \cite{perez2023poisson} being one of the most widely known and frequently used methods \cite{wang2022anomaly,lee2023improving}. In our work, we applied this gradient-based image processing technique to augment medical data.

The fundamental idea is to blend a selected region from a source image into a specified area of a target image, aiming to minimize color discrepancies at the blending boundary. The goal is to ensure that the gradients within the blended region closely resemble those of the source image, while the boundary seamlessly matches the target image. This approach is particularly useful for augmenting medical data by blending lesion areas into healthy tissue, thereby expanding the lesion dataset. Below, we provide a brief theoretical overview of Poisson Blending, following the original notation from \cite{perez2023poisson}.

\begin{figure}[h]
    \centering
    \includegraphics[width=0.8\linewidth]{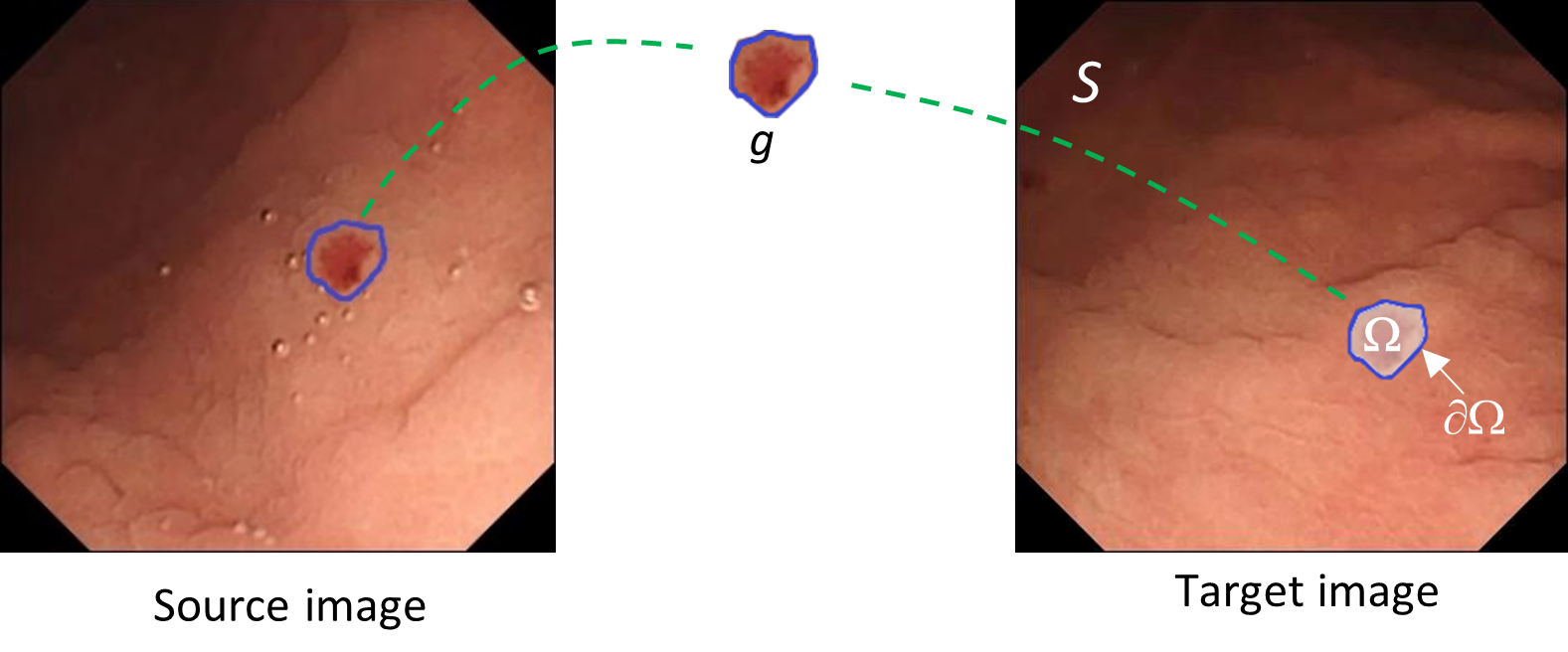}
    \caption{Interpretation of the introduced notation. The image definition domain $S$ represents the target image, with $\Omega$ denoting the blending location, where $\partial\Omega$ is the border of the location. The function $f^*$ is defined over $S$ minus the interior of $\Omega$ and represents e.g. the background of a target image. The function $g$ can be a lesion and the unknown $f$ defined over $\Omega$ represents the blended pathology. }
    \label{fig:figure2}
\end{figure}

Let  $S$, be the image definition domain, defined as a closed subset of  $\mathbb{R}^2$, and let  $\Omega$  be a closed subset of $S$ with boundary $\partial\Omega$. Let $f^*$ be a known scalar function defined over $S$ minus the interior of $\Omega$ and let $f$ be an unknown scalar function defined over the interior of  $\Omega$ and finally, $\mathbf{v}$ be a vector field defined over $\Omega$. Essentially, the function $f$ interpolates the destination function $f^*$ over $\Omega$, under the guidance of the vector field $\mathbf{v}$, which can be the gradient field of some source function $g$.

The interpolant $f$ of $f^*$ is defined as the solution of the following minimalization problem:
\begin{equation}
    \min_f\iint_{\Omega}{|\nabla f - \mathbf{v}|^2} \ \text{with} \ f|_{\partial\Omega}=f^*|_{\partial\Omega},
    \label{eq:interpolant}
\end{equation}
where $\nabla$ is the gradient operator. The solution of Eq. \ref{eq:interpolant} must satisfy the following Poisson equation with Dirichlet boundary conditions:

\begin{equation}
    \Delta f = \text{div}\ \mathbf{v} \ \text{over} \ \Omega,\ \text{with} \ f|_{\partial\Omega}=f^*|_{\partial\Omega},
    \label{eq:poisson-equation}
\end{equation}
where $\Delta.=\frac{\partial^{2}.}{\partial x^2} + \frac{\partial^{2}.}{\partial y^2}$ is the Laplacian operator and div $\mathbf{v}$ is the divergence of $\mathbf{v}$.

To provide an intuitive understanding, Figure \ref{fig:figure2} illustrates the relationship between real input image data and the concepts described above. In this example, the function $f^*$ could represent the background of a target image, $g$ could depict a lesion, and $f$ would correspond to the blended pathology. However, it's important to note that this is a simplified illustration for intuition, as digital images require consideration of the discrete case.

In our implementation, we selected the gradient field from the source lesion $g$ as the guidance field $\mathbf{v}$, as this is a fundamental choice.

Therefore, the interpolation is performed under the guidance of:
\begin{equation}
    \mathbf{v} = \nabla g,
\end{equation}
and Eq. \ref{eq:poisson-equation} takes the form:
\begin{equation}
    \Delta f = \Delta g \ \text{over} \ \Omega, \ \text{with} \ f |_{\partial\Omega}=f^*|_{\partial\Omega}.
\end{equation}
The above assumption introduces the so-called seamless cloning approach, which ensures the compliance of source and destination boundaries.

\subsubsection{Data preparation}
\label{sec:methods_data_preparation}
\begin{figure}
    \centering
    \includegraphics[width=0.8\linewidth]{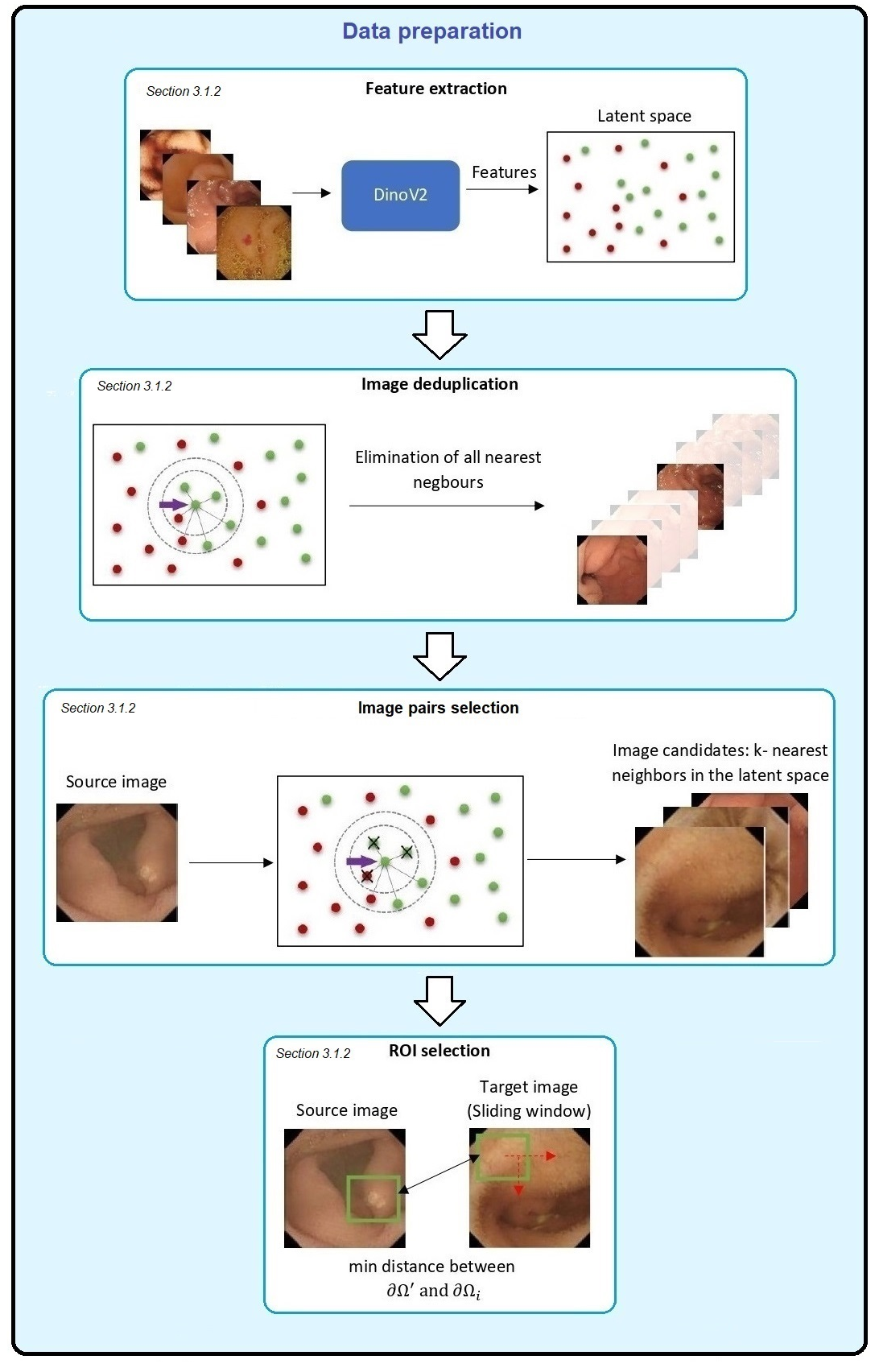}
    \caption{The data preparation diagram. Latent representations of images are extracted using DinoV2-Giant model \cite{oquab2023dinov2}. Deduplication stage increases data variability and quality by removing redundant images. Following this, image pairs and ROIs are selected for further processing.}
    \label{fig:figure3}
\end{figure}

Our data preparation procedure (Figure \ref{fig:figure3}) utilizes specific characteristics of VCE data and the Poisson Blending technique to maximize the data augmentation effectiveness. It consists of feature extraction, image deduplication, and image pairs and ROI selection steps, which are further described in detail.\\

\textit{Feature extraction and image deduplication}. A common challenge with VCE data is the presence of numerous seemingly duplicate images. This occurs when the capsule remains stationary in the intestine for extended periods, often due to slow or infrequent peristaltic movements. Removing these redundant images increases dataset diversity, which enhances the variety of pathological and healthy tissue pairings in the proposed pipeline, ultimately leading to a more diverse synthetic dataset.

The deduplication process, detailed in Appendix \ref{appendix:a}, begins by extracting latent vectors from all images in the training dataset using the DinoV2-Giant model \cite{oquab2023dinov2}. The dataset is then filtered iteratively, removing images that are most similar to the one currently sampled. Similarity is measured by the Euclidean distance between latent vectors, with a threshold determined experimentally.\\

\textit{Image pairs selection}. To augment data through Poisson Blending, firstly appropriate ROIs need to be found for lesion blending. To explain our approach to selecting an optimal blending location, we first introduce new definitions. Similar to Section \ref{sec:poisson_background}, let $S^{\prime}$ represent an image domain, defined as a closed subset of  $\mathbb{R}^2$ and let  $\Omega^{\prime}$  be a closed subset of $S^{\prime}$, where the previously introduced function $g$ is defined over $\Omega^{\prime}$. Additionally, a correction function $\tilde{f}$, defined on $\Omega$, is such that $f = g + \tilde{f}$  \cite{perez2023poisson}, which will help in understanding Poisson interpolation.

The Poisson Blending algorithm for seamless cloning is then based on solving the following Laplace equation:
\begin{equation}
     \Delta \tilde{f}=0 \text{ over } \Omega
    \label{eq:laplace-equation}
\end{equation}
with boundary conditions \cite{perez2023poisson}:
\begin{equation}
\tilde{f}|_{\partial\Omega}=(f^{*}-g)|_{\partial\Omega}.
    \label{eq:boundary-condition}
\end{equation}
Eq. \ref{eq:laplace-equation} and \ref{eq:boundary-condition} show that inside $\Omega$ the $\tilde{f}$ is a membrane interpolant of the mismatch ($f^* - g$) along the boundary $\partial\Omega$.

The function $\tilde{f}$ represents the adjustments needed for the lesion to blend seamlessly using the Poisson Blending algorithm. Therefore for the lesion to remain consistent after the blending, the difference $f^*-g$ must be minimal along the boundary $\partial\Omega$. If this difference is too large, it can cause color bleeding or blending artifacts \cite{afifi2015mpb}, potentially altering the semantics (or class) of $g$ and introducing label noise into the synthesized dataset. Our pipeline for finding pairs of $f^*$ and $g$ values consists of two steps: first, finding the most similar pair of images defined over $S$ and $S^{\prime}$, and then identifying the optimal location $\Omega$ for seamless cloning.

Identifying the most similar pair of source and target images relies on semantic search within the latent space of the DinoV2-Giant model \cite{oquab2023dinov2}. The process is detailed in Appendix \ref{appendix:a}. To identify $K$ target pairs for a given source, simply $K$ most similar images of healthy tissue are chosen for a given lesion. After selecting such pairs, choosing an appropriate blending location becomes essential. Not all areas in VCE images are suitable for seamlessly blending lesions. Regions like edges, folds, or holes are generally unsuitable because the source lesion cannot appear there without significant alteration. The Poisson Blending algorithm integrates the lesion as is, adjusting only the color for seamless blending but not modifying its texture or shape. Furthermore, if there is a large gradient difference at the boundary of the lesion and the new region (e.g., near folds or dark zones), blending artifacts may arise.\\

\textit{ROI selection}. To blend a pathological subregion from a source image into a target image, it is necessary to identify a ROI in the target image that minimizes the color difference to the ROI in the source image along the border pixels. This process is illustrated in Figure \ref{fig:figure3}.

Formally, let $\partial\Omega^{\prime}$ denote the set of border pixels for the lesion's bounding box in the source image, and $\partial\Omega_i$ represent the border pixels for the bounding boxes of each potential ROI in the target image, indexed by $i = 1, \dots, N$, where $N$ is the total number of ROIs generated using a sliding window algorithm. For all $i$, the number of border pixels is denoted as $|\partial\Omega^{\prime}| = |\partial\Omega_i| = M$.

The optimal ROI in the target image is determined by minimizing the average color difference between the corresponding border pixels of the source image and the respective ROIs. This is expressed as:
\begin{equation}
    \min_i{\{\frac{1}{M}\sum\limits_{j=1}^{M}{\sqrt{(p_{j}^{\prime} - p_{j})^2}}: p^{\prime}\in\partial\Omega^{\prime}, p\in\partial\Omega_{i}, i=1,\dots,N\}},
\end{equation}
where $p_j^{\prime}$ are the border pixels of ROI in the source image, and $p_j$ are the corresponding border pixels in the target image for each ROI $i$.\\ 

\textit{Poisson blending.} After completing the previous steps - selecting the appropriate source and target images and determining the suitable blending location - Poisson Blending can be performed as described in \cite{perez2023poisson} and in Section \ref{sec:poisson_background}.

\subsection{Image Inpainting Data Augmentation}
\label{sec:methods_image_inpainting}

The IIDA method leverages the unique strength of image inpainting — its capacity to fill missing image regions in a visually convincing manner — to generate realistic lesions in images of healthy tissue. By harnessing this feature, IIDA enhances synthetic sample generation and enables effective data augmentation. The approach involves training an image inpainting model to reconstruct masked-out lesions, conditioned on surrounding tissue. This conditioning directs the model to sample lesions that are better aligned with the geometry, brightness, contrast, texture, and shape of the background tissue. 

\begin{figure}[ht]
    \centering
    \includegraphics[width=0.8\linewidth]{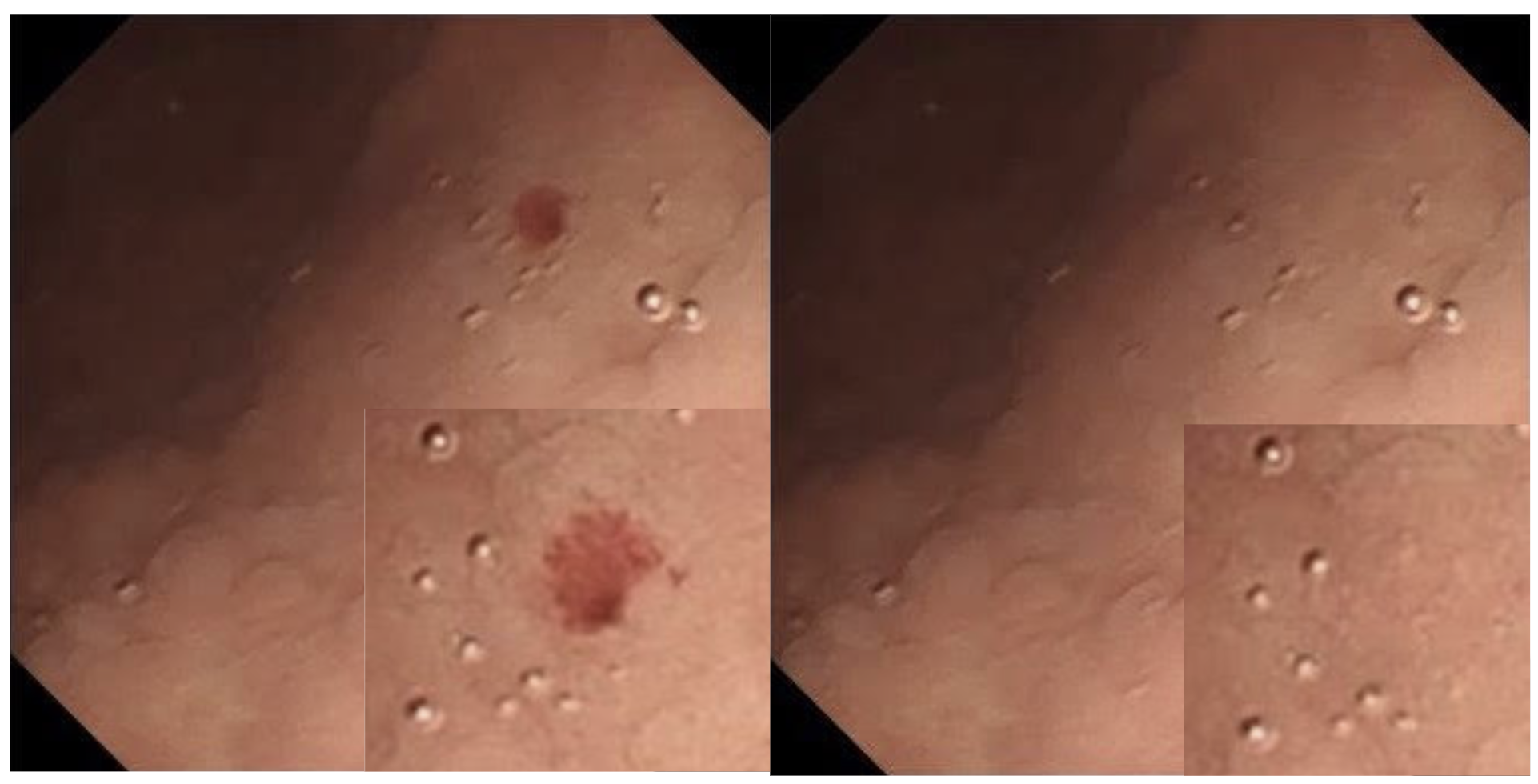}
    \caption{LaMa inpainting on a Kvasir Capsule image. On the left is an angiectasia sample from the Kvasir Capsule Dataset, and on the right is a sample generated by the pre-trained Big Lama-Fourier model \protect\cite{suvorov2022resolution} without fine-tuning. While this model can generate realistic tissue in capsule endoscopy images and effectively remove lesions, IIDA requires the opposite functionality. The generated content integrates seamlessly with the original tissue, showing no visible border.}
    \label{fig:figure4}
\end{figure}

IIDA requires selecting a backbone model for image inpainting; in this study, LaMa \cite{suvorov2022resolution} was chosen, as it demonstrated the ability to generate realistic capsule endoscopy images in a zero-shot setting (Figure \ref{fig:figure4}). It was fine-tuned on a subset of the Kvasir Capsule Dataset \cite{smedsrud2021kvasir} to achieve state-of-the-art data augmentation. 

To the best of our knowledge, this is the first method to use a pre-trained Image Inpainting model, fine-tuned to generate new objects for data augmentation in classification tasks on a limited medical imaging dataset.

The following sections will provide background information on the LaMa model, the model’s fine-tuning, and the data generation process.

\subsubsection{Background on LaMa}
\label{sec:lama_background}
Following notation introduced by Suvorov et al. \cite{suvorov2022resolution}, the mask is represented as  $m$, and the masked image is $x \odot m$. The LaMa model takes as input a stack of the masked image and the mask, denoted  $x^{\prime} = stack(x \odot m, m)$. The inpainting network, referred to as the generator or synthesis network, is denoted by $f_\theta(\cdot)$, where $\theta$ are the parameters of the network. The LaMa model processes $x^{\prime}$ using a fully convolutional approach, yielding a three-channel RGB image $\hat{x} = f_\theta(x^{\prime})$. In this study, bounding boxes provided for lesions in the Kvasir Capsule Dataset \cite{suvorov2022resolution} are used to prepare image-mask pairs.

Suvorov et al. \cite{suvorov2022resolution} introduced High Receptive Field Perceptual Loss (HRFPL). It is based on perceptual loss from Johnson et al. \cite{johnson2016perceptual}, which computes the distance between feature vectors extracted from a pretrained network $\phi(\cdot)$ applied to generated and original images. The formula for HRFPL loss, as defined by Suvorov et al. \cite{suvorov2022resolution}, is provided below:
\begin{equation}
\label{eq:suvorov2022resolution_eq_1}
\mathcal{L}_{HRFPL}(x, \hat{x}) = \mathcal{M}([\phi_{HRF}(x) - \phi_{HRF}(\hat{x})]^2),
\end{equation}
where $[\cdot - \cdot]^2$ is an element-wise operation, and $\mathcal{M}$ is the sequential two-stage mean operation (interlayer mean of intra-layer means).

LaMa uses adversarial loss based on image patches. Only patches that intersect with the masked region of interest are considered \textit{fake}. The non-saturating adversarial loss is defined by Suvorov et al. \cite{suvorov2022resolution} as:
\begin{equation}
\begin{split}
\label{eq:suvorov2022resolution_eq_2}
\mathcal{L}_D = &-\mathbb{E}_x\left[logD_\xi(x)\right] - \mathbb{E}_{x,m}\left[logD_\xi(\hat{x}) \odot m\right] \\
&- \mathbb{E}_{x,m}\left[log(1-D_\xi(\hat{x})) \odot (1 - m)\right]
\end{split}
\end{equation}

\begin{equation}
\label{eq:suvorov2022resolution_eq_3}
\mathcal{L}_G = -\mathbb{E}_{x,m}\left[logD_\xi(\hat{x})\right]
\end{equation}

\begin{equation}
\label{eq:suvorov2022resolution_eq_4}
L_{Adv} = sg_\theta (\mathcal{L}_D) + sg_\xi (\mathcal{L}_G) \rightarrow \min_{\theta,\xi},
\end{equation}
where $sg_{var}$ stops gradients w.r.t $var$ and $L_{Adv}$ is the joint loss to optimize.

Suvorov et al. \cite{suvorov2022resolution} also use two types of regularization on the discriminator network. The first one is $R_1=E_x\Vert \nabla D_\xi (x) \Vert^2$ gradient penalty \cite{mescheder2018training}, the second one is a discriminator-based perceptual loss $\mathcal{L}_{DiscPL}$ \cite{wang2018high}, also called feature matching loss, applied on features of a discriminator network. Their final loss function is:
\begin{equation}
\label{eq:suvorov2022resolution_eq_5}
\mathcal{L}_{final} = \kappa L_{Adv} + \alpha \mathcal{L}_{HRFPL} + \beta \mathcal{L}_{DiscPL} + \gamma R_1
\end{equation}
where $\kappa, \alpha, \beta, \gamma$ are weighting parameters for each part of loss function. Suvorov et al. \cite{suvorov2022resolution} used values of $\kappa = 10$, $\alpha = 30$, $\beta = 100$ and $\gamma = 0.001$.

\subsubsection{Fine-tuning}
\label{sec:methods_fine_tuning}
Multiple studies have shown that GANs fine-tuning is possible \cite{mo2020freeze,noguchi2019image,wang2020minegan,wang2018transferring}. LaMa utilizes training-time data augmentation for discriminator, which has been shown by Karras et al. \cite{karras2020training} to be effective when training on a limited dataset. The primary novelty of this paper is a modification of the standard training procedure of image inpainting models to fine-tuning on the task of lesion reconstruction.

Therefore in this work, the Big LaMa-Fourier model trained by Suvorov et al. \cite{suvorov2022resolution} on Places2 Dataset has been fine-tuned using a modified loss function:
\begin{equation}
\label{eq:lama_modified_loss}
\mathcal{L}_{final} = \kappa L_{Adv} + \alpha \mathcal{L}_{HRFPL} + \beta \mathcal{L}_{DiscPL} + \gamma R_1 + \boldsymbol{\delta} {\mathcal{L}_{L1}},
\end{equation}
where $\mathcal{L}_{L_1}(x, \hat{x})=||x - \hat{x}||_1$. Adding the L1 term between the reconstruction of lesion generated by the model, and the original pathology, with $\delta = 10$, makes the reconstruction task easier for the model, while other parts of the loss function, such as $\mathcal{L}_{HRFPL}$ help to leverage the problem of blurriness when applying L1 loss term. This term is introduced to ensure that the model does not deviate from what it learned prior to the fine-tuning process. The rest of the scaling factors, are set with the values used by Suvorov et al. \cite{suvorov2022resolution}. The LaMa model, as presented by Suvorov et al. \cite{suvorov2022resolution}, does not allow for class conditioning and the development of such methods is out of the scope of this paper. Therefore, a separate model is fine-tuned for each of the considered lesion classes. In the following paragraphs, the fine-tuning process of separate models for each class will be described.

Formally, let $\mathbb{X} \subset \mathbb{E}^n$ be a subset of Euclidean space corresponding to image dimensions, representing capsule endoscopy images, and $\mathbb{Y}=\{0, 1, \dots, M\}$ be the set of corresponding labels, where $0$ represents the healthy tissue and $1, 2, ..., M$ represent different lesion classes. Let $\mathcal{D}=\{ (x^{\prime}_i, y_i) \}_{i=1}^{N}$ be a dataset of $N$ observations, where $y_i\in \mathbb{Y}$, $x^{\prime}_i\in \mathbb{X}$  and $x^{\prime}_i = stack(x_i \odot m_i, m_i)$, such that an observation $x_i$ is coupled with a mask $m_i$ derived from a bounding box for a lesion or obtained through ROI selection procedure (Section \ref{sec:methods_data_preparation}) for healthy tissue.

The dataset $\mathcal{D}$ can be divided into separate sets for each class. Let $\mathcal{D}_i=\{(x^{\prime}, y) \in \mathcal{D}: y=i \}$ for $i=0, 1, ..., M$. The learning problem is to fine-tune a set of inpainting networks $\mathcal{F}=\{f_{\theta_i}\}_{i=1}^{M}$ from a pre-trained $f_{\theta}$ network using respective datasets $\mathcal{D}_i$ for $ i=1, ..., M$ and the loss function $\mathcal{L}_{final}$. Note that the dataset $\mathcal{D}_0$ of healthy images is not used for training. 

\subsubsection{Data generation process}
\label{sec:methods_data_generation}
Following the formulations introduced in the previous section, a general approach to generating synthetic data using image inpainting is described in the following paragraph.

To augment the dataset, a set of $K$ new observations is generated for each lesion class, using respective $f_{\theta_i} \in \mathcal{F}$. For every lesion class $i \in \{1, ..., M\}$ we sample $K$ observations from $\mathcal{D}_0$ and place them in respective sets $A_i$. Let $\tilde{\mathcal{D}}_i=\{(f_{\theta_i}(x^{\prime}), i): x^{\prime}\in \mathcal{A}_i\}$ for $i=1,...,M$ be the sets of new observations generated with respective inpainting networks $f_{\theta_i}$. A dataset used for the supervised training of a classifier is therefore composed of the real and generated data: $\mathcal{D}_{class}=\{\mathcal{D}_0, \mathcal{D}_1, \tilde{\mathcal{D}}_1, ..., \mathcal{D}_M, \tilde{\mathcal{D}}_M \}$. 

However, in this study, to ensure a fair comparison with PBDA, we used samples and ROIs obtained for PBDA using the data preparation procedure outlined in Section \ref{sec:methods_data_preparation}. This approach serves as a specific implementation of the general data sampling procedure previously described.

Therefore, the datasets $\mathcal{D}_i$ for $i=0, 1, ..., M$ were constructed from the official training split of the Kvasir Capsule Dataset. For each dataset $\mathcal{D}_i$, $i=1,...,M$, the inpainting model was fine-tuned for 500 epochs, incorporating a linear learning rate warm-up over the first 10 epochs. The Adam optimizer was used for both the generator and discriminator, with learning rates set at 0.001 and 0.0001, respectively, and a batch size of 16. Although the learning rate and optimizer settings align with those in \cite{suvorov2022resolution}, the batch size was reduced due to technical constraints. Models were trained on images with a resolution of 256x256, consistent with \cite{suvorov2022resolution}. The linear learning rate warm-up is a standard practice in fine-tuning deep learning models.

\section{Experiments}
\label{sec:experiments}
\subsection{Dataset}
\label{sec:methods_dataset}
The largest public dataset on capsule endoscopy, the Kvasir Capsule \cite{smedsrud2021kvasir}, served as the basis for our experiments. The dataset contains 47,238 labelled images and 117 videos making a total of 4,741,504 images, obtained from the Olympus Endocapsule 10 System \cite{olympus2013}. In this work, only the labelled subset and official classification data splits are used. The dataset consists of 14 categories. Three of them - Blood-Hematin, Polyp, and Ampulla of Vater contain very few images, thus they are not considered in the experiments and following tables as in \cite{smedsrud2021kvasir} and \cite{srivastava2022video}. Pylorus, Ileocecal Valve, and Normal Clean Mucosa classes are anatomical, non-pathological findings, therefore are put into one class - normal tissue. Classes are organized according to the World Endoscopy Association Minimal Standard Terminology version 3.0 (MST 3.0), without subcategories or intermediate level \cite{aabakken2014standardized}. Despite the availability of sequential information in the form of videos, the dataset remains quite small and each image is treated individually in this work.

\begin{table}[h]
    \centering
    \begin{tabular}{l|r}
        \toprule
        \textbf{Category} & \textbf{Count} \\
        \midrule
        Normal & 40074 \\
        Unclear View & 2906 \\
        Angiectasia & 866 \\
        Ulcer & 854 \\
        Foreign Body & 776 \\
        Lymphangiectasia & 592 \\
        Erosion & 523 \\
        Blood Fresh & 446 \\
        Erythema & 159 \\
        \bottomrule
    \end{tabular}
    \caption{Class distribution in the Kvasir Capsule Dataset. The sample sizes across classes can differ by several orders of magnitude.}
    \label{tab:kvasir_counts}
\end{table}

The dataset is highly imbalanced, as shown in Table \ref{tab:kvasir_counts}, which poses a significant challenge for training the classification model \cite{chlopowiec2023counteracting, johnson2019survey}. In addition, many lesions are very small and localized, covering only a minor part of the image. Classification of this data is therefore very challenging. Smedsrud et al \cite{smedsrud2021kvasir} achieved just a 25.23\% F1-Score macro using DenseNet-161, the best result to date, while Srivastava et al. \cite{srivastava2022video} obtained a 21.78\% F1-Score macro with their best model, despite conducting an extensive study of modern classification models on the dataset.

\subsubsection{Data split}
\label{sec:methods_patientwise_split}
Images in the Kvasir Capsule Dataset are extracted from videos, therefore random splitting may result in similar images, such as consecutive frames, being included in both the training and test sets. This could lead to data leak and bias in the test results. To ensure a fair evaluation and simulate a clinical setting, all frames from a single video are placed in distinct groups, keeping them together in the same splits - as proposed by authors of Kvasir Capsule Dataset \cite{smedsrud2021kvasir}. This simulates real-world performance, where algorithms are applied to images from incoming patients. The official data split, presented in Table \ref{tab:combined_kvasir_splits}, represents the described approach. Such data split further increased the class imbalance of the dataset, as images from different classes are unevenly distributed between videos. In this work, we use \textit{split 1} for training and \textit{split 0} for testing, following the approach from \cite{srivastava2022video}.

Additionally, Smedsrud et al. \cite{smedsrud2021kvasir} provided bounding boxes for almost all labeled pathological findings (Table \ref{tab:combined_kvasir_splits}). We use these bounding boxes to extract lesions for both PBDA and IIDA pipelines.

\clearpage

\begin{table}[htbp]
    \begin{center}
        \begin{subtable}{.49\linewidth}
            \centering
            \caption{Kvasir image count.}
            {\small{
                \begin{tabular}{l|r|r}
            \toprule
            \textbf{Category} & \textbf{Split 0} & \textbf{Split 1} \\
            \midrule
            Normal & \textbf{20488} & 19586 \\
            Unclear View & \textbf{1787} & 1119 \\
            Ulcer & \textbf{582} & 272 \\
            Blood Fresh & \textbf{424} & 22 \\
            Lymphangiectasia & \textbf{368} & 224 \\
            Foreign Body & 186 & \textbf{590} \\
            Erosion & 178 & \textbf{345} \\
            Angiectasia & 95 & \textbf{771} \\
            Erythema & 27 & \textbf{132} \\
            \bottomrule
            \end{tabular}
            }}
        \end{subtable}
        \begin{subtable}{.49\linewidth}
            \centering
            \caption{Kvasir bounding boxes count.}
            {\small{
                \begin{tabular}{l|r|r}
            \toprule
            \textbf{Category} & \textbf{Split 0} & \textbf{Split 1} \\
            \midrule
            Normal & 0 & 0 \\
            Unclear View & 0 & 0 \\
            Ulcer & \textbf{582} & 272 \\
            Blood Fresh & \textbf{424} & 22 \\
            Lymphangiectasia & \textbf{368} & 224 \\
            Foreign Body & 186 & \textbf{590} \\
            Erosion & 162 & \textbf{345} \\
            Angiectasia & 95 & \textbf{771} \\
            Erythema & 27 & \textbf{90} \\
            \bottomrule
        \end{tabular}
            }}
    \end{subtable}
    \end{center}
    \caption{a) Kvasir Capsule Dataset official splits. All the images from a single patient are confined to one split to ensure patient-level separation. This approach prevents any image overlap across splits. Given that individual patients contribute differing numbers of lesion images — and since multiple lesions can be present within a single patient — this splitting method increases class imbalance within the dataset. b) Bounding boxes count of the official splits of the Kvasir Capsule Dataset. The dataset offers localized annotations for almost all lesions.}
    \label{tab:combined_kvasir_splits}
\end{table}

\subsection{PBDA - qualitative results}
\label{sec:experiments_poisson_blending}
\begin{figure}[htpb]
    \centering
    \includegraphics[width=\linewidth]{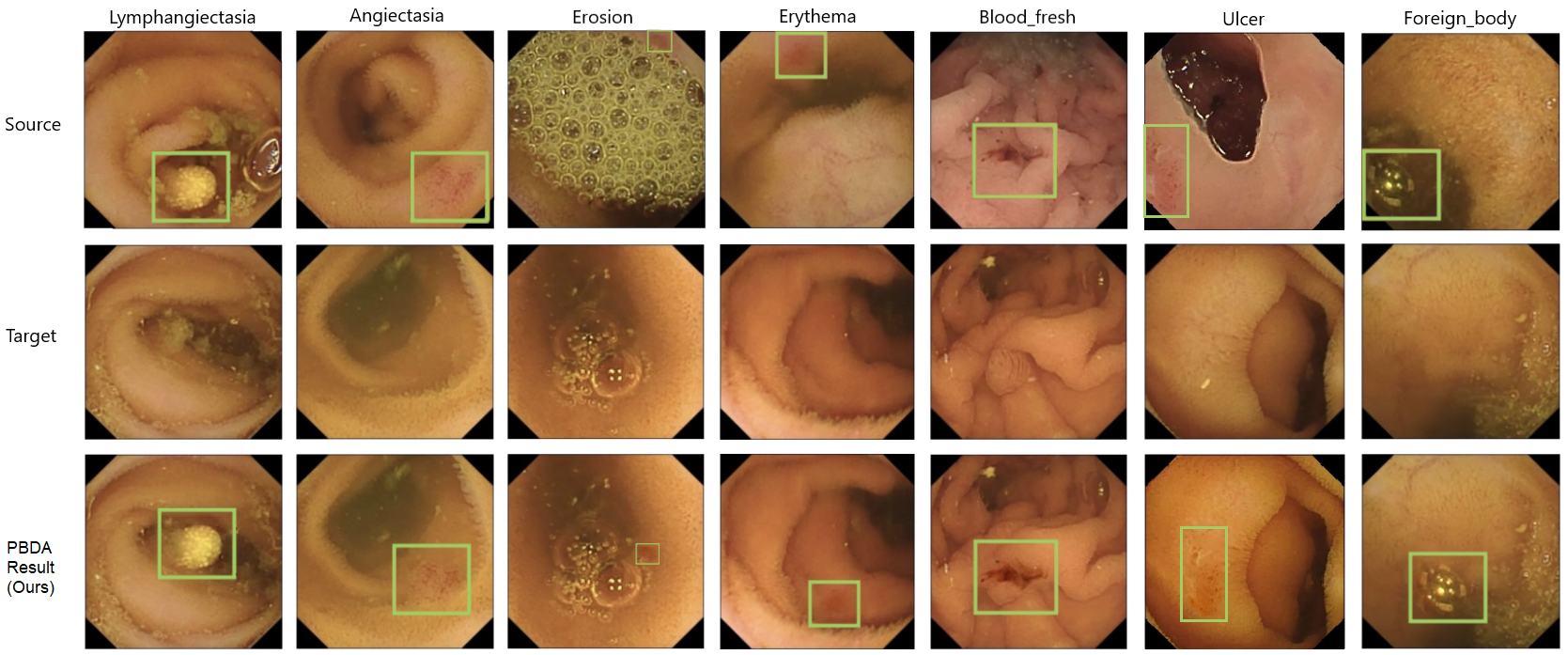}
    \caption{Samples generated using the PBDA pipeline, with the resulting lesions highlighted by a green bounding box. The PBDA pipeline demonstrates the ability to generate plausible lesions.}
    \label{fig:figure5}
\end{figure}

Figure \ref{fig:figure5} presents example synthetic samples generated by the PBDA pipeline, alongside their real counterparts. For capsule endoscopy data, PBDA effectively creates visually convincing lesions, closely replicating the color, texture, and brightness of the originals. It excels at blending high-contrast lesions that stand out from the background, such as ulcers, lymphangiectasia, angiectasia, and small fresh blood samples. The pipeline performs best when handling localized lesions — those confined to small, compact areas of the image — compared to more diffuse conditions, like extensive bleeding.

\begin{figure}[htpb]
    \centering
    \includegraphics[width=0.5\linewidth]{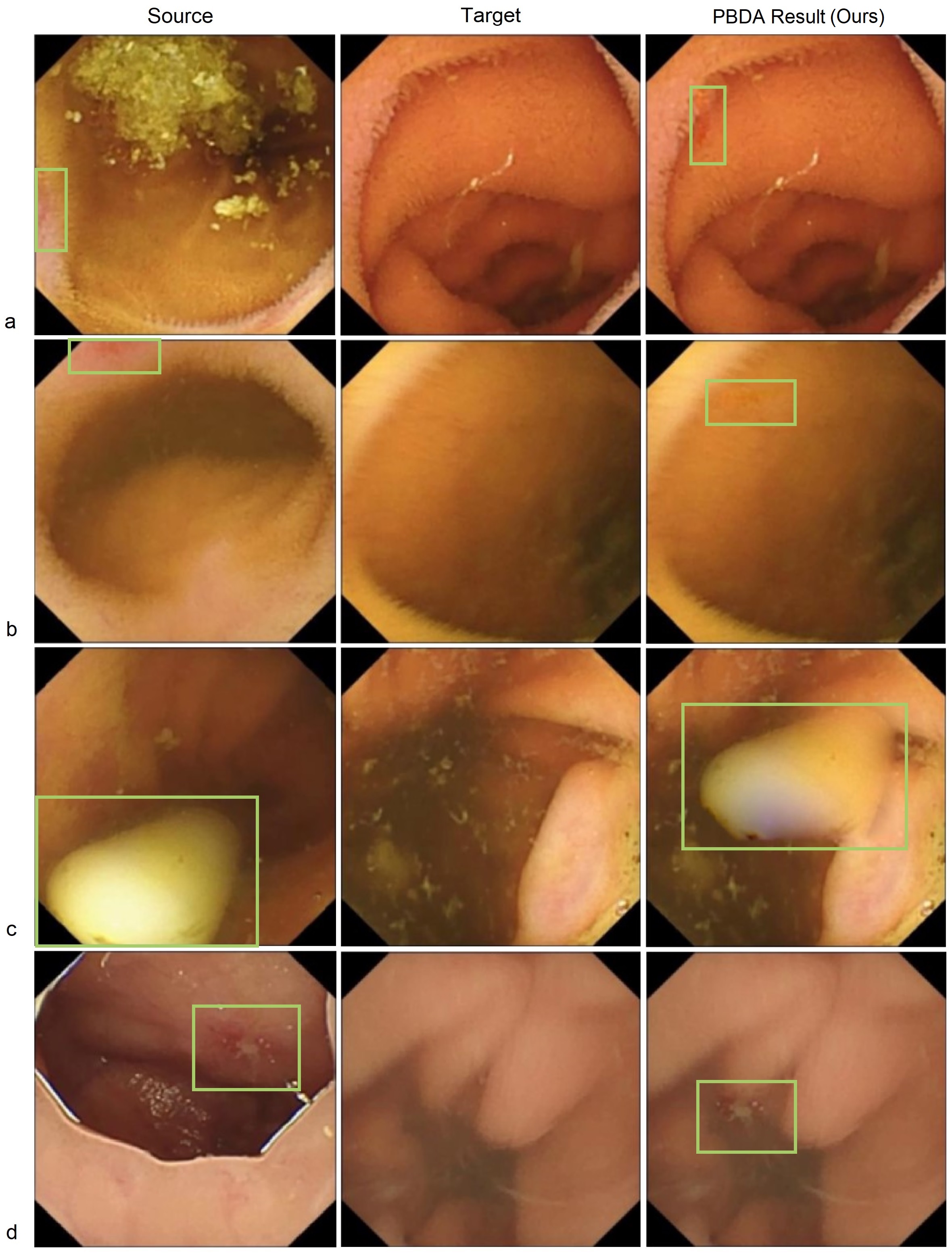}
    \caption{Corrupted a) erosion, b) erythema, c) foreign body, d) ulcer samples generated with PBDA pipeline. In some cases, the algorithm creates image compositions completely invisible to the human eye or with unrealistic colors.}
    \label{fig:figure6}
\end{figure}

Despite its strengths, the PBDA pipeline has limitations. When the source and target image pairs differ significantly, as illustrated in Figure \ref{fig:figure6}, it can produce unrealistic lesions, particularly in terms of color, as discussed in Section \ref{sec:methods_data_preparation}. Moreover, blending lesions that are either very small or very large poses challenges, making it difficult to create realistic results. Poor-quality blends often occur when the blending region on the target image contains a tissue fold or distant background.

More examples for each pathology class can be found in Appendix \ref{appendix:b}.

\subsection{IIDA - qualitative results}
\label{sec:experiments_lama}

This section presents experiments designed to evaluate the performance of the IIDA. The first experiment focuses on the method's ability to generate different types of lesions within a single image, testing whether it can produce diverse tissue types in the same context. Additionally, the effect of mask size on the generated content is analyzed. The experiments are repeated across multiple images for each pathological class, with the lesion mask selected at different locations to assess the method's adaptability in generating varied content across different areas of an image. Finally, the quality of the synthetic data produced by the proposed techniques is compared to that of state-of-the-art generative models.

\begin{figure}[htpb]
    \centering
    \includegraphics[width=0.6\linewidth]{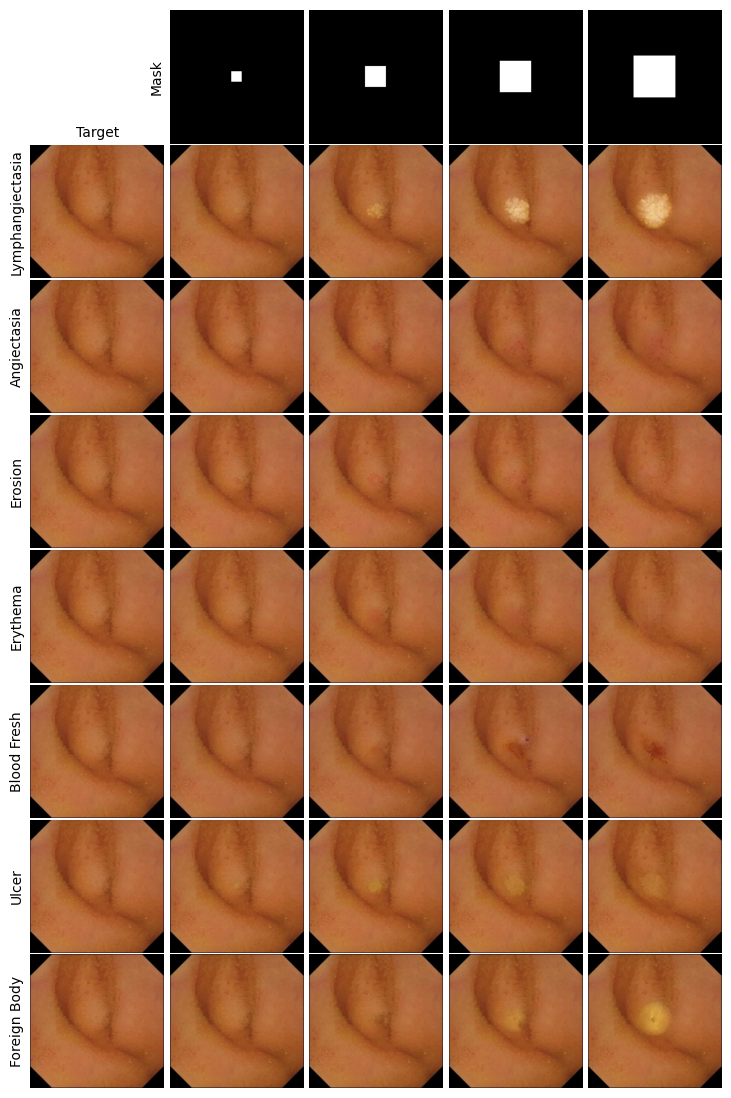}
    \caption{Lesions generated for each pathology class with different mask sizes, all positioned in the central region of the image. Generally, the models produce realistic content; however, with very small masks, the generator cannot produce lesions. A key advantage of the proposed method is its ability to generate diverse lesions within the same image. Zoom-in is recommended for best viewing.}
    \label{fig:figure7}
\end{figure}

\begin{figure}[htpb]
    \centering
    \includegraphics[width=\textwidth]{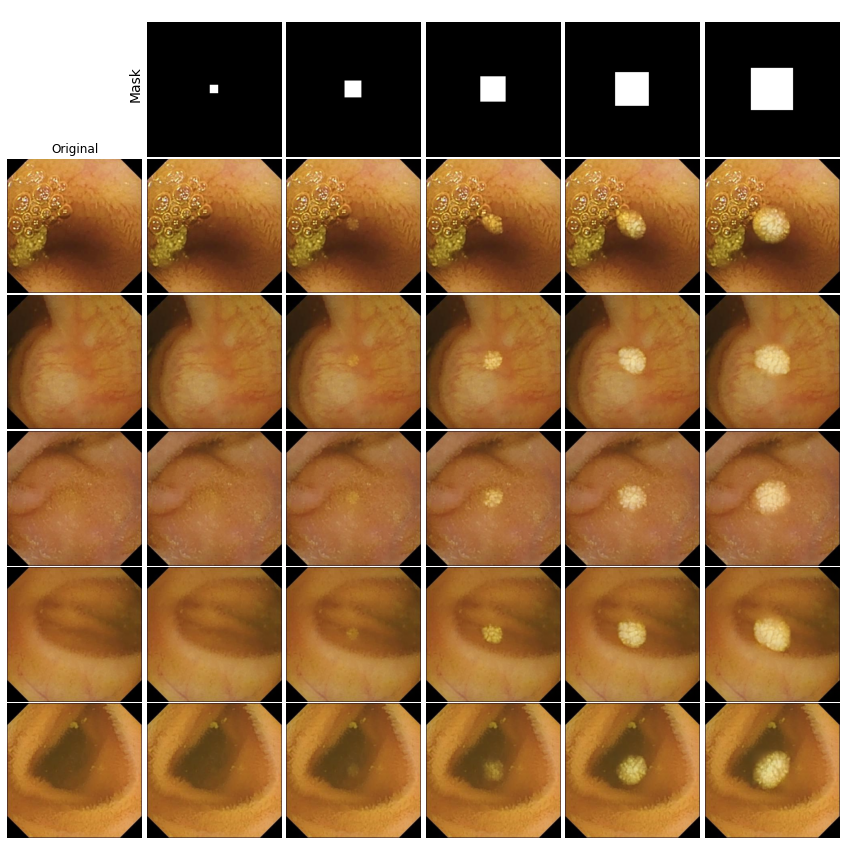}
    \caption{Lymphangiectasia generated with different mask sizes, all positioned in the central region of the image. The model demonstrates flexibility in content generation and effectively adapts to lighting, perspective, and tissue folds in the target image. }
    \label{fig:figure8}
\end{figure}

As demonstrated in Figure \ref{fig:figure7}, the model was provided with identical images, but with varying mask sizes centered in the same image region. The generated samples exhibit significant variability in size, shape, and texture, with lymphangiectasia being particularly well-represented. Additional examples of lymphangiectasia are provided in Figure \ref{fig:figure8}. The model successfully generated plausible lesions for most mask sizes, although it struggled with the smallest mask, failing to produce a lesion when the mask was too small. More examples for each pathology class can be found in Appendix \ref{appendix:c}.

\begin{figure}[htbp]
    \centering
    \includegraphics[width=0.77\linewidth]{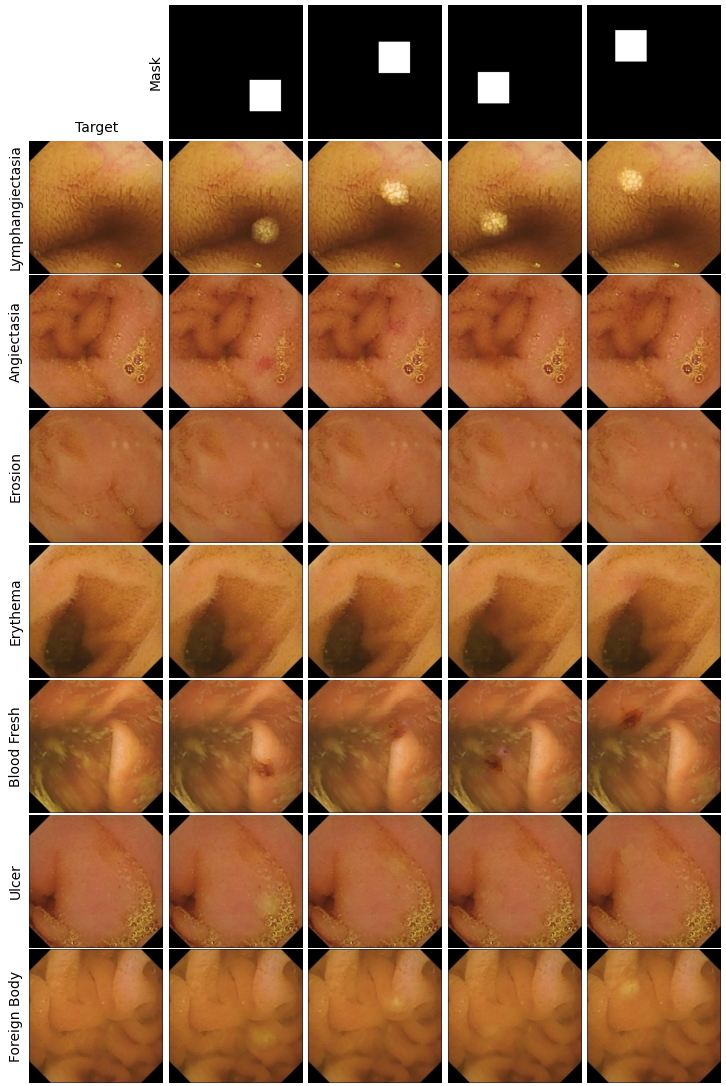}
    \caption{Lesions generated with a fixed mask size applied at different locations on an image. The model can adjust the generated lesion by altering its position, slightly modifying the content, and adapting the brightness. Zoom-in is recommended for best viewing.}
    \label{fig:figure9}
\end{figure}

The second experiment assessed the model’s ability to generate different pathological structures in various locations within the same image, using identical mask sizes (Figure \ref{fig:figure9}). The results show that the model can generate diverse pathological tissue in different areas, even when using the same mask size. This is a critical capability for generating varied pathological samples.

\begin{figure}[htpb]
    \centering
    \includegraphics[width=0.5\linewidth]{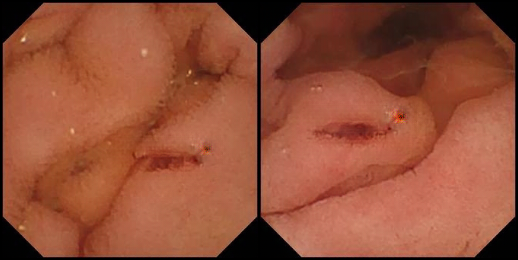}
    \caption{Fresh blood artifacts. The model produces characteristic artifacts when it is insufficiently trained, either during early fine-tuning stages or when provided with too few training samples.}
    \label{fig:figure10}
\end{figure}

The experiments led to the conclusion that while the model produced seamless content for each pathology class, only the lymphangiectasia, angiectasia, and erosion lesions were consistently of high quality. The variability in performance may be attributed to factors such as lesion complexity, distinguishability from the background, mask quality, dataset size, and intra-class variability. For example, Kvasir Capsule's Split 1 contains very few bounding boxes for the fresh blood and erythema classes (Table \ref{tab:combined_kvasir_splits}). Although fresh blood is a high-contrast class, which could occasionally lead to realistic lesion generation, artifacts were still introduced, as shown in Figure \ref{fig:figure10}. Despite the erythema class having more bounding boxes, its low contrast and similarity to healthy tissue may have hindered the model's ability to generate effective samples.

\begin{figure}[htbp]
    \centering
    \includegraphics[width=\linewidth]{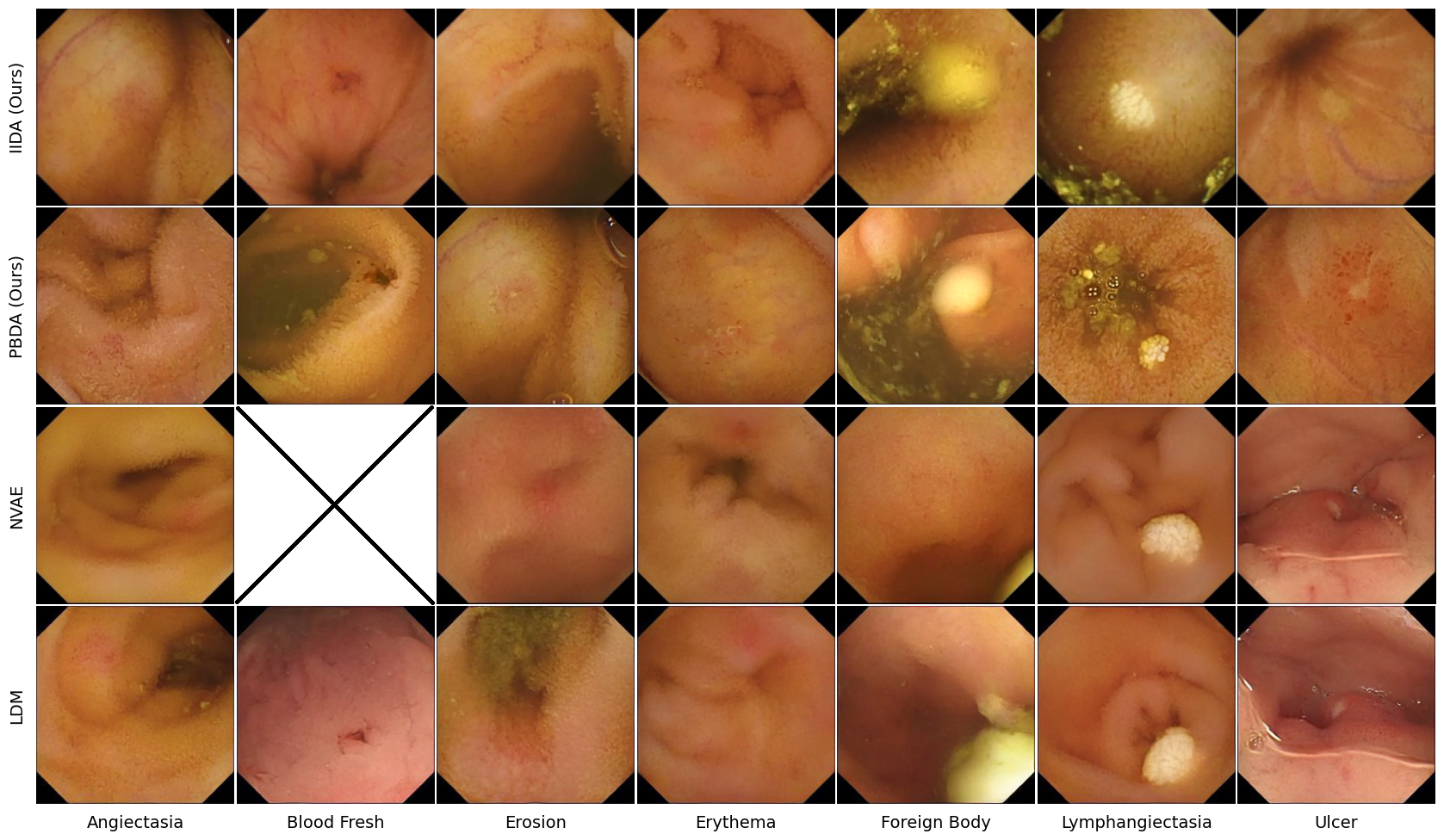}
    \caption{Visual comparison of evaluated methods. Each examined method can generate high-quality synthetic data, with most quality differences only visible in fine details. However, the proposed methods - PBDA and IIDA - modify only specific regions within real images of healthy tissue, so any artifacts are confined to a small part of the image. In contrast, NVAE and LDM occasionally create global distortions and often copy training data. The NVAE model for the blood fresh class failed to train, due to too low sample size.}
    \label{fig:figure11}
\end{figure}

To further validate the effectiveness of the proposed methods, NVAE and LDM, two modern generative models, were trained on the same data (Section \ref{sec:experiments_preparation}), and their sample quality was compared to that of the synthetic data generated by PBDA and IIDA (Figure \ref{fig:figure11}). While all methods produced high-quality synthetic data, differences were primarily observed in the details. The proposed methods, PBDA and IIDA, modify small regions of real images containing healthy tissue, which localizes any minor artifacts to specific areas of the image. In contrast, NVAE and LDM occasionally introduced global distortions or replicated training data, highlighting the advantages of the proposed approaches.

\clearpage

\subsection{Preparation of augmented dataset}
\label{sec:experiments_preparation}
In GDA, the standard approach to evaluating model performance involves measuring improvements in downstream task metrics compared to a baseline without GDA \cite{lee2024intraoperative,zheng2024toward,he2024image,bowles2018gan,trabucco2023effective}. For each proposed augmentation method - PBDA and IIDA - three versions of the augmented dataset were created. For each lesion class, synthetic images were generated in sufficient quantities to ensure that the total number of samples  per class in the training set reached 2,000, 5,000, or 10,000, depending on the experimental version. If any class exceeded these thresholds, it was limited to 2,000, 5,000, or 10,000 samples accordingly. For simplicity, these datasets are referred to as the 2k, 5k, and 10k datasets in the following sections. This approach ensured that each training set was balanced for the experiments.

PBDA experiments were prepared using the pipeline described in Section \ref{sec:poisson_blending_data_augmentation}. For IIDA, the same target images and locations on those images were used for augmentation, allowing for direct comparison between the two methods.

We also generated datasets with 2,000, 5,000, or 10,000 total samples per class, where $x\%$ of synthetic samples were created using PBDA, and the rest were created using IIDA. The values of $x$ were set to 20, 40, 60, and 80.

To compare our proposed methods with other GDA approaches, we trained NVAE and LDM models to generate capsule endoscopy images \textit{de novo}. Since NVAE is not class-conditional, separate models were trained for each pathology, using images from the training dataset, similar to the approach taken with the LaMa model. In contrast, LDM, being class-conditional, was trained on the entire Kvasir Capsule training dataset. As with the proposed methods, 2k, 5k, and 10k datasets were created using NVAE and LDM, and the resulting data were evaluated in classification experiments. NVAE was trained using the FFHQ setup provided in the official codebase \cite{vahdat2020nvaerepo}, while LDM was trained with the cin256-v2 configuration using the official codebase \cite{rombach2021highresolutionrepo}.

\subsection{Classification experiments}
\label{sec:experiments_classification}
To evaluate the effectiveness of our methods, we conducted experiments using various neural network architectures and compared their performance with standard techniques for addressing class imbalance. This comparison specifically assessed the influence of balancing the training dataset, which is a natural outcome of our proposed data augmentation strategy. We compared our best-performing model to the previous state-of-the-art approaches on the Kvasir Capsule dataset. Additionally, we investigated the impact of varying augmented sample sizes on classification accuracy, applying our techniques and comparing them with alternative GDA methods. Finally, we evaluated the lesion detection performance across the tested methods.

Several classification setups were designed to assess the effectiveness of the proposed data augmentation methods. Four models were compared: ResNet18 \cite{he2016deep} as a common baseline, ViT-T/16 \cite{dosovitskiy2020image} to evaluate transformer model performance on the Kvasir Capsule Dataset, and CAFormer-S18 \cite{yu2023metaformer} and ConvNext Tiny \cite{liu2022convnet} as modern architectures. All models were pretrained on ImageNet dataset \cite{deng2009imagenet} variants, with details provided in Table \ref{tab:experiments-configuration}. Standard data augmentation techniques, such as affine transformations (e.g., flip, rotation) and color transformations, were applied during training.

\begin{table}[htbp]
    \begin{center}
        \small{
        \begin{adjustbox}{width=\textwidth}
        \begin{tabular}{l|c|c|c|c}
        \toprule
        \textbf{Parameter} & \textbf{ResNet18} & \textbf{ConvNeXt-Tiny} & \textbf{ViT-T/16} & \textbf{CAFormer-S18} \\
        \midrule
        Version & resnet18 & convnext\_tiny  & vit\_tiny\_patch16\_224 & caformer\_s18 \\
        Pretraining  & ImageNet1k & ImageNet22k+ImageNet1k & ImageNet22k+ImageNet1k & ImageNet22k+ImageNet1k \\
        \midrule
        Train resolution & 224 & 224 & 224 & 224 \\
        Test resolution & 224 & 224 & 224 & 224 \\
        \midrule
        Optimizer & AdamP & AdamP & AdamP & AdamP \\
        Learning rate (LR) & \( 1.5 \times 10^{-4} \) & \( 1.5 \times 10^{-4} \) & \( 1.5 \times 10^{-4} \) & \( 3 \times 10^{-5} \) \\
        Betas & [0.9, 0.999] & [0.9, 0.999] & [0.9, 0.999] & [0.9, 0.999] \\
        Eps & \( 1 \times 10^{-8} \) & \( 1 \times 10^{-8} \) & \( 1 \times 10^{-8} \) & \( 1 \times 10^{-8} \) \\
        Weight decay & \( 1 \times 10^{-3} \) & \( 1 \times 10^{-3} \) & \( 1 \times 10^{-3} \) & \( 1 \times 10^{-3} \) \\
        Nesterov & \ding{51} & \ding{51} & \ding{51} & \ding{51} \\
        \midrule
        LR decay & cosine & cosine & cosine & cosine \\
        Decay epochs & 30 & 30 & 30 & 30 \\
        Initial warmup LR & \( 1 \times 10^{-7} \) & \( 1 \times 10^{-7} \) & \( 1 \times 10^{-7} \) & \( 1 \times 10^{-7} \) \\
        Warmup epochs & 3 & 3 & 3 & 3 \\
        Cycle decay & 0.1 & 0.1 & 0.1 & 0.1 \\
        \midrule
        Loss & Cross-entropy & Cross-entropy & Cross-entropy & Cross-entropy \\
        \midrule
        Rotation angles & [-15, 15] & [-15, 15] & [-15, 15] & [-15, 15] \\
        Rotate proba & 0.5 & 0.5 & 0.5 & 0.5 \\
        H. flip proba & 0.5 & 0.5 & 0.5 & 0.5 \\
        V. flip proba & 0.5 & 0.5 & 0.5 & 0.5 \\
        Brightness change limit & 0.1 & 0.1 & 0.1 & 0.1 \\
        Contrast change limit & 0.15 & 0.15 & 0.15 & 0.15 \\
        Brightness/Contrast proba & 0.5 & 0.5 & 0.5 & 0.5 \\
        Coarse dropout max holes & 20 & 20 & 20 & 20 \\
        Coarse dropout max height/max width & [8, 8] & [8, 8] & [8, 8] & [8, 8] \\
        Coarse dropout proba & 0.5 & 0.5 & 0.5 & 0.5 \\
        \bottomrule
        \end{tabular}
        \end{adjustbox}
        }
    \end{center}
    \caption{Overview of the training configurations for ResNet18, ConvNeXt-Tiny, ViT-T/16, and CAFormer-S18 models. The table includes details on model versions, pretraining, resolutions, optimizer settings, learning rate schedules, and other relevant parameters.}
    \label{tab:experiments-configuration}
\end{table}

We also explored class imbalance handling techniques, including Random Oversampling (ROS), Random Undersampling (RUS), thresholding, thresholding combined with ROS, and class weighting \cite{johnson2019survey}. In the thresholding with ROS approach, class counts were adjusted to match the number of samples in the angiectasia class. Classes with fewer samples were oversampled using ROS, while those with more samples were randomly undersampled. Thresholding alone only reduced the sample counts of more populous classes to the level of the angiectasia class. For comparison, models were also trained without any data sampling or class weighting.

\subsubsection{Comparison to state-of-the-art}
Experiments using PBDA, IIDA, NVAE, and LDM were conducted on the datasets described in Section \ref{sec:experiments_preparation}. Each experiment was run three times with different random seeds, and the average results with standard deviations are reported. Table \ref{tab:patientwise-comparison-to-reference} compares the classification performance of the best-performing model of our choice, CAFormer, to previous best results on the Kvasir Capsule Dataset official split. While few works have reported results on this data split, Srivastava et al. \cite{srivastava2022video} conducted an extensive study of modern classification models.

\begin{table}[htbp]
    \begin{center}
    \small{
        \begin{adjustbox}{width=\textwidth}
    \begin{tabular}{l|l|l|ccc|ccc}
    \toprule
 \multirow{2}{*}{\textbf{Model}} & \multirow{2}{*}{\textbf{Strategy}} & \multirow{2}{*}{\textbf{Data}} & \multicolumn{3}{c}{\textbf{Macro Average}} & \multicolumn{3}{c}{\textbf{Weighted Average}} \\
 \cmidrule(lr){4-6} \cmidrule(lr){7-9}
 & & & \textbf{F1-score} $\uparrow$ & \textbf{Precision} $\uparrow$ & \textbf{Sensitivity} $\uparrow$ & \textbf{F1-score} $\uparrow$ & \textbf{Precision} $\uparrow$ & \textbf{Sensitivity} $\uparrow$ \\ 
\midrule
ResNet-152 \cite{srivastava2022video} & \multirow[c]{7}{*}{Weighting} & \multirow[c]{7}{*}{Base} &  $14.63^{n/g}$ & $15.63^{n/g}$ & $20.49^{n/g}$ & $56.75^{n/g}$ & $72.95^{n/g}$ & $48.86^{n/g}$ \\	
Swin-S \cite{srivastava2022video} &  &  &  $15.25^{n/g}$ & $15.38^{n/g}$ & $23.88^{n/g}$ & $63.34^{n/g}$ & $73.90^{n/g}$ & $58.00^{n/g}$ \\
ConViT-B \cite{srivastava2022video} &  &   & $17.00^{n/g}$ & $17.69^{n/g}$ & $25.34^{n/g}$ & $61.60^{n/g}$ & $74.06^{n/g}$ & $55.41^{n/g}$ \\
DenseNet-169 \cite{srivastava2022video} &  &   & $14.83^{n/g}$ & $18.84^{n/g}$ & $22.62^{n/g}$ & $60.83^{n/g}$ & $71.41^{n/g}$ & $55.40^{n/g}$ \\	
FocalConvNet \cite{srivastava2022video} &  &  &  $21.78^{n/g}$ & $24.38^{n/g}$ & $27.45^{n/g}$ & $67.34^{n/g}$ & $75.57^{n/g}$ & $63.73^{n/g}$ \\
DenseNet-161 \cite{smedsrud2021kvasir} & & &  ${25.23}^{n/g}$ & ${29.33}^{n/g}$ & ${29.39}^{n/g}$ & - & - & - \\
ViT-B/16 \cite{regmi2023vision} & & & - & - & - & ${71.56}^{27.79}$ & ${68.41}^{29.85}$ &  ${71.56}^{28.99}$  \\
\midrule
 \multirow[c]{7}{*}{CAFormer-S18} & No strategy & \multirow[c]{2}{*}{Base} & $23.11^{1.79}$ & $38.51^{8.86}$ & $21.44^{2.52}$ & $83.35^{1.00}$ & $84.07^{0.96}$ & $\textbf{85.88}^{0.31}$ \\
  & RUS &  &  $25.26^{2.08}$ & $27.17^{3.41}$ & $30.60^{2.70}$ & $80.47^{0.38}$ & $82.06^{0.17}$ & $79.58^{0.76}$ \\
  \cmidrule{2-9}
  & 5k & LDM &  $20.46^{0.44}$ & $27.10^{5.03}$ & $21.82^{2.71}$ & $80.19^{2.04}$ & $81.20^{0.96}$ & $81.07^{3.89}$ \\
  & 2k & NVAE &  $23.92^{1.97}$ & $25.08^{2.66}$ & $29.24^{1.83}$ & $71.35^{3.38}$ & $73.31^{1.03}$ & $70.40^{5.26}$ \\
  \cmidrule{2-9}
  & \textbf{2k} & \textbf{IIDA (ours)} & $32.40^{1.71}$ & $\textbf{35.34}^{1.05}$ & $35.58^{2.99}$ & $82.93^{0.25}$ & $83.18^{1.43}$ & $83.65^{0.91}$ \\
  & \textbf{10k} & \textbf{PBDA (ours)} & $29.27^{3.43}$ & $29.22^{6.08}$ & $38.48^{2.31}$ & $81.35^{1.34}$ & $82.19^{1.26}$ & $82.19^{2.96}$ \\
   & $\textbf{2k}_{20\%PBDA - 80\%IIDA}$ & \textbf{PBDA+IIDA (ours)} & $\textbf{33.07}^{0.33}$ & $33.61^{1.02}$ & $\textbf{40.89}^{2.54}$ & $\textbf{83.83}^{0.98}$ & $\textbf{84.44}^{0.90}$ & $83.87^{1.09}$ \\
    \bottomrule
\end{tabular}
\end{adjustbox}}
\end{center}
\caption{The classification results are evaluated and compared to the previous benchmark on the Kvasir Capsule Dataset challenge using its official data split. In \protect\cite{srivastava2022video}, multiple common neural network architectures were tested for this challenge, with models trained using a weighted categorical cross-entropy strategy to address class imbalance, denoted as \textit{Weighting}. This study also includes CAFormer \protect\cite{yu2023metaformer}, one of the current best-performing models on the ImageNet challenge \protect\cite{deng2009imagenet}. Results are provided both without any data sampling strategy and with the best-performing sampling strategy. These baseline results are compared with those obtained through IIDA and PBDA, which substantially enhance classification accuracy and establish a new state-of-the-art for the Kvasir Capsule Dataset. In the \textit{Data} column the non-augmented dataset is labelled as \textit{Base}. Superscripts indicate the standard deviation across three experimental runs. The final row presents an experiment combining synthetic data, with 20\% generated via PBDA and 80\% via IIDA. The superscript n/g (not given) indicates that standard deviation was not given by respective authors.}
    \label{tab:patientwise-comparison-to-reference}
\end{table}

In Table \ref{tab:patientwise-comparison-to-reference}, the top experiments based on the macro average F1-Score using PBDA and IIDA are compared with the best performing sampling strategy for handling class imbalance in our experiments - RUS - and other results in the literature. CAFormer, one of the currently top-performing models on the ImageNet challenge, exceeded previous benchmarks, even when using only sampling techniques, as shown in Table \ref{tab:patientwise-comparison-to-reference}.

When PBDA and IIDA were applied during training, the models significantly outperformed previous highest results. PBDA improved F1-score macro average by 7.49 p.p. over Srivastava et al. \cite{srivastava2022video} and 4.04 p.p. over Smedsrud et al. \cite{smedsrud2021kvasir}. IIDA showed even larger gains, surpassing the same baselines by 10.62 and 7.17 p.p., respectively. Notably, CAFormer, trained on a 2k dataset balanced with 20\% PBDA and 80\% IIDA samples, achieved a new state-of-the-art F1-score macro average of 33.07\%.

The proposed methods were also evaluated against modern generative models NVAE and LDM, which generate images \textit{de novo} rather than modifying existing ones. Results demonstrated the superiority of PBDA and IIDA, as NVAE yielded only a minor classification improvement, while LDM adversely affected performance. Both generative models underperformed relative to the proposed techniques.

Our methods, PBDA and IIDA, achieved superior performance, setting a new state-of-the-art on the Kvasir Capsule Dataset. Both techniques outperformed the sampling strategies across all evaluated models, with IIDA consistently achieving better results (Table \ref{tab:patientwise_augmented_sample_size}). Specifically, the IIDA pipeline surpassed the best sampling method, RUS, by 7.14 p.p. and PBDA surpassed it by 4.01 p.p., while a combination of PBDA and IIDA reached improvement of 7.81 p.p. Both proposed methods greatly outperformed data augmentation using LDM and NVAE models.

The proposed techniques significantly improved lesion detection, yielding sensitivity increases of 6.19, 9.09, and 11.50 p.p. for IIDA, PBDA, and their combined use, respectively, over the previous state-of-the-art. These results indicate that incorporating synthetic lesions into real images of healthy tissue enhances the classifier's ability to detect pathological changes effectively.

Following prior studies, we included weighted average metrics in our comparison. However, due to substantial class imbalance within the dataset, these metrics are predominantly influenced by performance on the \textit{normal} class, with limited contribution from the pathological classes, therefore we believe that macro average is the most reliable and informative metric for the example under consideration. In our optimal configuration, the combined application of PBDA and IIDA achieved the highest weighted average F1-score and precision values.

\subsubsection{Study on the effect of augmented sample size}

\begin{table}[htbp]
    \begin{center}
        \begin{adjustbox}{width=\textwidth}
        \begin{tabular}{ll|llll|llll|llll|llll}
        \toprule
    \multirow{2}{*}{\textbf{Data}} & \multirow{2}{*}{\textbf{Strategy}} & \multicolumn{4}{c}{\textbf{CAFormer-S18}} \vline &  \multicolumn{4}{c}{\textbf{ConvNeXt-Tiny}} \vline & \multicolumn{4}{c}{\textbf{ResNet18}} \vline & \multicolumn{4}{c}{\textbf{ViT-T/16}} \\
    \cmidrule(lr){3-6} \cmidrule(lr){7-10} \cmidrule(lr){11-14} \cmidrule(lr){15-18}
 &  &  \textbf{AUC} $\uparrow$ & \textbf{F1-score} $\uparrow$ & \textbf{Sens.} $\uparrow$ & \textbf{Spec.} $\uparrow$ & \textbf{AUC} $\uparrow$ & \textbf{F1-score} $\uparrow$ & \textbf{Sens.} $\uparrow$ & \textbf{Spec.} $\uparrow$ & \textbf{AUC} $\uparrow$ & \textbf{F1-score} $\uparrow$ & \textbf{Sens.} $\uparrow$ & \textbf{Spec.} $\uparrow$ & \textbf{AUC} $\uparrow$ & \textbf{F1-score} $\uparrow$ & \textbf{Sens.} $\uparrow$ & \textbf{Spec.} $\uparrow$ \\ \midrule
\multirow[t]{7}{*}{\textbf{Base}} & No strategy & $86.78^{2.56}$ & $23.11^{1.79}$ & $21.44^{2.52}$ & $92.42^{0.87}$ & $79.87^{8.18}$ & $19.73^{3.30}$ & $22.23^{3.89}$ & $92.12^{1.11}$ & $73.08^{3.28}$ & $14.29^{1.51}$ & $17.64^{0.90}$ & $90.73^{0.48}$ & $63.37^{13.31}$ & $10.97^{3.31}$ & $14.94^{3.36}$ & $91.17^{1.10}$ \\
 & Weighting & $83.55^{4.12}$ & $23.09^{0.62}$ & $28.08^{6.29}$ & $93.03^{1.42}$ & $80.01^{3.87}$ & $18.93^{2.03}$ & $28.62^{3.53}$ & $93.10^{1.22}$ & $65.01^{4.07}$ & $14.72^{1.67}$ & $20.86^{1.60}$ & $92.04^{0.60}$ & $74.53^{4.77}$ & $14.96^{2.72}$ & $23.92^{1.87}$ & $91.87^{0.26}$ \\
 & ROS & $85.41^{4.31}$ & $24.14^{5.30}$ & $24.03^{5.49}$ & $92.16^{0.46}$ & $82.48^{2.81}$ & $22.31^{3.35}$ & $27.67^{5.02}$ & $91.81^{0.61}$ & $75.06^{5.88}$ & $19.13^{3.42}$ & $21.09^{4.75}$ & $91.59^{0.99}$ & $68.00^{16.85}$ & $10.44^{6.94}$ & $13.30^{6.08}$ & $89.90^{1.24}$ \\
 & RUS & $85.14^{2.47}$ & $25.26^{2.08}$ & $30.60^{2.70}$ & $93.16^{0.15}$ & $78.29^{5.99}$ & $19.98^{4.10}$ & $23.60^{5.70}$ & $92.91^{0.72}$ & $64.54^{10.72}$ & $15.33^{2.78}$ & $19.78^{1.90}$ & $92.43^{0.59}$ & $65.23^{14.90}$ & $13.67^{3.74}$ & $19.23^{3.10}$ & $92.59^{0.69}$ \\
 & Thresholded & $80.67^{7.38}$ & $21.92^{2.95}$ & $27.37^{2.21}$ & $93.26^{0.24}$ & $80.45^{4.69}$ & $18.89^{1.13}$ & $25.88^{3.48}$ & $92.32^{0.56}$ & $62.45^{3.25}$ & $13.86^{1.98}$ & $20.12^{1.71}$ & $92.52^{1.00}$ & $70.23^{9.43}$ & $14.53^{1.75}$ & $18.96^{1.60}$ & $92.00^{0.40}$ \\
 & Thresholded ROS & $83.31^{2.14}$ & $25.14^{3.64}$ & $28.58^{5.39}$ & $93.33^{0.55}$ & $74.72^{1.83}$ & $19.09^{0.90}$ & $26.44^{0.68}$ & $93.48^{0.95}$ & $67.93^{8.24}$ & $14.66^{2.67}$ & $20.76^{2.29}$ & $92.61^{0.55}$ & $75.94^{0.72}$ & $17.04^{0.47}$ & $25.19^{3.39}$ & $92.72^{0.93}$ \\

\midrule
\multirow[t]{3}{*}{\textbf{NVAE}} & 10k & $51.82^{3.67}$ & $9.64^{2.35}$ & $12.57^{5.17}$ & $88.89^{0.44}$ & $51.09^{6.07}$ & $7.55^{2.93}$ & $10.00^{2.77}$ & $89.05^{0.34}$ & $44.88^{1.96}$ & $5.56^{3.02}$ & $9.21^{1.48}$ & $88.77^{0.28}$ & $55.47^{13.15}$ & $9.74^{7.93}$ & $14.10^{7.91}$ & $89.46^{1.31}$ \\
 & 5k & $69.03^{13.33}$ & $20.40^{9.06}$ & $21.21^{5.68}$ & $90.72^{1.47}$ & $66.26^{11.84}$ & $16.61^{10.05}$ & $23.49^{8.39}$ & $90.86^{1.40}$ & $57.73^{16.17}$ & $11.20^{10.73}$ & $15.47^{6.68}$ & $89.99^{1.55}$ & $59.27^{15.35}$ & $10.70^{9.63}$ & $16.14^{9.01}$ & $89.93^{1.65}$ \\
 & 2k & $75.93^{4.90}$ & $23.92^{1.97}$ & $29.24^{1.83}$ & $92.21^{0.36}$ & $74.60^{3.88}$ & $19.88^{1.10}$ & $24.50^{3.89}$ & $91.93^{0.65}$ & $71.87^{3.39}$ & $18.64^{0.56}$ & $21.92^{0.35}$ & $91.82^{0.20}$ & $72.66^{4.15}$ & $17.60^{2.75}$ & $21.30^{3.40}$ & $91.85^{0.60}$ \\

\midrule
\multirow[t]{3}{*}{\textbf{LDM}} & 10k & $75.14^{2.81}$ & $19.62^{2.62}$ & $20.01^{5.13}$ & $92.31^{1.13}$ & $79.31^{6.48}$ & $19.44^{5.26}$ & $22.55^{3.02}$ & $92.52^{0.65}$ & $74.20^{2.81}$ & $19.54^{1.45}$ & $21.83^{2.51}$ & $92.06^{0.56}$ & $68.13^{15.84}$ & $11.91^{6.24}$ & $14.69^{6.17}$ & $90.55^{1.03}$ \\
 & 5k & $79.39^{3.07}$ & $20.46^{0.44}$ & $21.82^{2.71}$ & $91.95^{0.42}$ & $80.17^{3.62}$ & $17.36^{3.80}$ & $19.24^{3.01}$ & $91.96^{0.83}$ & $76.19^{7.97}$ & $21.00^{1.55}$ & $23.84^{3.24}$ & $92.39^{0.71}$ & $78.86^{1.62}$ & $19.38^{1.49}$ & $22.45^{2.29}$ & $91.91^{0.48}$ \\
 & 2k & $78.13^{3.42}$ & $18.57^{0.41}$ & $22.29^{1.05}$ & $92.25^{0.27}$ & $80.99^{1.19}$ & $22.45^{2.36}$ & $28.44^{0.66}$ & $92.84^{1.03}$ & $75.98^{3.82}$ & $16.82^{2.97}$ & $23.13^{2.07}$ & $92.40^{0.21}$ & $81.78^{0.61}$ & $19.65^{0.83}$ & $24.55^{1.82}$ & $92.82^{0.43}$ \\

\midrule
\multirow[t]{3}{*}{\textbf{PBDA (ours)}} & 10k & $83.15^{4.34}$ & $29.27^{3.43}$ & $38.48^{2.31}$ & $93.40^{0.57}$ & $76.73^{1.22}$ & $24.44^{0.85}$ & $34.05^{1.46}$ & $93.02^{0.60}$ & $75.34^{1.14}$ & $22.88^{1.24}$ & $33.86^{1.32}$ & $93.28^{0.53}$ & $74.57^{2.03}$ & $16.77^{1.56}$ & $30.34^{0.86}$ & $93.27^{0.54}$ \\
 & 5k & $81.33^{4.21}$ & $27.98^{2.59}$ & $38.17^{2.22}$ & $93.58^{0.55}$ & $76.99^{3.15}$ & $24.77^{3.23}$ & $34.44^{1.59}$ & $93.22^{0.59}$ & $77.39^{0.97}$ & $21.26^{1.26}$ & $33.54^{0.79}$ & $93.51^{0.20}$ & $72.43^{2.31}$ & $17.32^{0.83}$ & $30.44^{3.20}$ & $\textbf{93.34}^{0.17}$ \\
 & 2k & $81.89^{1.84}$ & $25.85^{1.90}$ & $36.71^{3.17}$ & $93.78^{0.35}$ & $75.22^{2.41}$ & $22.47^{2.82}$ & $36.41^{0.75}$ & $\textbf{93.83}^{0.37}$ & $71.47^{1.52}$ & $20.98^{0.98}$ & $31.82^{2.55}$ & $93.38^{0.33}$ & $73.55^{2.10}$ & $18.30^{0.41}$ & $\textbf{32.25}^{2.48}$ & $92.91^{0.14}$ \\
 
\midrule
\multirow[t]{3}{*}{\textbf{IIDA (ours)}} & 10k & $84.18^{2.84}$ & $27.43^{0.69}$ & $30.33^{0.82}$ & $92.50^{0.12}$ & $\textbf{84.53}^{2.71}$ & $22.43^{4.40}$ & $27.91^{1.53}$ & $91.68^{0.33}$ & $74.12^{2.96}$ & $21.47^{2.24}$ & $25.74^{5.66}$ & $92.06^{0.41}$ & $\textbf{81.80}^{2.88}$ & $\textbf{22.92}^{0.23}$ & $28.54^{4.41}$ & $91.74^{0.62}$ \\
  & 5k & $86.78^{1.60}$ & $29.68^{2.38}$ & $33.01^{3.04}$ & $92.58^{0.69}$ & $83.60^{3.71}$ & $25.83^{3.05}$ & $27.98^{2.53}$ & $92.45^{0.51}$ & $78.12^{1.76}$ & $22.18^{3.10}$ & $25.23^{3.30}$ & $92.13^{0.59}$ & $78.03^{1.17}$ & $20.21^{1.26}$ & $25.90^{3.19}$ & $92.09^{0.70}$ \\
 & 2k & $\textbf{88.30}^{1.51}$ & $32.40^{1.71}$ & $35.58^{2.99}$ & $93.63^{0.98}$ & $82.99^{4.73}$ & $26.38^{2.44}$ & $29.82^{0.62}$ & $92.50^{0.27}$ & $76.04^{2.40}$ & $24.12^{0.13}$ & $26.55^{0.71}$ & $92.33^{0.67}$ & $76.32^{1.15}$ & $19.45^{2.84}$ & $22.74^{1.17}$ & $92.15^{0.31}$ \\

 \midrule
 \multirow[t]{3}{*}{\textbf{20\% PBDA + 80\% IIDA (ours)}} & 10k & $87.90^{2.53}$ & $29.19^{1.66}$ & $39.33^{5.07}$ & $93.36^{0.64}$ & $79.93^{1.41}$ & $27.69^{0.69}$ & $33.68^{2.50}$ & $92.47^{0.09}$ & $\textbf{78.58}^{4.28}$ & $24.82^{2.07}$ & $35.46^{2.67}$ & $93.28^{0.89}$ & $77.24^{2.57}$ & $19.63^{0.17}$ & $31.64^{0.69}$ & $92.85^{0.17}$ \\
  & 5k & $85.71^{1.25}$ & $31.27^{1.11}$ & $\textbf{42.59}^{3.40}$ & $93.68^{0.67}$ & $82.21^{3.87}$ & $27.34^{1.87}$ & $37.43^{1.73}$ & $93.62^{0.96}$ & $77.82^{2.65}$ & $24.46^{1.05}$ & $\textbf{36.06}^{1.40}$ & $93.55^{0.16}$ & $79.61^{3.18}$ & $19.66^{1.05}$ & $28.96^{3.09}$ & $92.86^{0.54}$ \\
 & 2k & $85.24^{0.47}$ & $\textbf{33.07}^{0.33}$ & $40.89^{2.54}$ & $\textbf{94.24}^{0.55}$ & $82.29^{0.27}$ & $\textbf{28.54}^{1.24}$ & $\textbf{38.76}^{4.75}$ & $93.60^{0.86}$ & $76.93^{5.76}$ & $\textbf{25.81}^{1.01}$ & $35.16^{0.03}$ & $\textbf{93.84}^{0.69}$ & $71.98^{2.61}$ & $19.99^{0.40}$ & $26.57^{4.99}$ & $92.83^{0.08}$ \\

 \bottomrule

\end{tabular}

    \end{adjustbox}
    
    \end{center}
    \caption{Classification results on the official Kvasir Capsule data split. Columns report the macro average AUC, F1-score, sensitivity, and specificity on the test set. We compared several established and modern architectures, using different sampling methods to address the class imbalance, including ROS, RUS, thresholding, thresholding with ROS, and class weighting. CAFormer \protect\cite{yu2023metaformer} achieved the highest classification performance across most scenarios. The IIDA allows for the highest increase in F1-Score, followed by PBDA for each trained model. Superscript values represent the standard deviation from three experimental runs. Results in bold are the highest values in each column.}
    \label{tab:patientwise_augmented_sample_size}
\end{table}

Table \ref{tab:patientwise_augmented_sample_size} highlights how augmented sample size influences classification model performance. For most models using IIDA, a balanced dataset of 2k training samples yields the best results, with the exception of ViT-T/16, which performs optimally with a 10k dataset. In contrast, PBDA performs best with a 10k dataset, except for ViT-T/16 and ConvNext-Tiny.

Notably, the NVAE model experiences a significant performance drop with a 10k dataset, suggesting that the classifier may exploit shortcuts \cite{geirhos2020shortcut} when classifying large amounts of synthetic data. Similarly, ResNet18 and ViT-T/16 show considerable performance degradation with the 5k dataset. However, this issue does not arise with data generated by LDM, except for ViT, even though LDM also generates images \textit{de novo}.

\begin{figure}[htbp]
    \centering
    \includegraphics[width=0.65\linewidth]{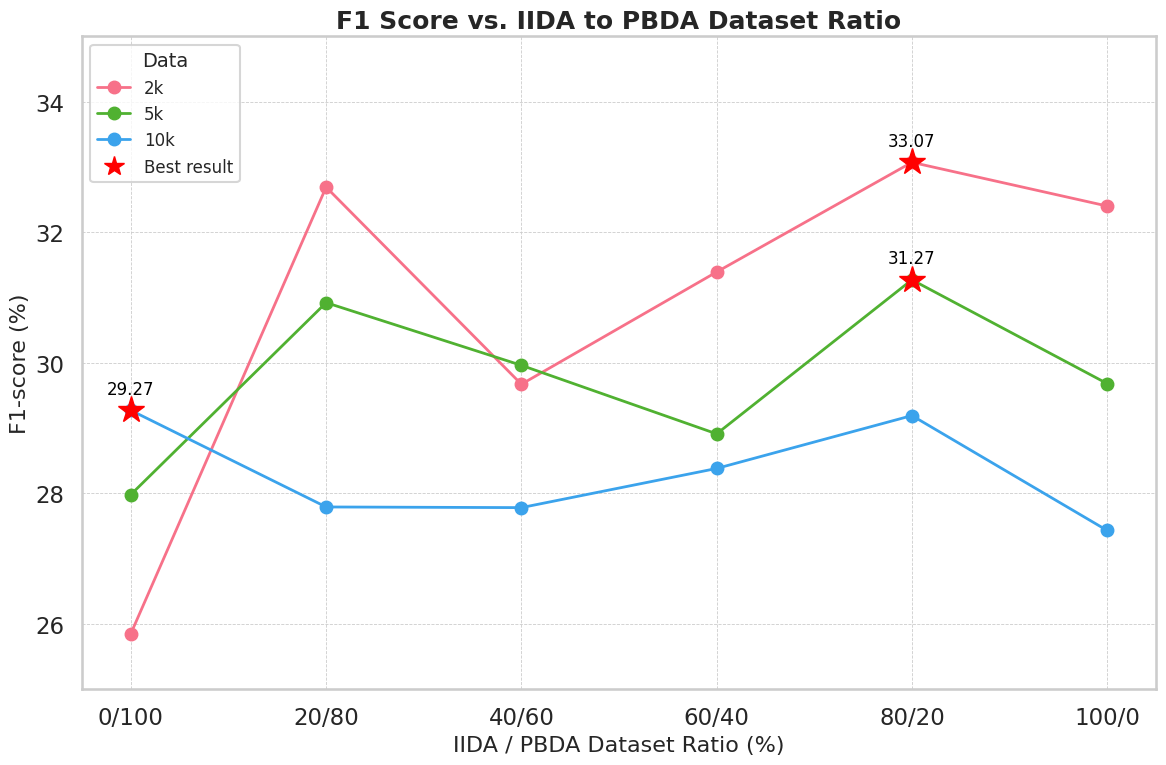}
    \caption{The chart displays F1-score values on the test set in relation to the proportion of samples generated using IIDA compared to those from PBDA. The results are presented across three distinct training set sizes: 2k, 5k, and 10k. }
    \label{fig:figure12}
\end{figure}

We conducted additional experiments where synthetic training data was a combination of $x\%$ PBDA-generated data and $(100 - x\%)$ IIDA-generated data. Figure \ref{fig:figure12} shows the results for varying proportions of these data generation methods. The optimal performance, particularly in terms of F1-Score macro average, was achieved with a 2k dataset where 20\% of the data came from PBDA and 80\% from IIDA. The highest macro average sensitivity of 42.59\% was reached using the 5k dataset. Across datasets of 2k, 5k, and 10k samples, there was no clear trend in performance improvement based on the proportion of PBDA to IIDA, except that the 20\% PBDA and 80\% IIDA configuration usually produced the best results. Similar to IIDA experiments, the 2k dataset yields the best results.

\clearpage

\subsubsection{Analysis of lesion detection}
The proposed methods significantly enhance lesion detection, particularly for lymphangiectasia and angiectasia classes, as indicated in Table \ref{tab:patientwise-per-class-comparison}. Using CAFormer with IIDA, the F1-scores for angiectasia and lymphangiectasia increased from 8.61\% to 36.14\% and from 23.99\% to 56.60\%, respectively. With PBDA, these scores improved to 17.6\% and 62.05\% for the respective classes. Combining these methods resulted in improved detection across most pathological classes, with a notable increase in sensitivity.

\begin{table}[htbp]
    \begin{center}
    \small{
        \begin{adjustbox}{width=\textwidth}
\begin{tabular}{l|ll|ll|ll|ll|ll|ll}
\toprule
\multirow{2}{*}{\textbf{Data}} & \multicolumn{2}{c}{\textbf{RUS}} & \multicolumn{2}{c}{\textbf{PBDA+IIDA (ours)}} & \multicolumn{2}{c}{\textbf{IIDA (ours)}} & \multicolumn{2}{c}{\textbf{PBDA (ours)}} & \multicolumn{2}{c}{\textbf{NVAE}} & \multicolumn{2}{c}{\textbf{LDM}} \\
\cmidrule(lr){2-3} \cmidrule(lr){4-5} \cmidrule(lr){6-7} \cmidrule(lr){8-9} \cmidrule(lr){10-11} \cmidrule(lr){12-13}
 & \textbf{F1-score} $\uparrow$ & \textbf{Sens.} $\uparrow$ & \textbf{F1-score} $\uparrow$ & \textbf{Sens.} $\uparrow$ & \textbf{F1-score} $\uparrow$ & \textbf{Sens.} $\uparrow$ & \textbf{F1-score} $\uparrow$ & \textbf{Sens.} $\uparrow$ & \textbf{F1-score} $\uparrow$ & \textbf{Sens.} $\uparrow$ & \textbf{F1-score} $\uparrow$ & \textbf{Sens.} $\uparrow$ \\
 \midrule
 \textbf{Average} & $25.26^{2.08}$ & $30.60^{2.70}$ & $\textbf{33.07}^{0.33}$ & 
 $\textbf{40.89}^{2.54}$ & $32.40^{1.71}$ & $35.58^{2.99}$ & $29.27^{3.43}$ & $38.48^{2.31}$ & $23.92^{1.97}$ &$29.24^{1.83}$ & $22.45^{2.36}$ & $28.44^{0.66}$ \\
 \midrule
\textbf{Normal} & $89.47^{0.40}$ & $87.86^{1.06}$ & $\textbf{92.63}^{0.42}$ & $92.68^{0.79}$ & $92.10^{0.75}$ & $92.64^{3.16}$ & $92.23^{1.47}$ & $\textbf{93.05}^{4.12}$ &  $78.24^{3.74}$ & $76.09^{6.43}$ & $89.51^{0.86}$ & $88.18^{2.05}$ \\
\textbf{Unclear view} & $49.85^{2.19}$ & $53.76^{6.19}$ & $49.51^{9.21}$ & $45.64^{14.17}$ & $43.64^{10.61}$ & $46.88^{22.67}$ & $19.97^{10.65}$ & $14.75^{10.70}$ &  $\textbf{56.55}^{5.80}$ & $\textbf{65.77}^{11.82}$ & $34.43^{8.29}$ & $28.05^{9.26}$ \\
\midrule
\textbf{Erosion} & $9.07^{3.97}$ & $40.82^{16.25}$ & $14.95^{0.94}$ & $53.00^{7.65}$ & $\textbf{15.98}^{0.87}$ & $46.82^{6.56}$ & $15.79^{3.75}$ & $\textbf{55.06}^{5.92}$ & $10.05^{2.10}$ & $27.99^{7.97}$ & $2.91^{0.81}$ & $12.55^{7.11}$ \\
\textbf{Angiectasia} & $8.61^{3.31}$ & $9.47^{2.79}$ & $22.86^{7.76}$ & $43.16^{5.26}$ & $\textbf{36.14}^{10.26}$ & $\textbf{44.91}^{20.69}$ & $17.60^{5.83}$ & $33.68^{1.05}$ & $2.84^{2.11}$ & $3.54^{2.76}$ & $12.20^{1.21}$ & $31.23^{16.85}$ \\
\textbf{Ulcer} & $2.10^{1.06}$ & $1.32^{0.65}$ & $2.23^{0.41}$ & $1.37^{0.30}$ & $2.02^{1.91}$ & $1.20^{1.20}$ & $2.88^{2.09}$ & $2.00^{1.39}$ & $0.53^{0.32}$ & $0.40^{0.20}$ & $\textbf{4.85}^{4.08}$ & $\textbf{3.61}^{3.31}$ \\
\textbf{Lymphangiectasia} & $23.99^{4.90}$ & $16.12^{3.86}$ & $56.88^{8.09}$ & $50.00^{14.42}$ & $56.60^{4.56}$ & $41.39^{3.29}$ & $\textbf{62.05}^{9.34}$ & $\textbf{64.49}^{7.54}$ & $25.57^{8.47}$ & $16.62^{5.78}$ & $19.64^{13.26}$ & $13.95^{9.51}$ \\
\textbf{Erythema} & $5.15^{2.55}$ & $17.28^{4.28}$ & $6.44^{6.07}$ & $\textbf{25.93}^{22.53}$ & $0.00^{0.00}$ & $0.00^{0.00}$ & $5.14^{7.41}$ & $16.05^{16.70}$ & $\textbf{8.98}^{2.35}$ & $15.07^{6.03}$ & $0.00^{0.00}$ & $0.00^{0.00}$ \\
\textbf{Foreign body} & $39.14^{10.02}$ & $48.75^{10.44}$ & $\textbf{52.13}^{5.26}$ & $56.27^{3.29}$ & $45.14^{7.76}$ & $46.42^{10.37}$ & $47.75^{8.75}$ & $67.20^{11.64}$ & $32.51^{1.87}$ & $57.68^{12.63}$ & $38.37^{20.11}$ & $\textbf{78.32}^{22.12}$ \\
\textbf{Blood fresh} & $0.00^{0.00}$ & $0.00^{0.00}$ & $0.00^{0.00}$ & $0.00^{0.00}$ & $0.00^{0.00}$ & $0.00^{0.00}$ & $0.00^{0.00}$ & $0.00^{0.00}$ & $0.00^{0.00}$ & $0.00^{0.00}$ & $\textbf{0.15}^{0.27}$ & $\textbf{0.08}^{0.14}$ \\
\bottomrule
\end{tabular}
\end{adjustbox}}
\end{center}
\caption{Results on the official Kvasir Capsule data split per class, showing unweighted F1-scores for the best experimental setups across all methods. PBDA and IIDA significantly enhance lesion detection, with IIDA yielding substantial classification improvements for consistently well-generated classes: F1-scores for angiectasia increased from 8.61\% to 36.14\%, and for lymphangiectasia from 23.99\% to 56.60\%. Both methods, individually and in combination, led to a notable increase in sensitivity for pathological classes. Results in bold are the highest values in each row for each metric.}
\label{tab:patientwise-per-class-comparison}
\end{table}

These findings highlight the primary advantage of local lesion generation in medical image classification: enhanced lesion detection. The lymphangiectasia and angiectasia classes showed the most significant gains, benefiting from the high-quality generation by the inpainting model. However, data augmentation effectiveness was limited for the fresh blood class, likely due to a domain shift between training and test data. Additionally, the underrepresentation of fresh blood and erythema samples in the training set further reduced the model's ability to generate these classes effectively.

Data augmentation using LDM did not yield consistent classification improvements across pathological classes while using NVAE improved the results only for erosion, lymphangiectasia, and erythema.

\section{Discussion}
\label{sec:discussion}
This study introduces two novel methods for generating local lesions, aimed at enhancing capsule endoscopy image data augmentation in scenarios with limited data. The first method, PBDA, does not require model training, making it applicable across all dataset sizes. The second method, IIDA, requires a little amount of data but further improves classification performance. When used together, these techniques achieve state-of-the-art lesion recognition results.

Extensive qualitative and quantitative experiments were conducted to evaluate both methods. Sections \ref{sec:discussion-pbda}, \ref{sec:discussion-iida} and \ref{sec:discussion-pbda+iida} provide detailed discussions on PBDA, IIDA, and their combined application, respectively. In Section \ref{sec:discussion-comparison}, the results are compared to previous approaches using the Kvasir Capsule Dataset, as well as to augmentation techniques based on other generative models. Finally, Section \ref{sec:discussion-future-work} outlines potential directions for future research.

\subsection{Poisson Blending Data Augmentation}
\label{sec:discussion-pbda}
In this work, we introduced the PBDA pipeline aimed at maximizing Poisson Blending effectiveness, by leveraging the unique characteristics of capsule endoscopy data. The deduplication step increases the variance of healthy tissue by removing abundant, near-duplicate images. Identification of the blending location is optimized for minimizing the adjustment to a blended lesion, thereby improving the consistency of blended samples. The pipeline excels at lesion blending when the source sample is clearly visible, contrasts with the surrounding tissue, and is accurately labeled and localized. By effectively utilizing the excessive healthy tissue, PBDA successfully augments the training set without the need for additional model training, making it applicable to any available data sample. 

Despite its strengths, PBDA has certain limitations. The algorithm for selecting blending locations could be refined to avoid regions such as folds or dark areas, which can hinder effective blending.  Additionally, when image pairs are poorly matched, the algorithm may fail to produce a visible lesion, or color bleeding and blending artifacts may occur. This may result in altering the semantics (class) of the blended lesion and introducing label noise into the synthesized dataset. This issue is more pronounced when lesion masks are of poor quality or when the lesion is small, obscured, or partially occluded. Another drawback is that PBDA requires bounding boxes, which are more labor-intensive and expensive to generate compared to simple class annotations. Moreover, PBDA does not create new lesion shapes, sizes, or textures; it only blends existing lesions seamlessly. 

Experimental results (Table \ref{tab:patientwise_augmented_sample_size}) demonstrate PBDA's effectiveness across varying sample sizes and model architectures. For the CAFormer architecture, all tested datasets augmented with PBDA outperform previous state-of-the-art results. However, no clear performance trend was observed across different architectures, suggesting that the choice of architecture and augmentation strategy should be jointly optimized based on the dataset and task. Additionally, PBDA consistently improves sensitivity compared to IIDA but at the cost of a reduced F1-score. The lower F1-scores for healthy tissue and unclear views, combined with substantial improvements for lesions  (Table \ref{tab:patientwise-per-class-comparison}), indicate that PBDA may produce more varied lesions, therefore enhancing generalization, while also more frequently failing to generate any lesion, leading to augmenting the training set with mislabeled samples.

\subsection{Image Inpainting Data Augmentation}
\label{sec:discussion-iida}
The IIDA method introduced in this work fine-tunes an image inpainting model to convincingly fill missing image regions with realistic lesions. As demonstrated in Figures \ref{fig:figure7}, \ref{fig:figure8} and \ref{fig:figure9} IIDA adapts the shape, size, and brightness of generated lesions to blend seamlessly with the surrounding tissue. It also enables the generation of structures with depth, such as lymphangiectasia, which may protrude into the intestinal lumen.  Unlike PBDA, IIDA performs well at generating diffuse conditions like bleeding and large objects like foreign bodies. 

IIDA requires a little amount of data for training. For example, synthetic lymphangiectasia generated from a model trained on 224 samples (Table \ref{tab:combined_kvasir_splits}) improved the F1-score for this class by 32.61 p.p. over RUS (Table \ref{tab:patientwise-per-class-comparison}). However, the data requirement is inconsistent. The model trained on 272 ulcer samples failed to enhance performance, leading to a 0.08 p.p. decrease in F1-score. This suggests that data requirements may depend on factors such as annotation quality, intra-class variance, and contrast. Furthermore, currently, the model is not class-conditional, requiring separate models for each class. Like PBDA, IIDA also relies on bounding boxes and benefits from identifying good blending locations.

Results in Section \ref{sec:experiments_classification} show that IIDA consistently outperforms the baseline across all architectures and augmented data sizes. IIDA also consistently achieves higher F1-score than PBDA, albeit with lower sensitivity. Moreover, IIDA shows a smaller reduction in F1-score for the unclear view class compared to PBDA (-6.21 vs. -29.88 p.p., Table \ref{tab:patientwise_augmented_sample_size}). These findings, along with the qualitative results in Section \ref{sec:experiments_lama} suggest that lesions generated by fine-tuned LaMa are more consistent, but less diverse, which can contribute to overfitting in classification models.

\subsection{Combining PBDA and IIDA}
\label{sec:discussion-pbda+iida}
The proposed solutions effectively utilize abundant healthy tissue to enhance lesion recognition. As shown in Table \ref{tab:patientwise_augmented_sample_size}, PBDA achieves higher sensitivity than IIDA, though with a lower F1-score. However, Figure \ref{fig:figure12} and Table \ref{tab:patientwise-per-class-comparison} demonstrate that a combination of PBDA and IIDA outperforms both methods individually, improving the F1-score to 33.07\% using the 2k dataset and sensitivity to 42.59\% when training with the 5k dataset. These methods adopt different data augmentation strategies: PBDA applies classical image processing techniques to blend lesions, whereas IIDA generates lesions while conditioned on background tissue. The results suggest that a combination of these methods produces a more diverse dataset, leading to better generalization. The experiments confirm that using both approaches together yields new state-of-the-art augmentation performance. 

Despite their complementary strengths, PBDA and IIDA have distinct requirements for successful application. As discussed in Sections \ref{sec:discussion-pbda} and \ref{sec:discussion-iida} PBDA excels at blending lesions that contrast with surrounding tissue, are clearly visible, accurately labeled, and localized. It enhances training by attempting to eliminate background bias \cite{wad2022equivariance} but only relocates existing lesions to new backgrounds. In contrast, IIDA adapts generated lesions' characteristics - such as shape, size, brightness, and texture - to match the surrounding tissue, creating potentially novel lesions on new backgrounds. However, as a generative model, IIDA may be prone to overfitting \cite{karras2020training}. Successful use of both methods requires a dataset that contains lesions appropriate for PBDA and sufficient samples for training IIDA's image inpainting model. Nevertheless, as shown in Section \ref {sec:experiments_classification} applying either method already improves performance significantly. By combining these approaches, researchers can create robust augmentation strategies that leverage the strengths of both PBDA and IIDA, advancing the field of lesion detection. 

\subsection{Comparison to other methods}
\label{sec:discussion-comparison}
\begin{figure}[htpb]
    \centering
    \includegraphics[width=0.7\textwidth]{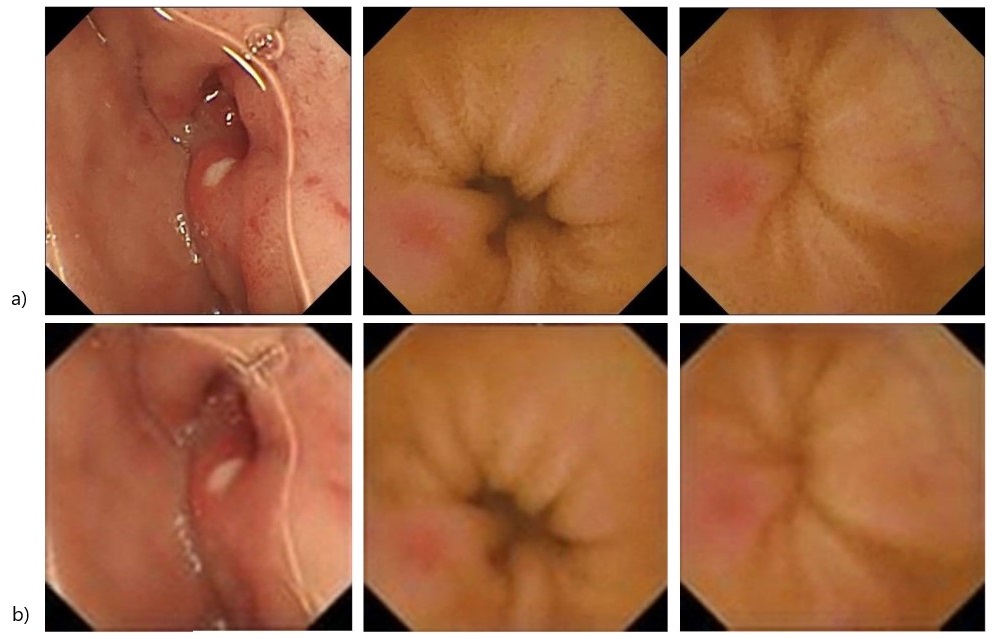}
    \caption{Data copying problem of generative models. Generative models, which generate images \textit{de novo}, when applied in a limited data regime, usually copy data. Row a) depicts real images from the Kvasir Capsule Dataset. Row b) shows images generated with VAE model of Diamantis et al. \protect\cite{diamantis2024intestine} . The data from \protect\cite{diamantis2024intestine} could be blurred, as it is copied directly from the manuscript (Figure 4) - we do not have access to the original data, nor the model. Despite the blurriness, it is clearly visible that this model copies the data from the training dataset, thus providing little value for Generative Data Augmentation.}
    \label{fig:figure13}
\end{figure}

In Sections \ref{sec:experiments_lama} and \ref{sec:experiments_classification} we conducted a comprehensive comparison of IIDA and PBDA against alternative methods. Both qualitative and quantitative experiments demonstrated the superiority of our approaches over augmentation using \textit{de novo} images generated by NVAE and LDM. These generative models occasionally produce global distortions and suffer from data copying issues \cite{bhattacharjee2023data,meehan2020non}, which is particularly evident when sampling \textit{de novo} images (Figure \ref{fig:figure13}). PBDA and IIDA may face similar challenges, as PBDA essentially replicates lesions in new locations, and IIDA, as a generative model, might also be prone to lesion copying. However, our methods differ by modifying only selected regions of existing images, limiting the potential for replication errors. The key question is whether our methods can introduce novel lesions into the dataset. As previously discussed, PBDA effectively mitigates dataset bias, while IIDA appropriately adjusts the generated lesions to the surrounding tissue. Although neither method can generate lesions outside the distribution of the training data, the combination of PBDA and IIDA has set a new state-of-the-art, as shown in Table \ref{tab:patientwise-comparison-to-reference}, which suggests that the issue of data copying is largely mitigated.

To evaluate the effect of balancing the training set via augmentation, we compared different data sampling strategies with our methods  (Table \ref{tab:patientwise_augmented_sample_size}). The best-performing sampling technique, RUS, improved F1-score by 2.15 p.p. over the baseline for the top-performing architecture, CAFormer. This improvement demonstrates the importance of balancing the training set, though PBDA and IIDA proved to be significantly more impactful overall.

We further compared our results to other studies that used the official Kvasir Capsule Dataset splits for training their models (Table \ref{tab:patientwise-comparison-to-reference}) \cite{regmi2023vision, smedsrud2021kvasir, srivastava2022video}.  Our methods established a new state-of-the-art on the Kvasir Capsule Dataset, improving F1-score by 7.84 p.p. over the previous best result. It is important to note that several recent studies \cite{malik2024multi, alam2022rat, oukdach2023conv, oukdach2022gastrointestinal} employed a random split, where consecutive frames from video sequences may appear in both the training and test sets, reducing the reliability of direct comparisons. Some studies \cite{fonseca2022abnormality} do not clearly report their data partitioning strategy, while others \cite{bordbar2023wireless, lima2022classification} deviate from the official dataset split. In contrast, our work adheres to the dataset split proposed by Smedsrud et al. \cite{smedsrud2021kvasir}, which better approximates real-world clinical scenarios by emphasizing performance on unseen patients. Additionally, our study only utilizes publicly available data, ensuring reproducibility.

The results in Table \ref{tab:patientwise_augmented_sample_size} also show that modern architectures learn more effectively from synthetic data. For instance, CAFormer, trained on the IIDA 2k dataset, achieved a 7.14 p.p. improvement in F1-score compared to training with class imbalance handling method, while ResNet improved by 4.99 p.p. These findings suggest that integrating modern architectures with novel augmentation techniques, such as those proposed in this work, is essential to overcoming the challenges posed by limited medical datasets.

Our methods also led to substantial improvements in lesion recognition, as shown in Table \ref{tab:patientwise-per-class-comparison}. This advance addresses the common issue of limited data and expert availability in medical image analysis, potentially accelerating the development of practical Computer-Aided Diagnosis (CAD) systems for clinical use \cite{chan2020computer, hwang2018application}. Accurate and automatic lesion detection in capsule endoscopy could enhance the precision, consistency, and efficiency of the diagnostic process.

\subsection{Future work}
\label{sec:discussion-future-work}
The success of the proposed methods opens new research avenues in Generative Data Augmentation. Future work should focus on evaluating the generality of IIDA and PBDA by testing their effectiveness across diverse datasets and imaging tasks. Improvements could also be made by developing more advanced ROI selection algorithms, such as those using object detection, semantic segmentation, or depth estimation. Additionally, incorporating a class-conditional inpainting model for IIDA may enhance performance. Applying both methods with irregular masks, which were unavailable in the Kvasir Capsule Dataset, could further improve results. Since both methods can utilize bounding boxes and masks, the synthesized data is automatically annotated, making it suitable for evaluation in object detection and semantic segmentation tasks.

While the primary goal of data augmentation is to enhance the performance of downstream models, conducting a detailed analysis of the generated data with input from domain experts could provide valuable insights. Such a study might assess the preference rate of IIDA and PBDA compared to other methods and evaluate whether experts can distinguish between synthetic and real lesions. 

Future extensions of these methods could also involve video inpainting techniques \cite{quan2024deep, kim2019deep} to take advantage of the spatio-temporal characteristics of capsule endoscopy data. By generating consistent lesions across consecutive frames, these methods could better mimic the appearance of real lesions in video sequences.

\section{Conclusion}
\label{sec:conclusion}
This study demonstrates that high-quality data augmentation is feasible even in low-data settings common to medical imaging. The proposed approach combines conditional generative models, specifically Image Inpainting Data Augmentation, with classical image processing techniques, such as Poisson Blending Data Augmentation. When applied to the Kvasir Capsule Dataset, both methods independently establish new state-of-the-art benchmarks and demonstrate even greater effectiveness when combined.

A key innovation of this approach is its reliance on generating local lesions within abundant images of healthy tissue, which enhances downstream tasks. This contrasts with GDA methods that rely on \textit{de novo} image generation. By avoiding data copying, the proposed techniques produce high-quality, diverse synthetic data, significantly improving classification model performance.

\section*{CRediT authorship contribution statement}
\textbf{Adrian B. Chłopowiec:} Writing - original draft, Validation, Visualization, Software, Methodology, Data curation, Conceptualization, Investigation. \textbf{Adam R. Chłopowiec:} Writing - original draft, Validation, Software, Methodology, Conceptualization, Formal analysis, Investigation. \textbf{Krzysztof Galus:} Writing - original draft, Validation, Software, Visualization, Data curation. \textbf{Wojciech Cebula:} Writing - original draft, Validation, Software. \textbf{Martin Tabakov:} Writing - original draft, Visualization, Conceptualization, Supervision, Formal analysis.

\section*{Declaration of competing interest}
The authors declare the following financial interests/personal relationships which may be considered as potential competing interests: Adrian B. Chłopowiec, Adam R. Chłopowiec and Martin Tabakov were employed by the company BioCam LLC.
All other authors declare that they have no competing interests.

\section*{Data availability}
Kvasir Capsule Dataset is open-source and cited where appropriate. To access the dataset please refer to relevant citation for link and guideline. All data enquiries should be mailed to chlopowiec.adrian@gmail.com.

\section*{Declaration of generative AI and AI-assisted technologies in the writing process.}
During the preparation of this work the authors used ChatGPT in order to improve readability and language of the manuscript text. After using this tool, the authors reviewed and edited the content as needed and take full responsibility for the content of the published article.

\section*{Funding}
This work was partially supported by the statutory funds of the Department of Artificial Intelligence, Wroclaw University of Science and Technology and partially supported by the Polish Ministry of Science and Higher Education, NCBR grant No POIR.01.01.01-00-2504/20.

\section*{Acknowledgements}
We would like to express our gratitude to Michał Karol for his substantive help, guidance and engaging discussions.

{\small
\bibliographystyle{ieee_fullname}

}

\clearpage

\onecolumn

\appendix
\section{Algorithms details}
\label{appendix:a}

\begin{algorithm}[htbp]
    \caption{The deduplication algorithm improves dataset diversity by iteratively removing redundant images from each video. It calculates the latent vector distance between images and discards those with distances below an experimentally determined threshold.}
    \label{alg:deduplication}
    \begin{algorithmic}
        \State \texttt{final\_images} $\gets$ list()
        \State \texttt{threshold} $\gets$ T \Comment{Experimentally chosen threshold for latent vectors distance, T = 356}
        \State \texttt{train\_dataset} $\gets$ \texttt{load\_train\_data()}
         \Comment{Contains latent vectors of images}
        \State \texttt{patients\_datasets} $\gets$
        \texttt{train\_dataset.groupby('patient\_id')}
        \State
        \For{\texttt{patient\_dataset in patients\_datasets}}
            \While{\texttt{cardinality(patient\_dataset) > 0}}
                \State \texttt{query} $\gets$ \texttt{sample(patient\_dataset)} \Comment{Sample one image from patient dataset}
                \State \texttt{drop(patient\_dataset, query)}
                \State \texttt{distances} $\gets$ \texttt{find\_sim\_imgs(query, patient\_dataset)}
                \State \Comment{Find most similar latent vectors by l2 distance}
                \State \texttt{duplicated} $\gets$ \texttt{get\_duplicates\_leq\_threshold(distances, threshold)}
                \State \texttt{drop(patient\_dataset, duplicated)}
                \State \texttt{final\_images.append(query)}
            \EndWhile
        \EndFor
        \State
        \State \texttt{// load\_train\_data() - loads the training data}
        \State \texttt{// train\_dataset.groupby() - groups the images by specified parameter}
        \State \texttt{// sample() - samples an image from the dataset}
        \State \texttt{// drop() - removes an image from the dataset}
        \State \texttt{// find\_sim\_imgs() - returns distances from query image to all other images}
        \State \texttt{// get\_duplicates\_leq\_threshold() - returns indices of images, for which distance to the query image are lower than specified threshold}
    \end{algorithmic}
\end{algorithm}

\clearpage

\begin{figure}
    \centering
    \includegraphics[width=\linewidth]{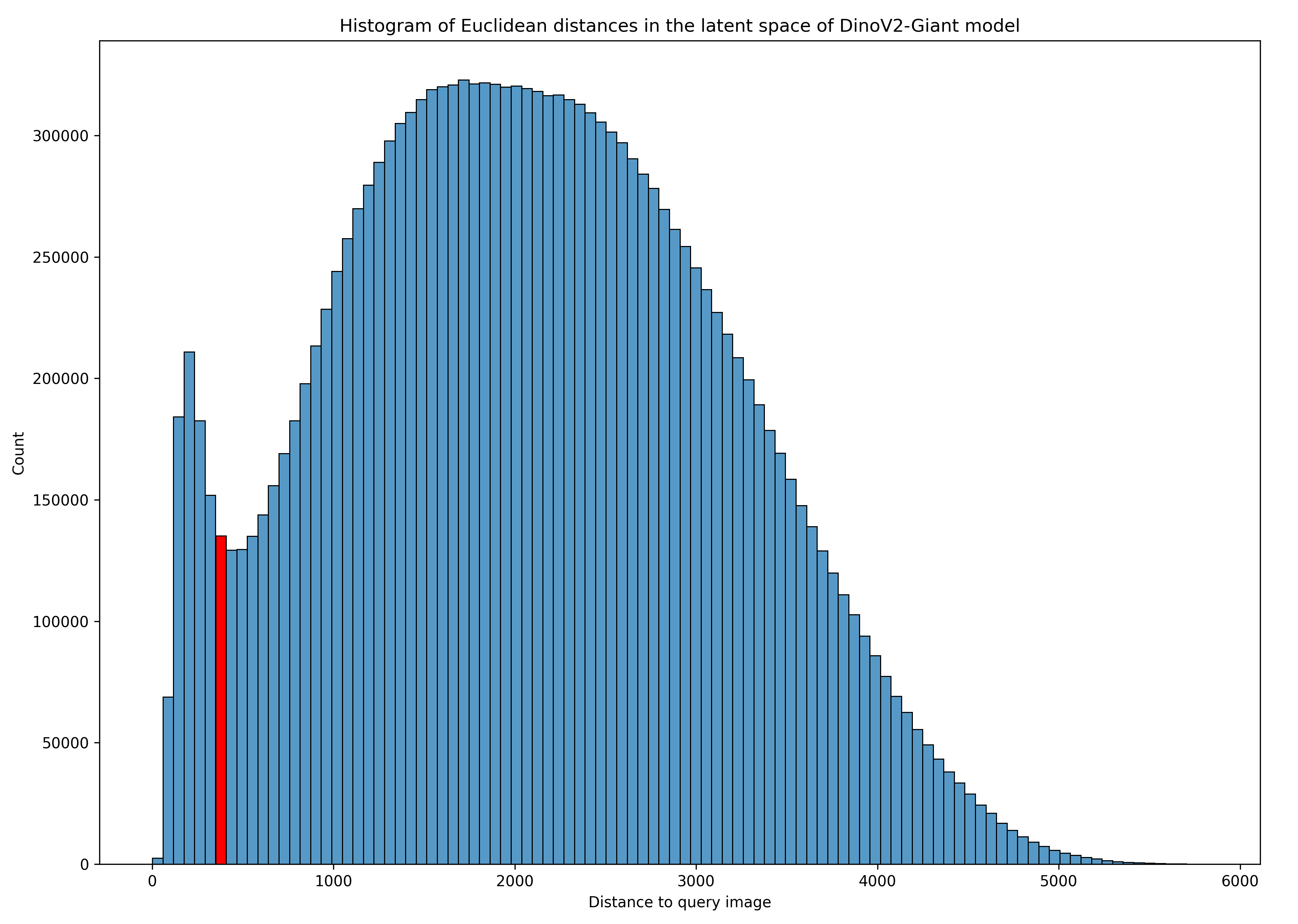}
    \caption{Histogram of Euclidean distances in the latent space of DinoV2-Giant model \cite{oquab2023dinov2}. The distances are calculated pairwise for images from individual videos and aggregated across all patient videos in the train dataset. The red bar represents the chosen threshold of 356, positioned at the division between two modes in the bimodal distribution. The first mode primarily reflects nearly duplicate images. The threshold was set slightly toward the first mode to prioritize the inclusion of all clinically relevant images over the complete elimination of redundancy.}
\end{figure}

\begin{algorithm}[htbp]
    \caption{The image pair selection algorithm improves the quality of generated samples by choosing the most similar source and target image pairs. For each given image, it selects the closest neighbor in the DinoV2 latent space. }
    \label{alg:place_selection}
    \begin{algorithmic}
        \State \texttt{lesions\_dataset} $\gets$ \texttt{load\_train\_data("lesion")}
        \State \Comment{Get the train dataset of only lesion samples}
        \State \texttt{normals\_dataset} $\gets$ \texttt{load\_train\_data("normal")}
        \State \Comment{Get the train dataset of only normal samples}
        \State \texttt{lesion\_sample} $\gets$ \texttt{sample(lesions\_dataset)}
        \State \texttt{sample\_patient} $\gets$ \texttt{lesion\_sample.get("patient\_id")}
        \State \texttt{normals\_dataset} $\gets$ \texttt{normals\_dataset.remove(sample\_patient)}
        \State \Comment{Remove all sample\_patient images from normals\_dataset}
        \State \texttt{matching\_normal\_sample} $\gets$ \texttt{find\_sim\_img(sample, normals\_dataset)}
        \State \Comment{Get the closest normal image by l2 distance on DinoV2 latents}
        \State
        \State \texttt{// load\_train\_data() - loads the training data}
        \State \texttt{// lesion\_sample.get() - gets the specified information about sample}
        \State \texttt{// normals\_dataset.remove() - removes all elements according to parameter}
        \State \texttt{// sample() - samples an image from the dataset}
        \State \texttt{// find\_sim\_img() - returns the most similar image to the sample by l2 distance from the given dataset}
    \end{algorithmic}
\end{algorithm}

\clearpage
\section{Additional samples generated with PBDA}
\label{appendix:b}
\begin{figure}[h]
    \centering
    \includegraphics[width=0.50\textwidth]{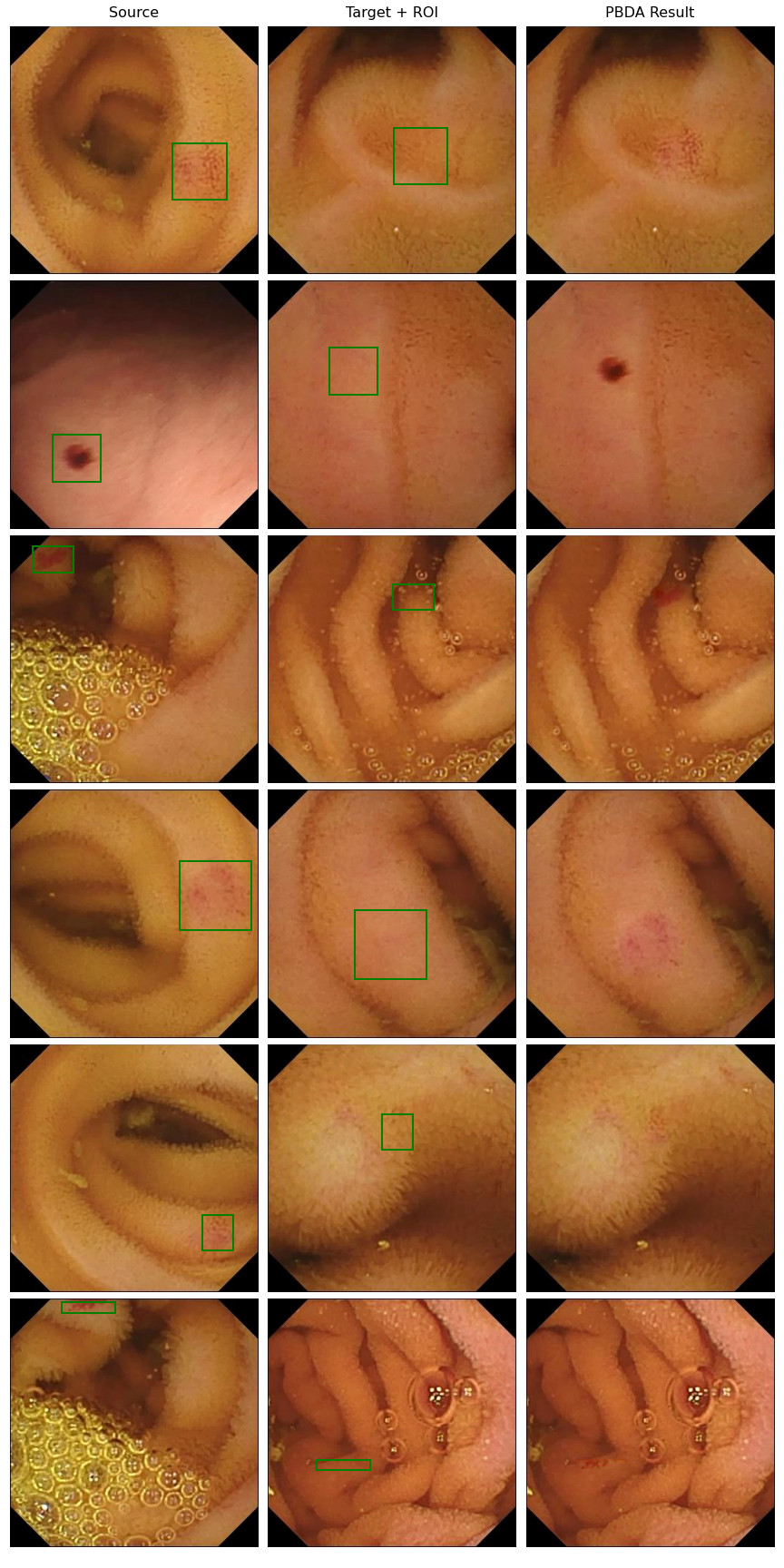}
    \caption{Qualitative examples of generated Angiectasia.}
    \label{fig:B_figure1}
\end{figure}

\begin{figure}[h]
    \centering
    \includegraphics[width=0.6\textwidth]{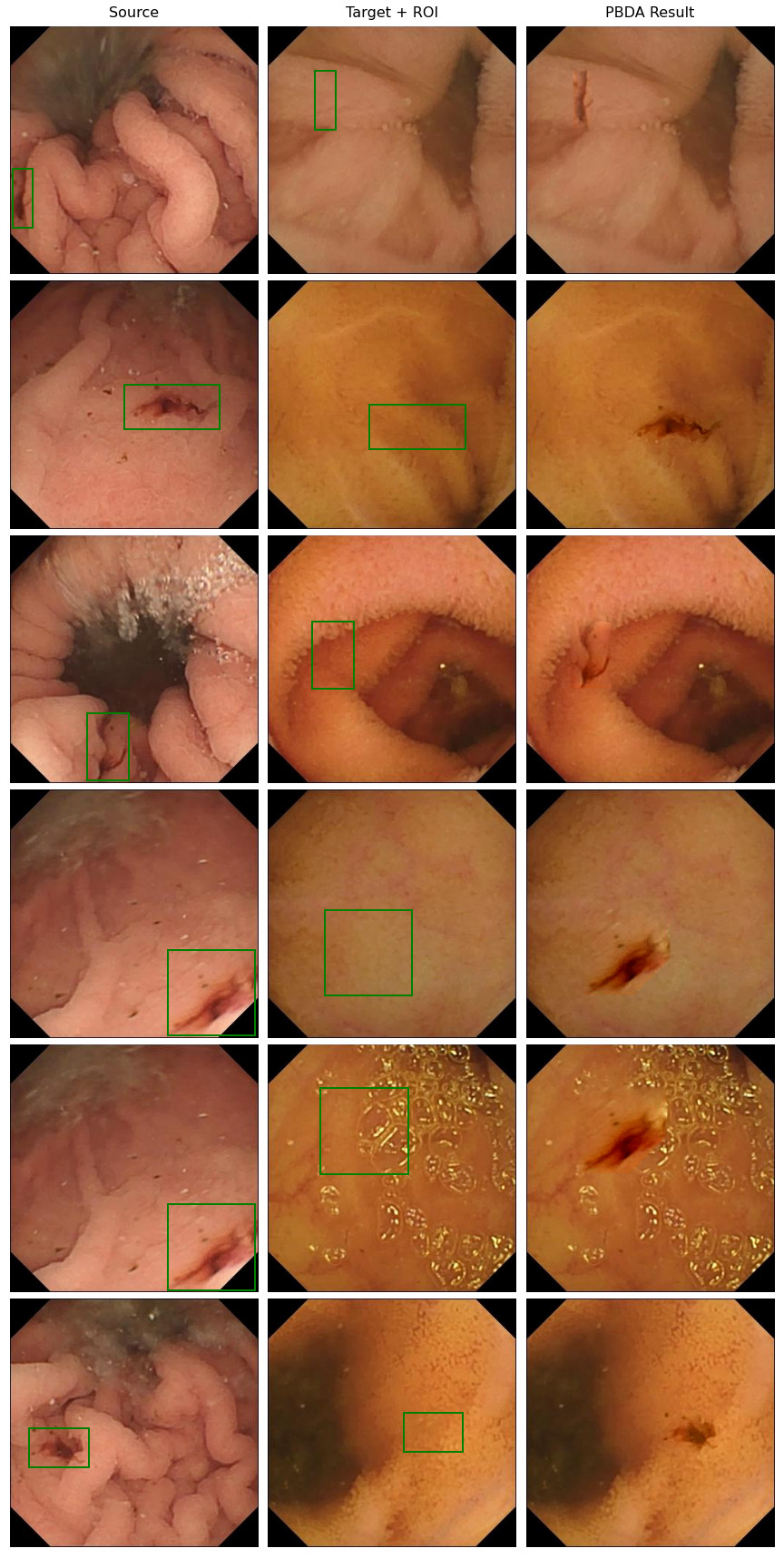}
    \caption{Qualitative examples of generated Fresh Blood.}
    \label{fig:B_figure2}
\end{figure}

\begin{figure}[h]
    \centering
    \includegraphics[width=0.6\textwidth]{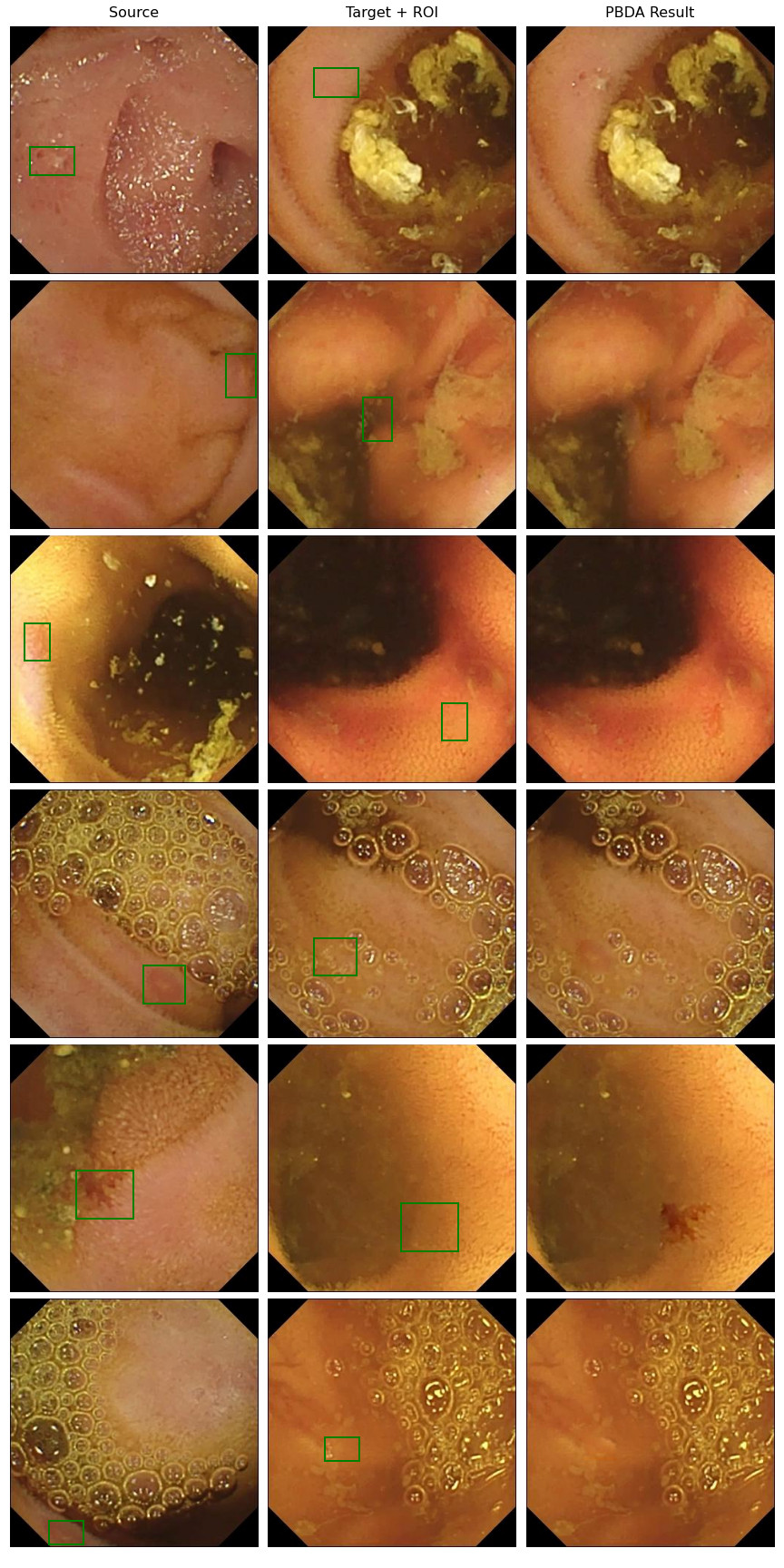}
    \caption{Qualitative examples of generated Erosion.}
    \label{fig:B_figure3}
\end{figure}

\begin{figure}[h]
    \centering
    \includegraphics[width=0.6\textwidth]{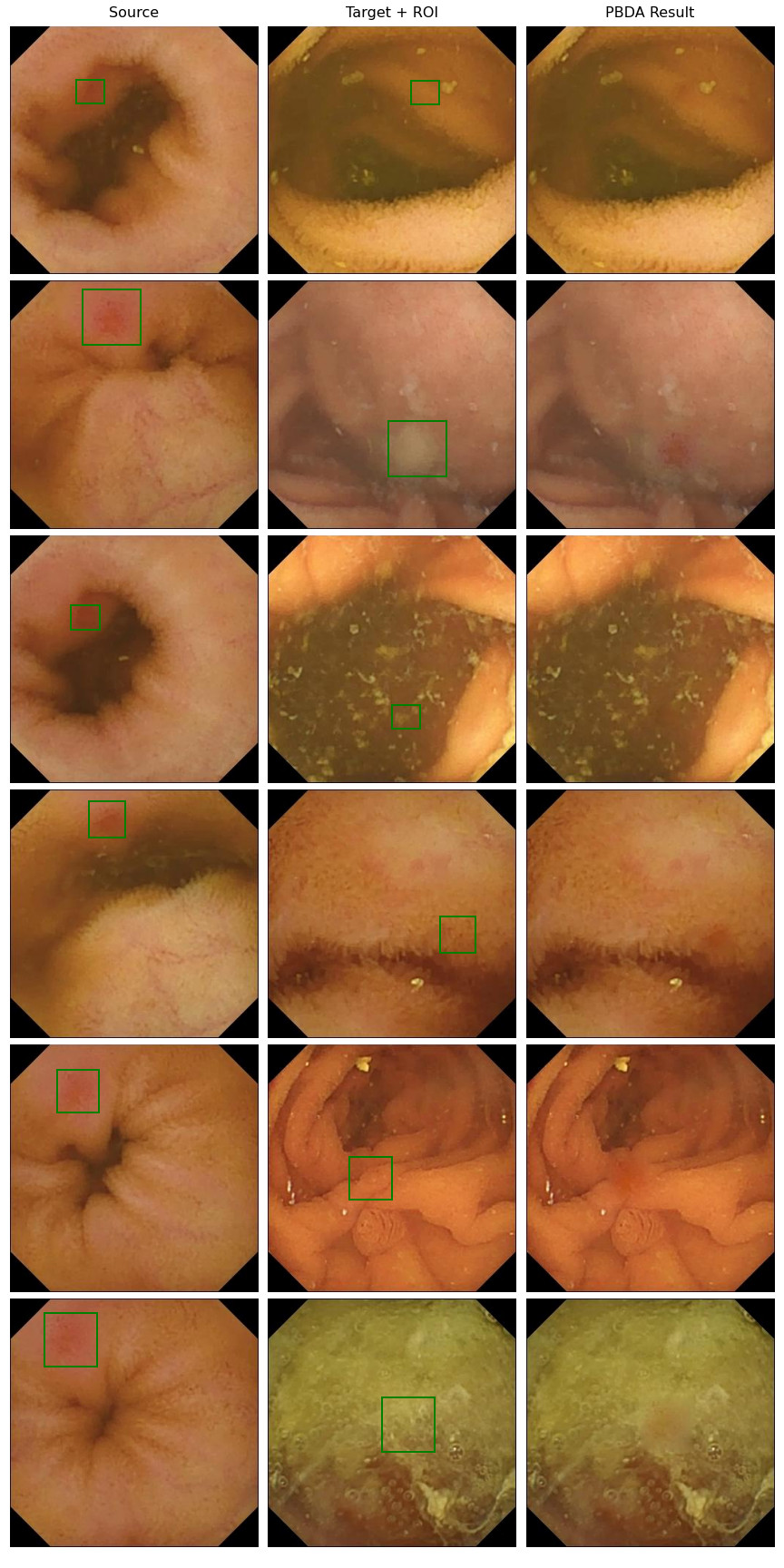}
    \caption{Qualitative examples of generated Erythema.}
    \label{fig:B_figure4}
\end{figure}

\begin{figure}[h]
    \centering
    \includegraphics[width=0.6\textwidth]{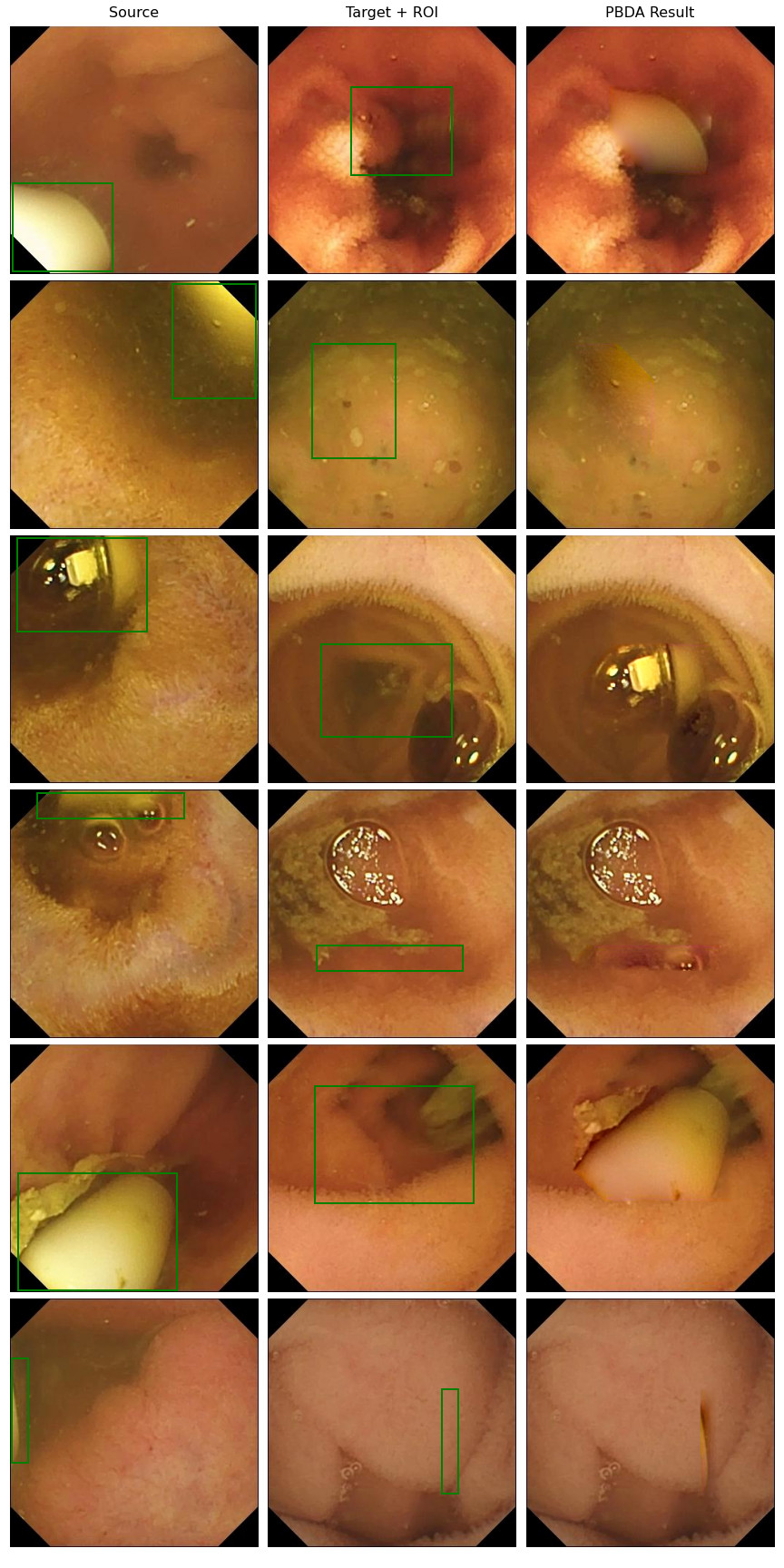}
    \caption{Qualitative examples of generated Foreign Body.}
    \label{fig:B_figure5}
\end{figure}

\begin{figure}[h]
    \centering
    \includegraphics[width=0.6\textwidth]{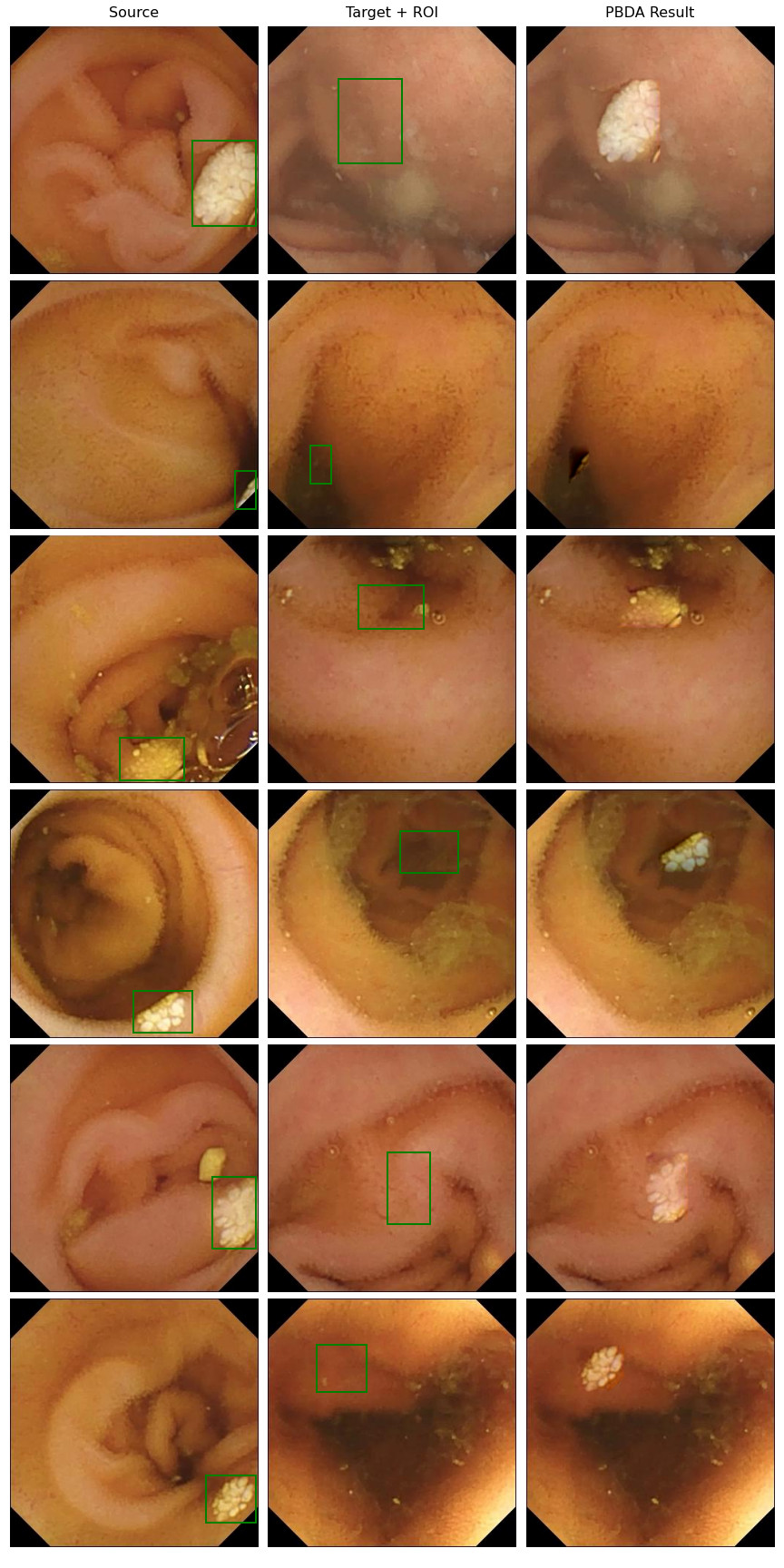}
    \caption{Qualitative examples of generated Lymphangiectasia.}
    \label{fig:B_figure6}
\end{figure}

\begin{figure}[h]
    \centering
    \includegraphics[width=0.6\textwidth]{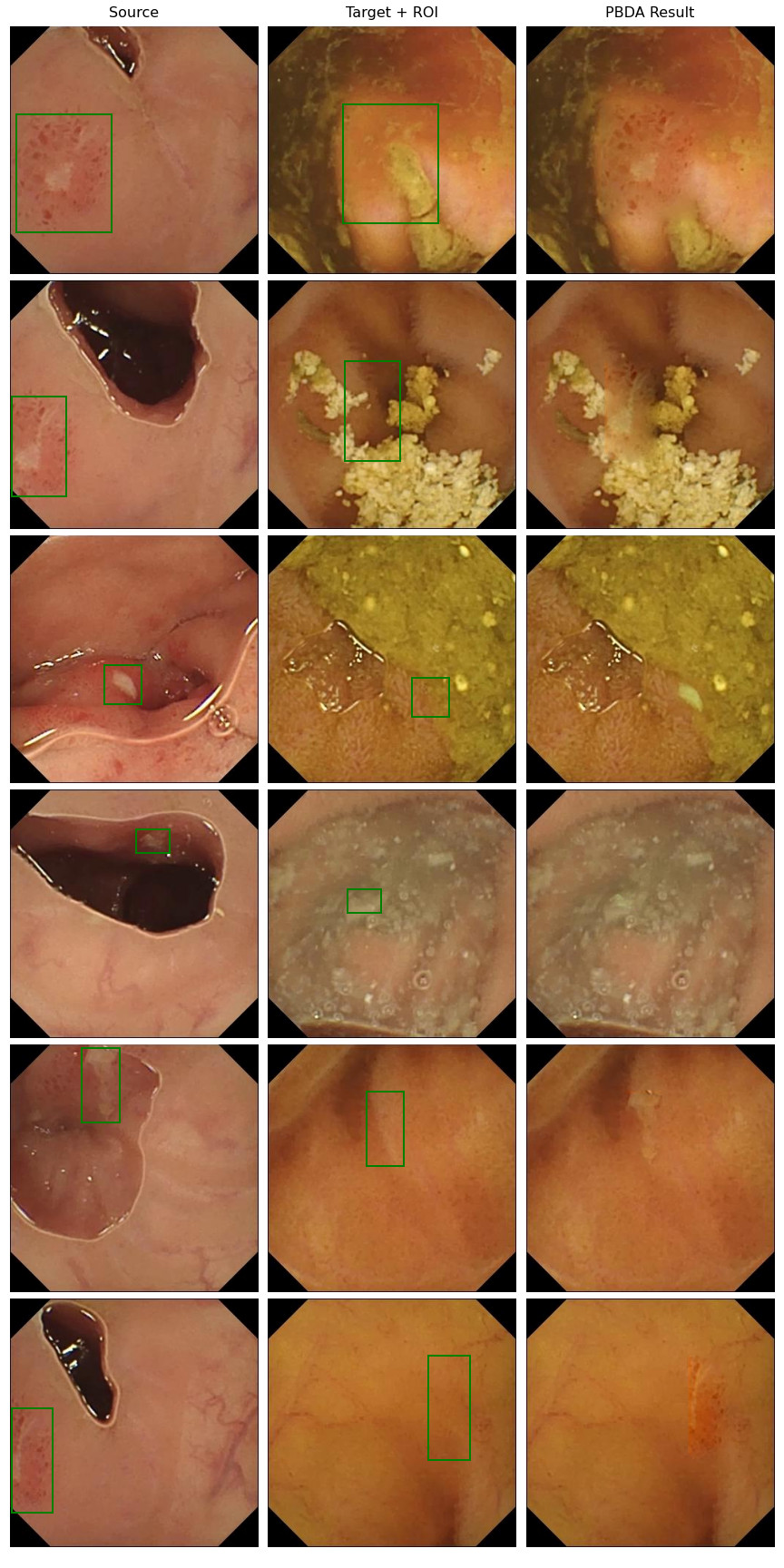}
    \caption{Qualitative examples of generated Ulcer.}
    \label{fig:B_figure7}
\end{figure}
\clearpage
\section{Additional samples generated with IIDA}
\label{appendix:c}
\begin{figure}[h]
    \centering
    \includegraphics[width=\textwidth]{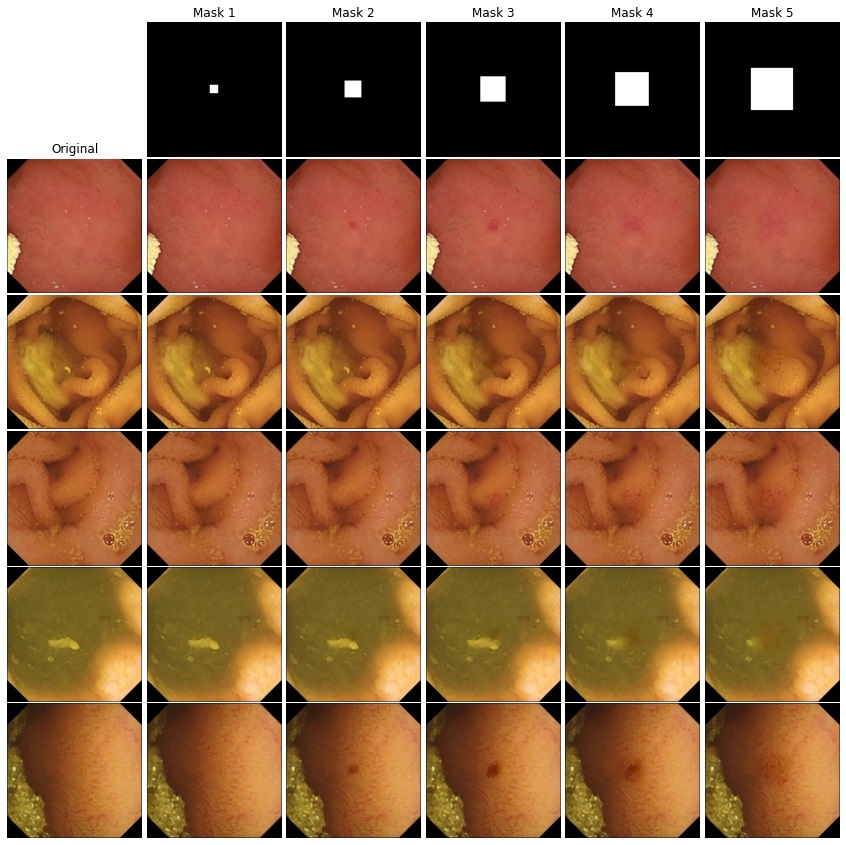}
    \caption{Qualitative examples of generated Angiectasia w.r.t different mask sizes in the center of the image.}
    \label{fig:C_figure1}
\end{figure}

\begin{figure}[h]
    \centering
    \includegraphics[width=\textwidth]{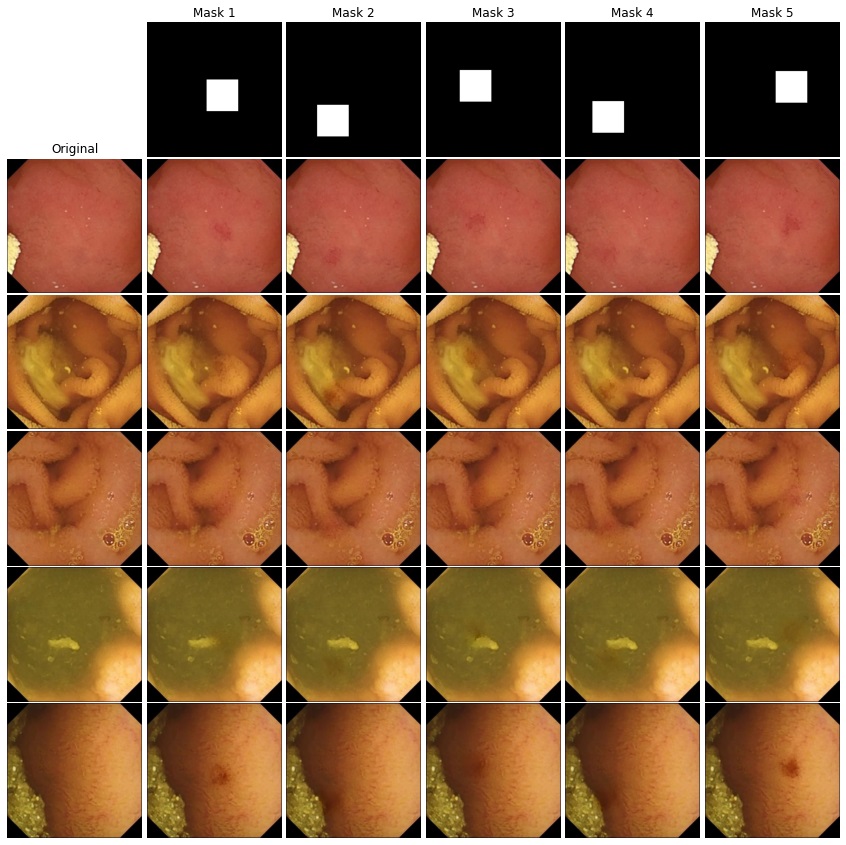}
    \caption{Qualitative examples of generated Angiectasia w.r.t the same mask size in different places in the image.}
    \label{fig:C_figure2}
\end{figure}

\begin{figure}[h]
    \centering
    \includegraphics[width=\textwidth]{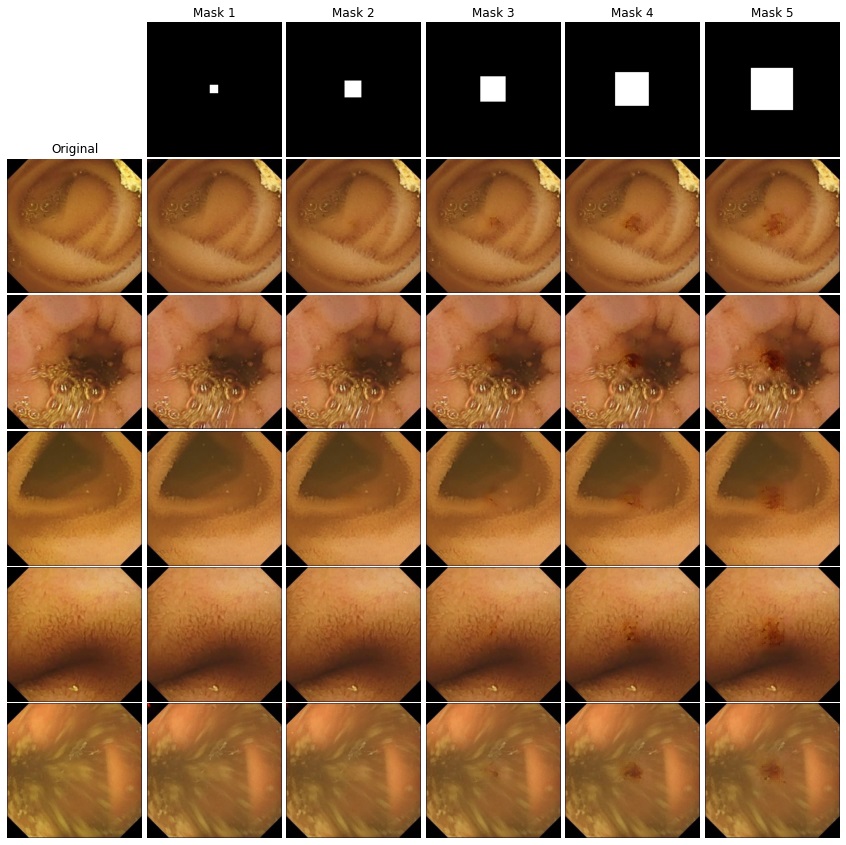}
    \caption{Qualitative examples of generated Fresh Blood class w.r.t different mask sizes in the center of the image.}
    \label{fig:C_figure3}
\end{figure}

\begin{figure}[h]
    \centering
    \includegraphics[width=\textwidth]{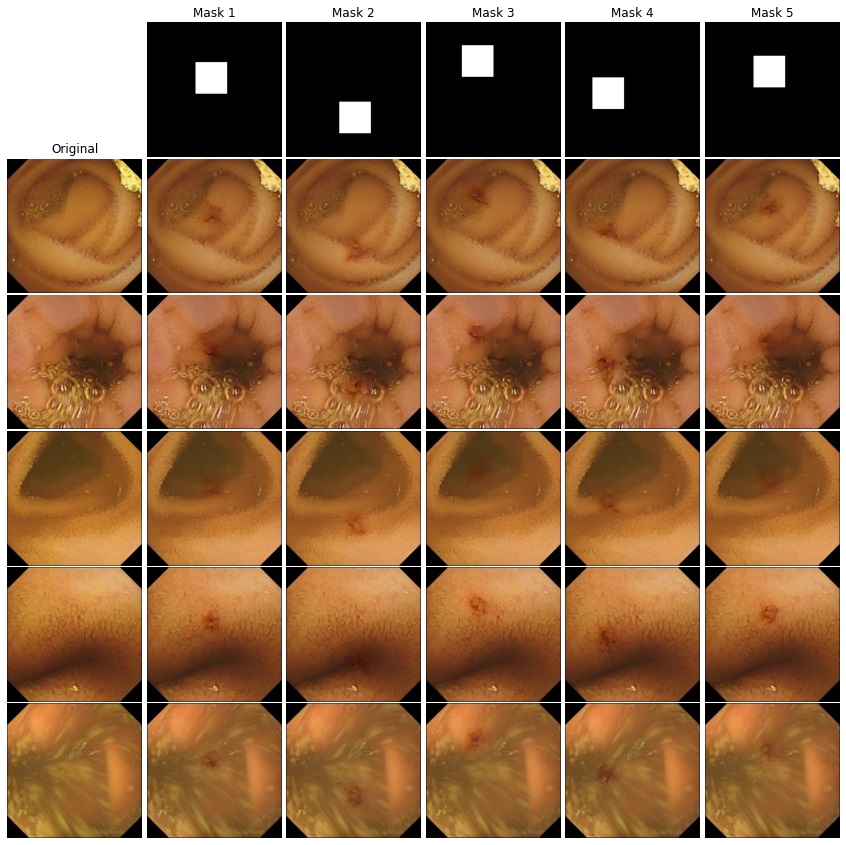}
    \caption{Qualitative examples of generated Fresh Blood w.r.t the same mask size in different places in the image.}
    \label{fig:C_figure4}
\end{figure}

\begin{figure}[h]
    \centering
    \includegraphics[width=\textwidth]{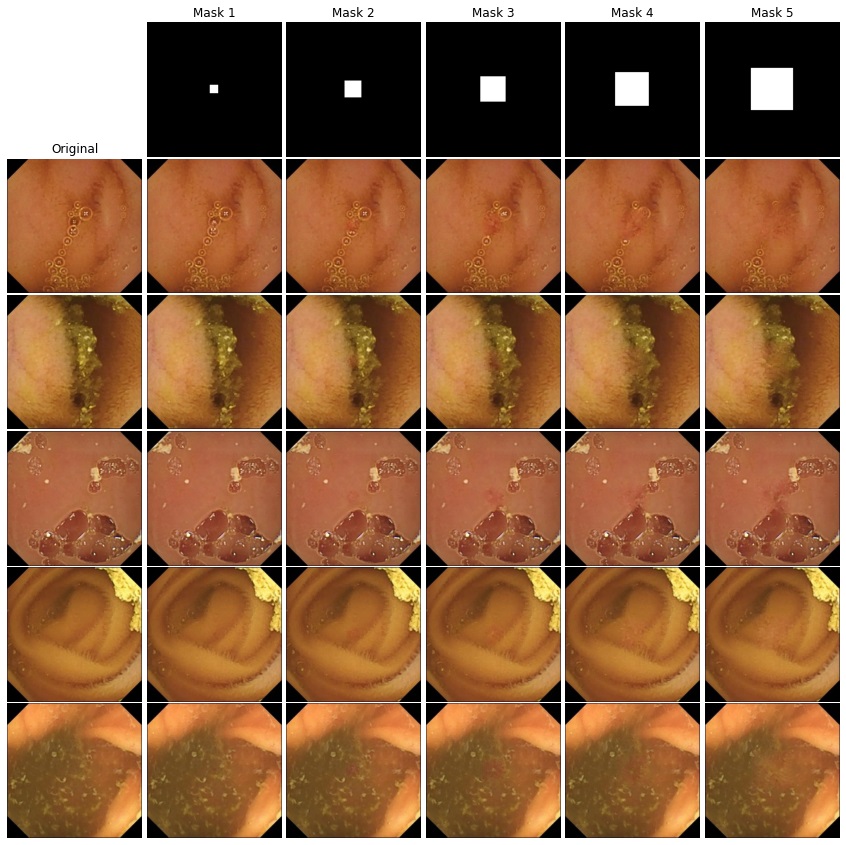}
    \caption{Qualitative examples of generated Erosion w.r.t different mask sizes in the center of the image.}
    \label{fig:C_figure5}
\end{figure}

\begin{figure}[h]
    \centering
    \includegraphics[width=\textwidth]{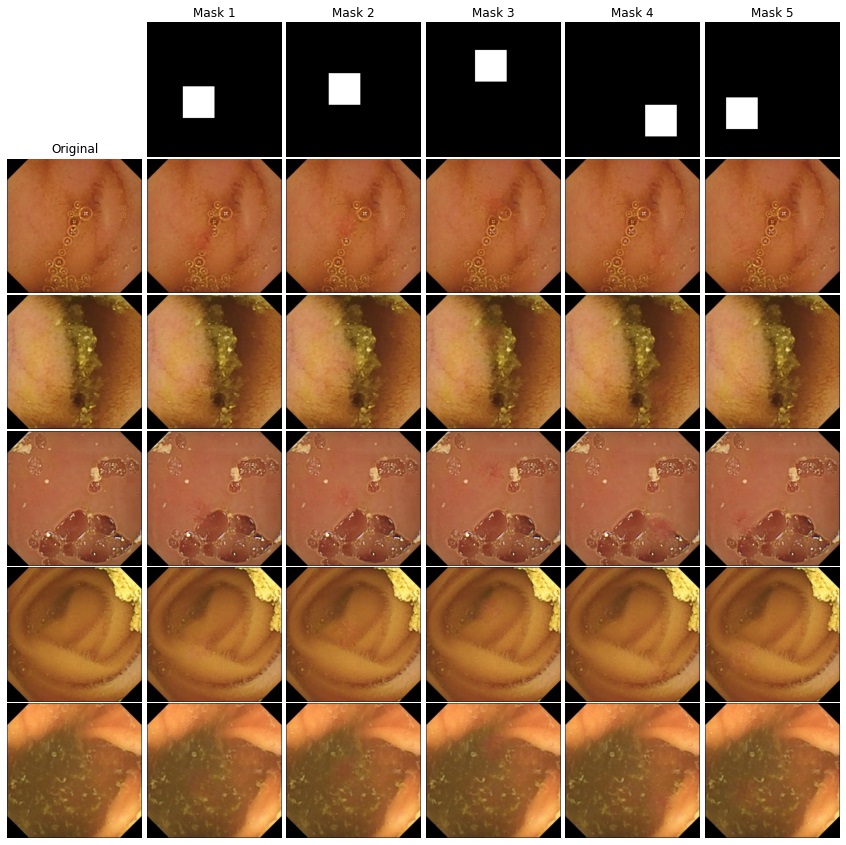}
    \caption{Qualitative examples of generated Erosion w.r.t the same mask size in different places in the image.}
    \label{fig:C_figure6}
\end{figure}

\begin{figure}[h]
    \centering
    \includegraphics[width=\textwidth]{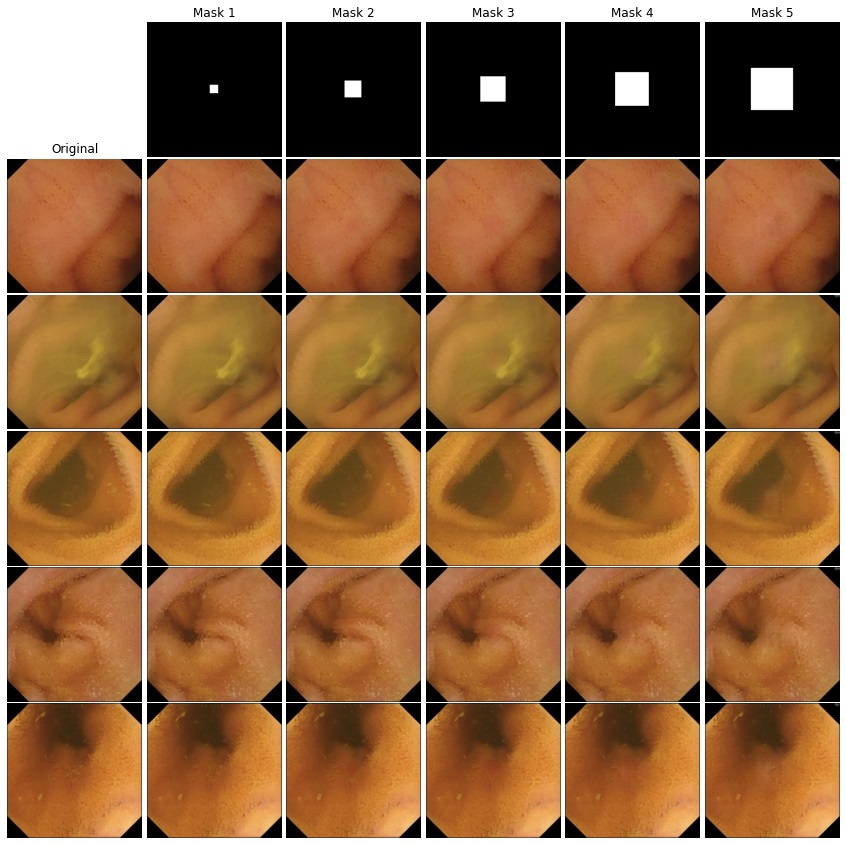}
    \caption{Qualitative examples of generated Erythema w.r.t different mask sizes in the center of the image.}
    \label{fig:C_figure7}
\end{figure}

\begin{figure}[h]
    \centering
    \includegraphics[width=\textwidth]{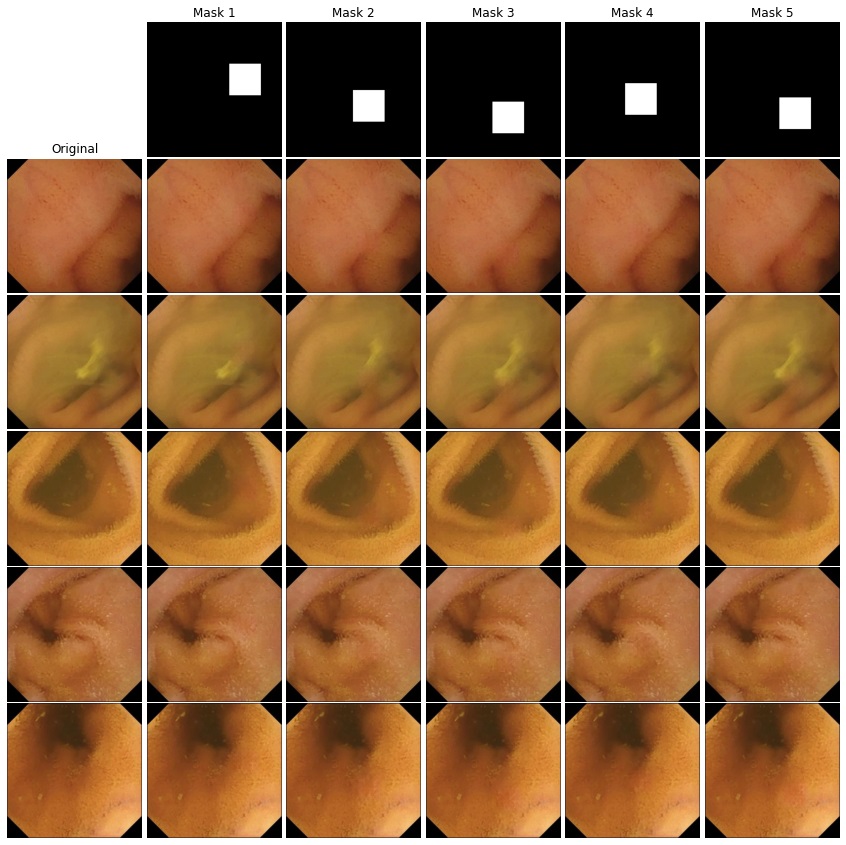}
    \caption{Qualitative examples of generated Erythema w.r.t the same mask size in different places in the image.}
    \label{fig:C_figure8}
\end{figure}

\begin{figure}[h]
    \centering
    \includegraphics[width=\textwidth]{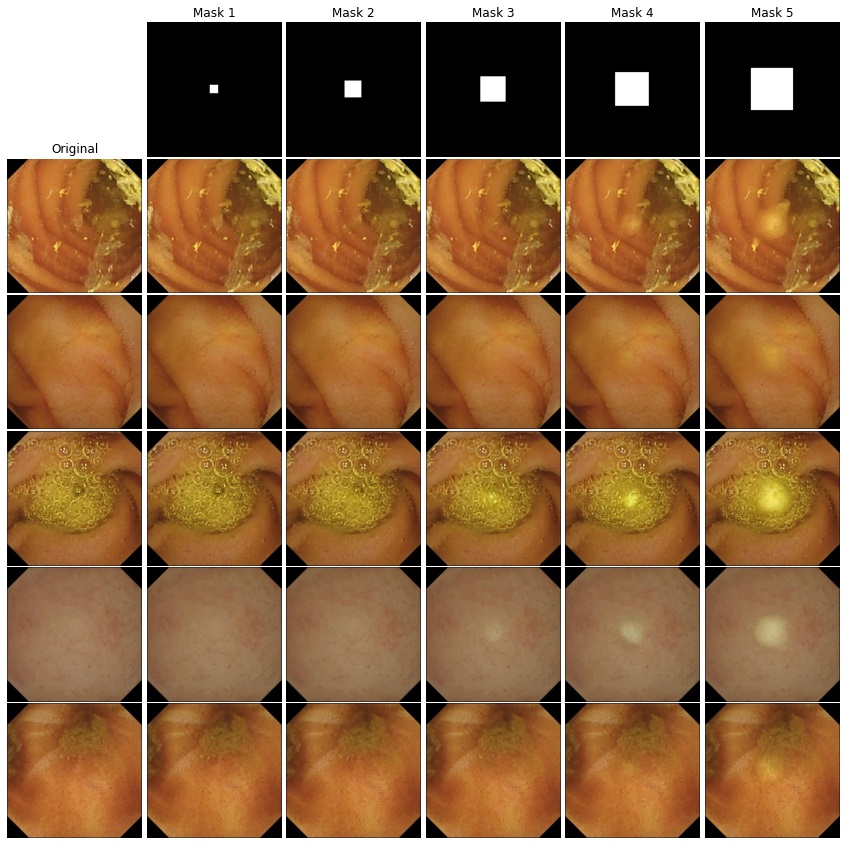}
    \caption{Qualitative examples of generated Foreign Body w.r.t different mask sizes in the center of the image.}
    \label{fig:C_figure9}
\end{figure}

\begin{figure}[h]
    \centering
    \includegraphics[width=\textwidth]{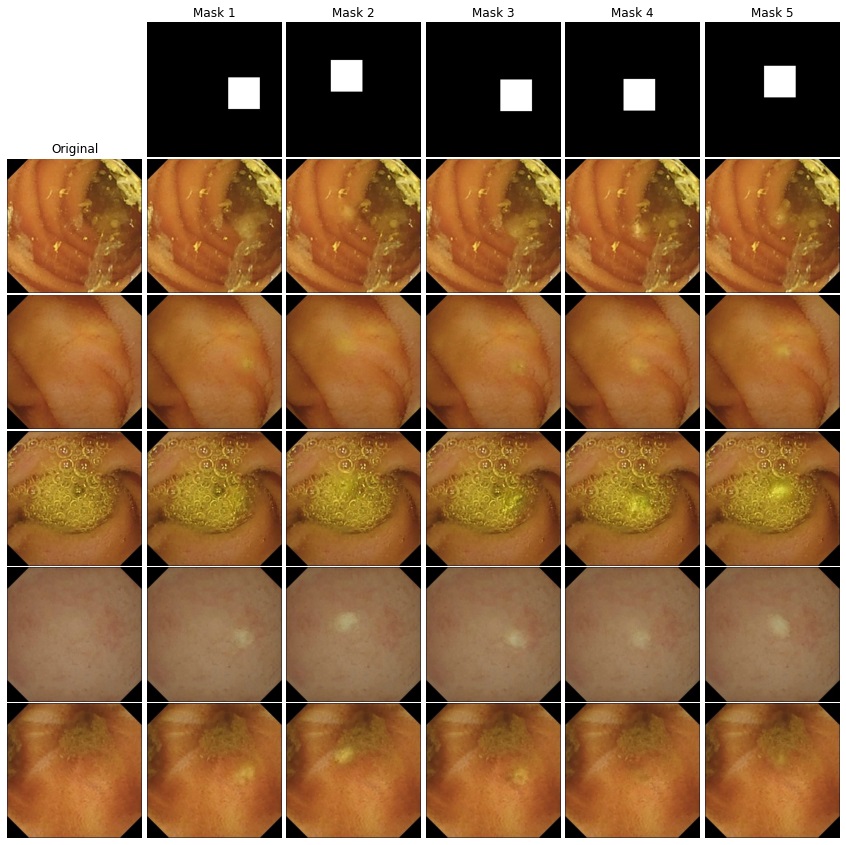}
    \caption{Qualitative examples of generated Foreign Body w.r.t the same mask size in different places in the image.}
    \label{fig:C_figure10}
\end{figure}

\begin{figure}[h]
    \centering
    \includegraphics[width=\textwidth]{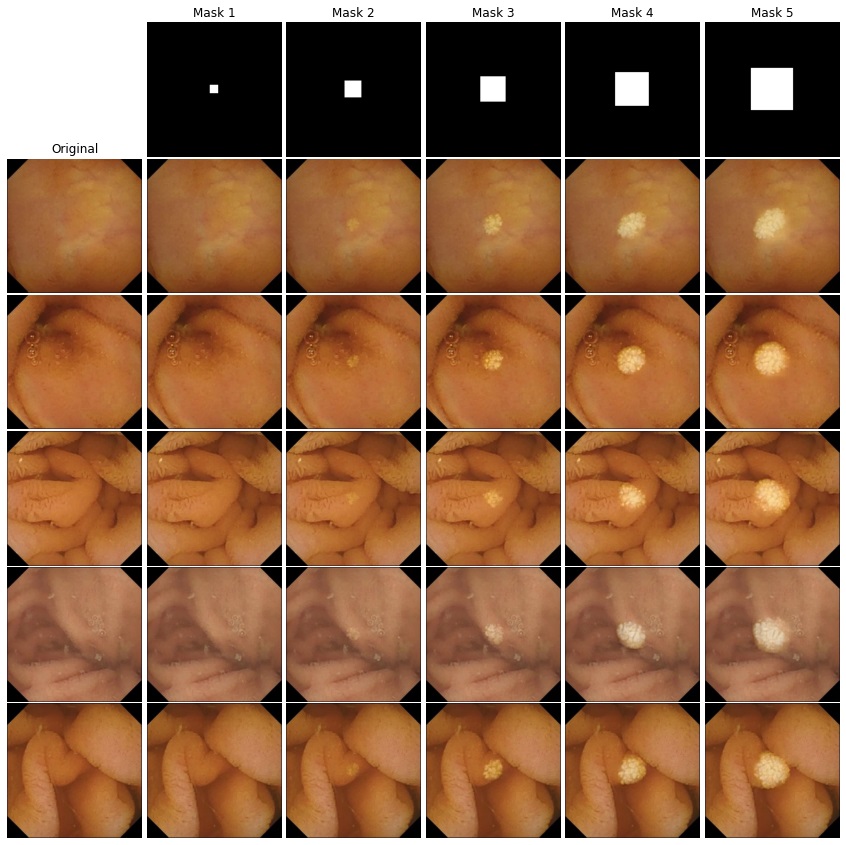}
    \caption{Qualitative examples of generated Lymhpangiectasia w.r.t different mask sizes in the center of the image.}
    \label{fig:C_figure11}
\end{figure}

\begin{figure}[h]
    \centering
    \includegraphics[width=\textwidth]{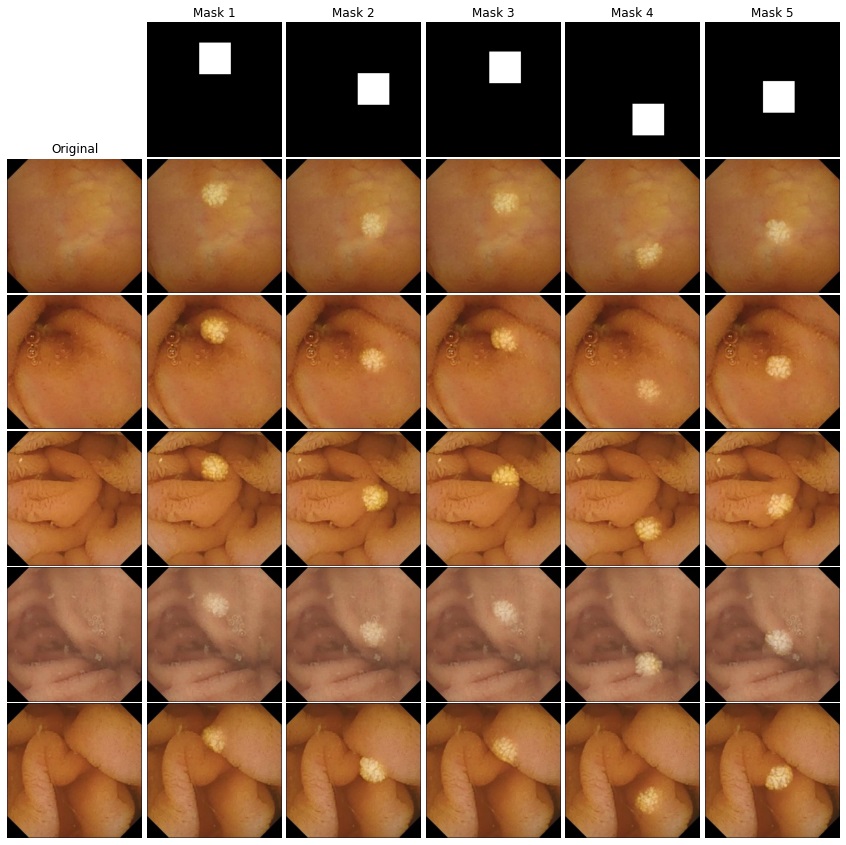}
    \caption{Qualitative examples of generated Lymphangiectasia w.r.t the same mask size in different places in the image.}
    \label{fig:C_figure12}
\end{figure}

\begin{figure}[h]
    \centering
    \includegraphics[width=\textwidth]{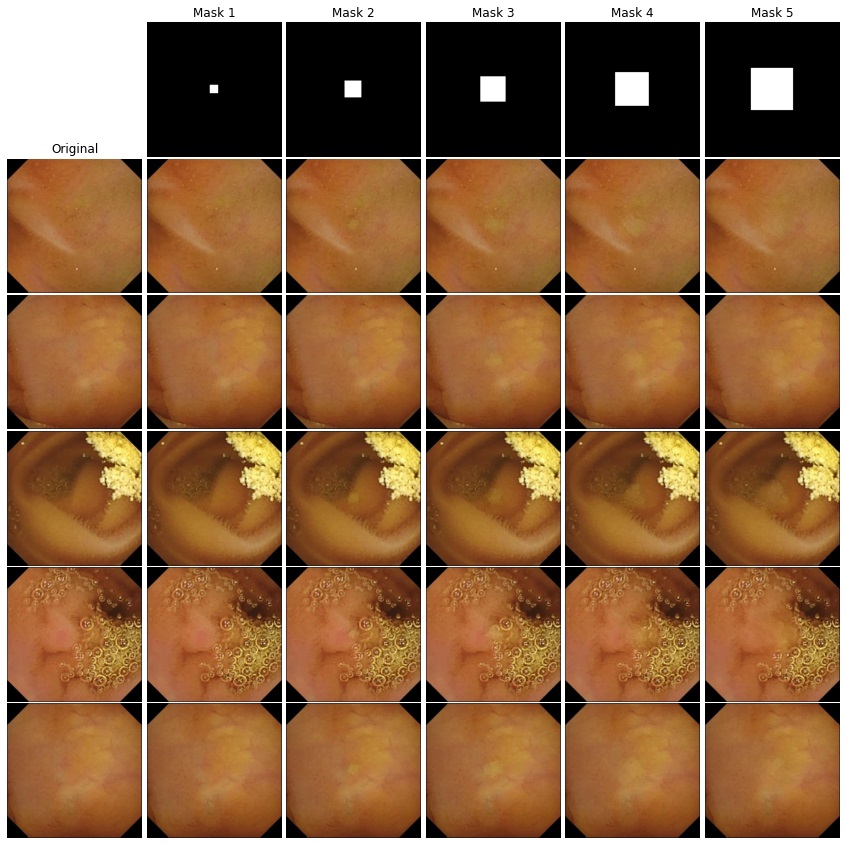}
    \caption{Qualitative examples of generated Ulcer w.r.t different mask sizes in the center of the image.}
    \label{fig:C_figure13}
\end{figure}

\begin{figure}[h]
    \centering
    \includegraphics[width=\textwidth]{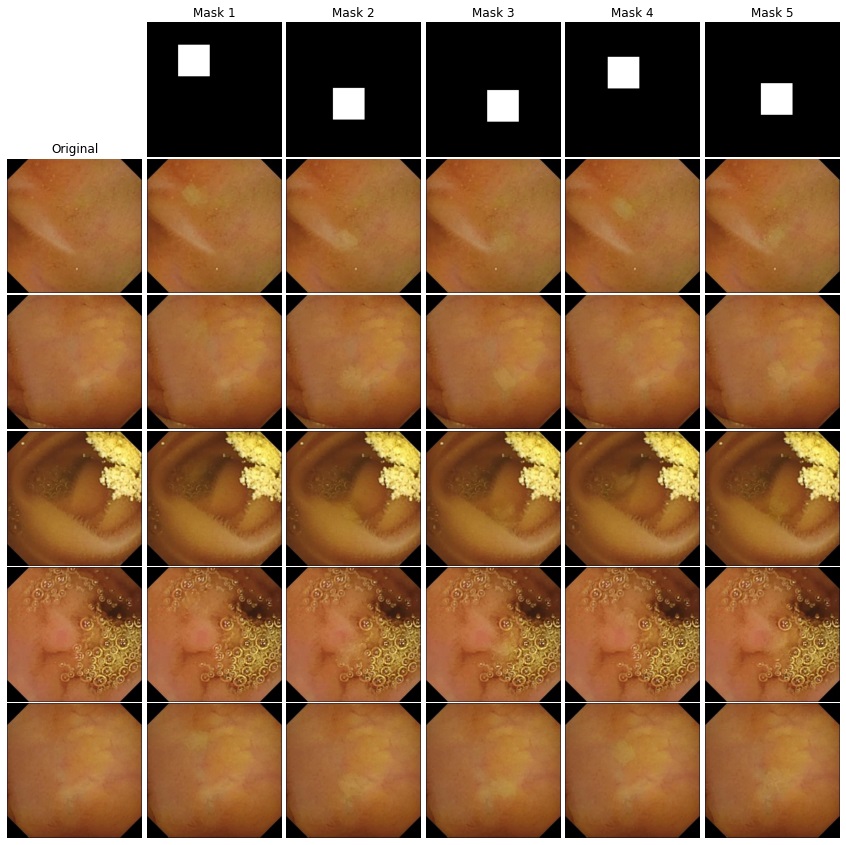}
    \caption{Qualitative examples of generated Ulcer w.r.t the same mask size in different places in the image.}
    \label{fig:C_figure14}
\end{figure}

\end{document}